%% file: main.tex
\documentclass[letterpaper, 10 pt, conference]{ieeeconf}  

\IEEEoverridecommandlockouts                              

\usepackage{algorithm}
\usepackage{algorithmic}
\usepackage{amsmath}
\usepackage{amstext}
\usepackage{amssymb}
\usepackage{amsfonts}
\usepackage{mathtools}

\usepackage{amsthm}
\usepackage{latexsym}
\usepackage{bbm}
\usepackage{balance}
\usepackage{dsfont}
\usepackage{nicefrac}
\usepackage[super]{nth}
\usepackage{wasysym}
\usepackage{pifont}

\DeclareMathOperator*{\argmax}{\mathrm{argmax}}
\DeclareMathOperator*{\argmin}{\mathrm{argmin}}
\newcommand*\diff{\mathop{}\!\mathrm{d}}

\newcommand{\ie}{\textit{i.e.}}
\newcommand{\eg}{\textit{e.g.}}

\newcommand{\method}{\textsc{SFT-CoP}}
\newcommand{\sfql}{\textsc{SFQL}}
\newcommand{\pdql}{\textsc{PDQL}}
\newcommand{\rasfql}{\textsc{RaSFQL}}
\newcommand{\cpo}{\textsc{CPO}}

\usepackage{booktabs}
\usepackage{epsfig}
\usepackage{graphicx}
\usepackage{float}
\usepackage{wrapfig}
\usepackage{textcomp}
\usepackage{subcaption}
\usepackage{etoolbox}
\usepackage{tikz}

\usepackage{url}

\usepackage[capitalize,noabbrev]{cleveref}

\newcommand{\mypara}[1]{\vspace{0.1cm}\noindent\textbf{#1}}

\theoremstyle{plain}

\newtheorem{proposition}{Proposition}
\newtheorem{lemma}{Lemma}

\theoremstyle{definition}

\newtheorem{assumption}{Assumption}
\theoremstyle{remark}

\title{\LARGE \bf
Safety-Constrained Policy Transfer with Successor Features
}

\author{Zeyu Feng$^{1}$, Bowen Zhang$^{1}$, Jianxin Bi$^{1}$, and  Harold Soh$^{1,2}$
\thanks{$^{1}$Dept. of Computer Science, National University of Singapore \texttt{\{zeyu, bowenzhang, jianxin.bi, harold\}@comp.nus.edu.sg}. $^{2}$Smart Systems Institute (SSI), NUS.}
}

\begin{document}

\maketitle
\thispagestyle{empty}
\pagestyle{empty}

\begin{abstract}
    In this work, we focus on the problem of safe policy transfer in reinforcement learning: we seek to leverage existing policies when learning a new task with specified constraints. This problem is important for safety-critical applications where interactions are costly and unconstrained policies can lead to undesirable or dangerous outcomes, e.g., with physical robots that interact with humans. We propose a Constrained Markov Decision Process (CMDP) formulation that simultaneously enables the transfer of policies and adherence to safety constraints. Our formulation cleanly separates task goals from safety considerations and permits the specification of a wide variety of constraints. Our approach relies on a novel extension of generalized policy improvement to constrained settings via a Lagrangian formulation. We devise a dual optimization algorithm that estimates the optimal dual variable of a target task, thus enabling safe transfer of policies derived from successor features learned on source tasks. Our experiments in simulated domains show that our approach is effective; it visits unsafe states less frequently and outperforms alternative state-of-the-art methods when taking safety constraints into account.
\end{abstract}
\input{sections/intro.tex}
\input{sections/background.tex}
\input{sections/methods.tex}
\input{sections/relatedwork.tex}
\input{sections/results.tex}
\input{sections/conclusion.tex}


\clearpage
\clearpage
\bibliography{reference}
\bibliographystyle{IEEEtran}

\appendices
\input{sections/appendix.tex}

\end{document}

%% file: sections/intro.tex
\section{Introduction} \label{sec:1}

Reinforcement learning (RL) is an attractive paradigm for developing agents that learn through interactions with an environment---the framework is not only elegant and principled, but also supported by recent empirical evidence showing RL agents can achieve superhuman performance on complex tasks~\cite{mnih2015human}. Nevertheless, RL remains challenging to employ in environments (especially real-world settings) where interactions are costly and safety is a principal concern. The cause for this can be traced back to two well-known deficiencies of canonical RL methods: the high sample complexity associated with learning and the risks associated with unconstrained exploration. Recognition of this problem has given rise to a body of work on Safe-RL~\cite{Garcia2015}, which attempt to maximize expected returns while ensuring thresholds on system performance and safety are met. 

In this paper, we build upon this line of work and investigate how \emph{safety-constrained} policies can be \emph{transferred} across tasks with different objectives. Specifically, we develop transfer learning for Constrained Markov Decision Processes (CMDPs)~\cite{altman1999constrained}. Our approach is motivated by the advantages afforded by the constrained criterion compared to other safety formulations: CMDPs are well-studied models that cleanly separate task goals from safety considerations. Similar to very recent work on risk-aware policy transfer~\cite{gimelfarb2021riskaware}, we employ the powerful successor features (SF)~\cite{NIPS2017_350db081} framework as our main mechanism for transfer. But unlike \cite{gimelfarb2021riskaware}, our setup naturally encompasses cases where risk cannot be captured by the variance of the return and enables the specification of a wide variety of safety considerations. In essence, transfer in CMDPs promotes safe behavior via both sample reduction and constraint adherence.

Our safe-policy transfer framework rests upon a Lagrangian formulation of CMDP with successor features. In contrast to the unconstrained and risk-aware settings, our error analysis reveals the importance of the dual variable to induce safe policy transfer. With this insight, we introduce a dual optimization method for approximating the optimal Lagrangian multiplier, along with a proof of convergence. Taken together, our theoretical findings and algorithmic development provides a principled approach for safe-policy transfer. 

Our framework---which we call the \textbf{S}uccessor \textbf{F}eature \textbf{T}ransfer for \textbf{Co}nstrained \textbf{P}olicies (\method)---enables agents to exploit structure in both the task objectives and constraints to more quickly learn safe and optimal behavior on new tasks. Experiments on three benchmark problems~\cite{gimelfarb2021riskaware,NIPS2017_350db081}---a discrete gridworld environment and two continuous physical robot simulations---show \method{} is able to achieve positive transfer in terms of both task objectives and safety considerations. In particular, we see that \method{} compares favorably to safety-agnostic transfer and a state-of-the-art risk-sensitive method. In summary, this paper makes three key contributions:
\begin{itemize}
	\item A generalized policy improvement theorem for constrained policy transfer via successor features;
 	\item A dual optimization algorithm (with a proof of convergence) that enables safe transfer of source policies to a new target task; 
 	\item Experimental results on benchmark RL problems that validate the effectiveness of transfer learning with safety constraints.
 \end{itemize}

%% file: sections/background.tex
\section{Preliminaries} 
In this section, we give a succinct review of background material, specifically Markov Decision Processes (MDPs) and Constrained MDPs (CMDPs). For more information, please refer to excellent survey articles~\cite{1994MDP,altman1999constrained}. 

\mypara{Markov Decision Process (MDP)}. We begin with the popular MDP framework~\cite{1994MDP} for modeling sequential decision-making problems. Formally, an MDP is denoted as a tuple $M = \left(\mathcal{S}, \mathcal{A}, p, r, \gamma\right)$, where $\mathcal{S}$ is the state space, $\mathcal{A}$ is an action set, $p$ is a transition probability function, $r$ is a reward function and $\gamma$ is a discount factor. After an agent executes an action $a$ in state $s$, it transitions to a new state $s'$ and receives a reward. This transition is governed by    $p:\mathcal{S}\times\mathcal{A}\times\mathcal{S}\rightarrow\left[0,\infty\right]$, which models the distribution of the next state $p\left(s' \vert s,a\right)$.  The function $r\left(s,a,s^{\prime}\right)$ provides the reward, which we assume to be bounded. An agent's policy $\pi\in\Pi$ dictates its action given a particular state and we focus on stochastic policies where $\pi(s)$ is $p(\cdot|s)$. 

Given an MDP, we seek the optimal policy $\pi^*$ that maximizes the expected discounted sum of rewards or value function $\max_{\pi\in\Pi} V_r^{\pi}\left(s_0\right) = \mathbb{E}_{\pi} \left[R_r\vert s_0\right]$, where $R_r = \sum_{t=0}^{\infty}\gamma^tr\left(s_t,a_t,s_{t+1}\right)$ is the reward return for a specific trajectory. The expectation above is taken with respect to trajectories generated by applying the policy $\pi$ in the environment starting from an initial state $s_0$. To ease our exposition, we assume the initial state to be fixed, but our work readily extends to the more general case where $s_0$ is drawn from a prior distribution of starting states. The value function $V_r^{\pi}$ is related to the \emph{action-value} function $Q_r^{\pi}$, where $V_r^{\pi}\left(s_0\right)=\mathbb{E}_{a\sim\pi\left(s_0\right)} Q_r^{\pi}\left(s_0,a\right)$ and $Q_r^{\pi}\left(s_0,a\right)=\mathbb{E}_{\pi} \left[\sum_{t=0}^{\infty}\gamma^t r\left(s_t,a_t,s_{t+1}\right)\vert s_0, a_0=a\right]$.

Given the action-value function $Q_r^{\pi}$ of a particular policy $\pi$, Bellman's celebrated \emph{policy improvement theorem}~\cite{sutton2018reinforcement} provides a means of obtaining a policy $\pi'$ that is guaranteed to be no worse than $\pi$ and possibly better. The  policy $\pi^{\prime}$ is obtained by acting greedily with respect to $Q_r^{\pi}$, \ie, $\pi^{\prime}\left(s\right)\in\argmax_a Q_r^{\pi}\left(s,a\right)$. Under certain conditions, successive evaluation of $Q_r^{\pi}$ and improvement leads to an optimal policy $\pi^*$ --- this forms the basis of RL methods that employ dynamic programming. 

\mypara{Constrained MDP (CMDP)}. In many decision-making scenarios, constraints play an important role, \eg, to account for safety or ethical considerations. To model such problems, we can extend the MDP setup above to include a constraint return $R_c = \sum_{t=0}^{\infty}\gamma^tc\left(s_t,a_t,s_{t+1}\right)$ of some utility function values $c\left(s,a,s^{\prime}\right)\in\left[0,1\right]$. For example, the utility function $c$ can be an indicator function $c\left(s,a,s^{\prime}\right)=\mathds{1}(s\in S)$ of whether an agent is in a safe set $S$~\cite{paternain2022safe}. 

The CMDP $M^c = (\mathcal{S}, \mathcal{A}, p, r, c,\allowbreak \gamma)$~\cite{altman1999constrained}, has corresponding value $V_c^{\pi}\left(s_0\right)=\mathbb{E}_{a\sim\pi\left(s_0\right)}\left[ Q_c^{\pi}\left(s_0,a\right)\right]$ and action-value functions $Q_c^{\pi}\left(s_0,a\right)=\mathbb{E}_{\pi} \left[\sum_{t=0}^{\infty}\gamma^t c\left(s_t,a_t,s_{t+1}\right)\vert s_0, a_0=a\right]$ associated with the constraint function $c$. Our goal is to obtain an optimal policy that maximizes the expected reward return while ensuring the expected \emph{utility return} is met, 
\begin{equation}
	\max_{\pi\in\Pi} \quad V_r^{\pi}\left(s_0\right) \quad\quad\quad \text{s.t.} \quad V_c^{\pi}\left(s_0\right) \geq \tau.
\end{equation}
Here, $\tau$ is the threshold we seek to enforce for the constraint\footnote{We set $\tau\in\left(0,\nicefrac{1}{1-\gamma}\right]$ to avoid trivial solutions.}. Prior work has shown that constraining the utility value function as above is a reasonable approach for safety~\cite{Ray2019}; returning to our indicator function example above, setting $V_c^{\pi}\left(s_0\right) = \mathbb{E}_{\pi}\left[\sum_{t=0}^{\infty}\gamma^t \mathds{1}\left(s_t\in S\right)\vert s_0\right]$ provides a guarantee on the probability of the agent being in the safe set $S$ when following the policy $\pi$~\cite{paternain2022safe}. Note that we can easily accommodate (i) \emph{cost} constraints $\tilde{c}$ via a simple transformation to utilities $c\left(s_t,a_t,s_{t+1}\right) = 1 - \tilde{c}\left(s_t,a_t,s_{t+1}\right)$ and (ii) multiple constraints by specifying a set of thresholds $\{\tau^k\}_{k=1}^K$ for $K$ expected utility returns $\{V_{c^k}^{\pi}\left(s_0\right)\}_{k=1}^K$.

%% file: sections/methods.tex
\section{Policy transfer for CMDPs}

In this work, we aim to transfer a set of policies, specifically $Q$ functions  learned on a set of source CMDP tasks, to a new target task. In the RL setting, the agent only observes scalar rewards/utilities when interacting with the environment, i.e., the agent does not know the reward and constraint functions and has to quickly learn them.  

Our approach is based on the successor features (SFs) framework~\cite{NIPS2017_350db081} where the source and target tasks occur within the same environment (with the same transition function), but have different reward functions. Applied to the standard MDP framework, the set of tasks can be expressed as
\begin{equation}
	\label{eq:sf_general}
	\begin{split}
	    \mathcal{M}_\phi \left(\mathcal{S}, \mathcal{A}, p, \gamma\right) = \\ \{ M_i(\mathcal{S}, \mathcal{A}, p, & r_i, \gamma) \vert r_i\left(s,a,s^{\prime}\right)= 
	     \phi\left(s,a,s^{\prime}\right)^{\top}w_{r,i}\},
	\end{split}
\end{equation}
where each task $M_i\in\mathcal{M}_\phi$ is associated with a specific reward function $r_i$~\cite{pmlr-v80-barreto18a}. A key assumption is that the reward $r_i$ can be computed using a shared feature function $\phi\left(s,a,s^{\prime}\right)\in\mathbb{R}^d$ and a unique weight vector $w_{r,i}\in\mathbb{R}^d$~\cite{NIPS2017_350db081}.

The advantage of using the model~(\ref{eq:sf_general}) is that the action-value function is given by $Q_{r,i}^{\pi_i}\left(s,a\right) = \psi^{\pi_i}\left(s,a\right)^{\top} w_{r,i}$ where $\psi^{\pi_i}$ are the \emph{successor features}. In other words, we can quickly evaluate a policy $\pi_i$ on a new task $M_j$ by computing a dot-product. The successor features $\psi^{\pi_i}\left(s,a\right)$ satisfy the Bellman equation and can be computed via dynamic programming, and $w_{r,j}$ can be directly plugged-in (if known) or learned using standard learning algorithms. This property is especially desirable for transfer learning where we are given $N$ source tasks $\left\{M_i\right\}_{i=1}^N$ from $\mathcal{M}_{\phi}$ with their corresponding action-value functions $\left\{Q_{r,i}^{\pi_i}\right\}_{i=1}^N$, and seek to rapidly learn a new task $M_{N+1}\in\mathcal{M}_{\phi}$. Under certain conditions, the generalized policy improvement (GPI) theorem guarantees that the policy $\pi(s)\in\argmax_a\max_i Q_{r,j}^{\pi_i}(s,a)$ is no worse than the individual source policies.

To account for safety, we extend the above SF framework to CMDPs, where the set of tasks,
\begin{equation}
	\label{eq:sf}
	\begin{split}
		& \mathcal{M}_{\phi}^c \left( \mathcal{S}, \mathcal{A}, p, \gamma\right) = \{  M^c\left(\mathcal{S}, \mathcal{A}, p, r_i, c_i, \gamma\right) \vert r_i\left(s,a,s^{\prime}\right) \\
		& \;\;\; = \phi\left(s,a,s^{\prime}\right)^{\top}w_{r,i}, c_i\left(s,a,s^{\prime}\right)=\phi\left(s,a,s^{\prime}\right)^{\top}w_{c,i} \},
	\end{split}
\end{equation}
take on an additional utility weight vector $w_{c,i}$ for each task. To simplify our exposition and without loss of generality, we use the same $\phi$ function for both $r$ and $c$\footnote{We can handle the more general case of different $\phi$ functions by concatenating their outputs into a single vector and expanding the corresponding weights, which conforms to Eq.~(\ref{eq:sf}).}. We focus on a single constraint but the extension to multiple constraints is straightforward. We highlight that unlike the standard MDP setting, the optimal policy for a CMDP depends on \emph{both} the reward function and constraints; a naive application of GPI does not guarantee that the transferred policy will adhere to the specified constraints. In our setup, we model the value functions of both the reward and utility functions using SFs, which enables efficient policy and constraint evaluation during transfer.

\subsection{General policy improvement in CMDP}

In the following analysis, we seek to better understand the conditions under which positive transfer can occur in CMDPs.
We begin by revisiting the optimal policy $\pi_j^*$ of a target CMDP $M_j^c$. The optimal value function is $V_{r,j}^{\pi_j^*}\left(s\right) = \max_{\pi\in\Pi} \left\{ V_{r,j}^{\pi}\left(s\right) \vert V_{c,j}^{\pi}\left(s\right) \geq \tau \right\}$, where we see that the class of policies is restricted to those whose expected utility returns satisfy the threshold constraints. The dual function is
\begin{equation}
	\label{eq:dual}
	d\left(\lambda_j\right) = \max_{\pi\in\Pi}L\left(\pi, \lambda_j\right),
\end{equation}
where $L\left(\pi, \lambda_j\right) = V_{r,j}^{\pi}\left(s\right) + \lambda_j\left( V_{c,j}^{\pi}\left(s\right) - \tau \right)$ is the Lagrangian and $\lambda_j\in\mathbb{R}_+$ is the Lagrange multiplier (or dual variable). The optimal dual variable minimizes the Lagrangian: $\lambda_j^* = \argmin_{\lambda_j\geq0} d\left(\lambda_j\right)$, and as we will see, plays an important role during transfer. 

Given the above, we extend GPI to the multi-task CMDP setting given in Eq.~(\ref{eq:sf}). In the following, we adopt the standard assumption that a feasible policy exists, \ie,
\begin{assumption}
	\label{th:slater}
	\textnormal{(Slater's condition).} There exists a policy $\bar{\pi}\in\Pi$ such that $V_{c,j}^{\bar{\pi}}(s) \geq \tau$,
\end{assumption}
\noindent and introduce our first key result:
\begin{proposition}
	\label{th:gpi}
	Let $Q_{r,j}^{\pi_i^*}$ and $Q_{c,j}^{\pi_i^*}$ be the action-value functions of an optimal policy of $M_i^c\in\mathcal{M}_{\phi}^c$ when executed in $M_j^c\in\mathcal{M}_{\phi}^c$. Given approximations $\left\{ \tilde{Q}_{r,j}^{\pi_i} \right\}_{i=1}^N$ and $\left\{ \tilde{Q}_{c,j}^{\pi_i} \right\}_{i=1}^N$ such that $\left\lvert Q_{r,j}^{\pi_i^*}\left(s,a\right) - \tilde{Q}_{r,j}^{\pi_i}\left(s,a\right)\right\rvert \leq \epsilon$
	and $\left\lvert Q_{c,j}^{\pi_i^*}\left(s,a\right) - \tilde{Q}_{c,j}^{\pi_i}\left(s,a\right)\right\rvert \leq \epsilon$, $\forall s, a$ and $i\in\left\{1,\dots,N\right\}$. Consider the policy 
	\begin{equation}
	    \label{eq:gpi}
	    \pi(s)\in\argmax_a\max_{i\in[N]}\left\{ \tilde{Q}_{r,j}^{\pi_i}(s,a) + \tilde{\lambda}_j \tilde{Q}_{c,j}^{\pi_i}(s,a)\right\},
	\end{equation}
	by specifying a $\tilde{\lambda}_j$ for the new task $M_j^c$. If Slater's condition~(\ref{th:slater}) holds, we have
	\begin{align}
		& Q_{j,\lambda_j^*}^{\pi_j^*}\left(s,a\right) - Q_{j,\tilde{\lambda}_j}^{\pi}\left(s,a\right) \nonumber \\
		\leq & \min_i \, \frac{2}{1-\gamma} \Big( \phi_{\max}\lVert w_{r,j}-w_{r,i} \rVert + \phi_{\max}\lVert \lambda_j^*w_{c,j}-\lambda_i^*w_{c,i} \rVert \nonumber \\
		& + \left\vert\lambda_j^*-\tilde{\lambda}_j\right\vert + \epsilon(1+\tilde{\lambda}_j) \Big) + 2\tau\left\vert\lambda_j^*-\lambda_i^*\right\vert,
	\end{align}
    where $Q_{j,\lambda}^{\pi}\left(s,a\right) \coloneqq Q_{j,r}^{\pi}\left(s,a\right) + \lambda\left(Q_{j,c}^{\pi}\left(s,a\right) - \tau\right)$ for any $\pi$, $j$ and $\lambda$, and $\phi_{\max}=\max_{s,a}\lVert\phi\left(s,a\right)\rVert$.
\end{proposition}
The detailed proofs can be found in Appendix~\ref{sec:app}. The proposition above upper-bounds the differences in action-values of the transfer policy in Eq. (\ref{eq:gpi}) and the optimal policies of the target task $M_j^c$. The bound formally captures the roles of multiple factors in safe \emph{transfer}. Similar to the GPI theorem with successor features, the first term $\lVert w_{r,j}-w_{r,i} \rVert$ characterizes the difference in the task objectives. However, our constrained version in Prop.~\ref{th:gpi} reveals the critical role of the optimal dual variable $\lambda_j^*$ and its estimation $\tilde{\lambda}_j$ in bounding the loss. We draw attention to the remaining terms in the bound; we see that the second and last elements characterize the similarity between tasks in terms of their utilities and optimal dual variables. This reflects that intuition that the performance of the agent depends on the similarity to source tasks in terms of the safety constraints. The term $\left\vert\lambda_j^*-\tilde{\lambda}_j\right\vert$ highlights the gap caused by estimating the optimal dual variable; the larger the estimation error, the larger the loss incurred by the agent.  

To shed additional light on the transfer policy, let us draw a connection to the optimal CMDP value function. We leverage the following lemma on strong duality\footnote{This lemma is also used in the proof of Proposition~\ref{th:gpi}.},
\begin{lemma}
    \label{th:duality}
	\textnormal{[Strong duality]~\cite{NEURIPS2019_c1aeb651}}. If Slater's condition~(\ref{th:slater}) holds, then strong duality holds $V_{r,j}^{\pi_j^*}\left(s\right) = d\left(\lambda_j^*\right)$.
\end{lemma}
Combining the above with the definition of the dual function in Eq.~(\ref{eq:dual}) yields
\begin{align}
	V_{r,j}^{\pi_j^*}\left(s\right) & = \max_{\pi\in\Pi} V_{r,j}^{\pi}\left(s\right) + \lambda_j^*\left( V_{c,j}^{\pi}\left(s\right) - \tau \right) \nonumber \\
	& = \max_{\pi\in\Pi,a=\pi(s)} Q_{r,j}^{\pi}\left(s,a\right) + \lambda_j^*\left( Q_{c,j}^{\pi}\left(s,a\right) - \tau \right) \label{eq:dualvq}
\end{align}
The second equality above holds because the optimal policies have corresponding optimal action-value functions, since the Lagrangian can be equivalently written as a value function $L\left(\pi,\lambda_j^*\right) = \mathbb{E}_{\pi} \left[ \sum_{t=0}^{\infty}\gamma^t r_j^{\lambda_j^*}\left(s_t,a_t,s_{t+1}\right) \vert s_0=s\right]$ with a single reward function $r_j^{\lambda}(s,a,s^{\prime}) = r_j\left(s,a,s^{\prime}\right) + \lambda\left( c_j\left(s,a,s^{\prime}\right) - \left(1-\gamma\right)\tau \right)$. Let us compare Eq.~(\ref{eq:dualvq}) against the policy in  Eq.~(\ref{eq:gpi}). We see that the value of the policy matches that of the optimal value function if (i) $\tilde{\lambda}_j = \lambda_j^*$, (ii) the true action-value functions of source policies are available, and (iii) the set of source policies $\{\pi_i^*\}_{i=1}^N$ equals policy space $\Pi$ (or at least covers the optimal region). While less formal than Prop.~\ref{th:gpi}, this analysis reveals the ingredients needed for strong positive transfer (and potential sources of policy under-performance), which again includes the optimal dual $\lambda_j^*$.

Taken together, both Prop.~\ref{th:gpi} and the above discussion indicate that a good estimate of the optimal dual variable is important for transfer in CMDPs. We could learn it by training a policy on the target task but this would typically entail many interactions with the environment, defeating the purpose of transfer. Instead, we propose a practical optimization method to approximate $\lambda_j^*$ \emph{without} having to resort to RL on the target task.

\subsection{Optimal dual estimation} \label{sec:dual_est}

Recall that the optimal dual variable for target $M_j^c$ is:
\begin{equation}
    \label{eq:opt_dual}
	\lambda_j^*\in \argmin_{\lambda_j\geq0}\max_{\pi\in\Pi} V_{r,j}^{\pi}\left(s\right) + \lambda_j\left( V_{c,j}^{\pi}\left(s\right) - \tau \right).
\end{equation}
Given the source tasks, we can approximate the optimal dual via maximizing among source policies,
\begin{equation}
	\label{eq:dual_approx}
	\hat{\lambda}_j\in \argmin_{\lambda_j\geq0}\max_{i\in[N]} V_{r,j}^{\tilde{\pi}_i}\left(s\right) + \lambda_j\left( V_{c,j}^{\tilde{\pi}_i}\left(s\right) - \tau \right),
\end{equation}
where $\tilde{\pi}_i$ is the source policy given by the source $Q$ functions ($\tilde{Q}_{r,i}^{\pi_i}$ and $\tilde{Q}_{c,i}^{\pi_i}$) and dual variable $\tilde{\lambda}_i$. However, strong duality does not hold for Eq.~(\ref{eq:dual_approx}) because the domain of the primal problem is non-convex. As a workaround, we relax the domain of the primal problem (\ie, the policy space) to allow stochastic combination of policies: $\Pi_c := \texttt{Conv}\left(\{\tilde{\pi}_i\}_{i=1}^N\right)$. A policy $\pi_{\alpha}(a\vert s) = \frac{\sum_{i=1}^n\alpha_i\rho^{\tilde{\pi}_i}(s,a)}{\int_{\mathcal{A}}\sum_{i=1}^n\alpha_i\rho^{\tilde{\pi}_i}(s,a)\diff a} \in \Pi_c$ has value function $V_{r,j}^{\pi_{\alpha}}(s) = \sum_{i=1}^N\alpha_i V_{r,j}^{\tilde{\pi}_i}(s) = \sum_{i=1}^n\alpha_i\int_{\mathcal{S},\mathcal{A}}\rho^{\tilde{\pi}_i}(s,a)r_j(s,a)\diff s\diff a$, where $\sum_{i=1}^N\alpha_i=1$, $\alpha_i\geq0$ for any $i$ and $\rho^{\pi}(s,a)=\sum_{t=0}^{\infty}\gamma^t P(s_t=s,a_t=a\vert\pi)$ is occupation measure. Effectively, we expand the space of the finite number of policies from source tasks and assume the existence of a feasible policy in this space,
\begin{assumption}
	\label{th:slater_trans}
	\textnormal{(Slater's condition for policy transfer)}. There exists a policy $\bar{\pi}\in\Pi_c$ such that $V_{c,j}^{\bar{\pi}}(s) \geq \tau$.
\end{assumption}

Then, strong duality holds and we can use the dual variable $\tilde{\lambda}_j^{\alpha} \in \argmin_{\lambda_j\geq0}\max_{\pi_{\alpha}\in\Pi_c} V_{r,j}^{\pi_{\alpha}}\left(s\right) + \lambda_j\left( V_{c,j}^{\pi_{\alpha}}\left(s\right) - \tau \right)$ to approximate the optimal Lagrange multiplier. The value of $\tilde{\lambda}_j^{\alpha}$ can be obtained by optimization with alternative updates:
\begin{align}
	i^{(t+1)} & = \argmax_{i\in\left[N\right]} V_{r,j}^{\tilde{\pi}_i}\left(s\right) + \lambda^{(t)}\left( V_{c,j}^{\tilde{\pi}_i}\left(s\right) - \tau \right) \label{eq:up_1}\\
	\lambda^{(t+1)} & = \text{P}_{\mathbb{R}_{\geq0}} \left( \lambda^{(t)} - \eta^{(t)}\left( V_{r,j}^{\tilde{\pi}_{i^{\left(t+1\right)}}}(s)-\tau \right) \right), \label{eq:up_2}
\end{align}
where $\text{P}_{\mathbb{R}_{\geq0}}$ projects an input onto the non-negative orthant. When the step size $\eta^{(t)}$ is chosen properly, this update process converges to the optimal solution,
\begin{proposition}
    \label{th:convergence}
	If the sequence of step sizes $\left\{\eta^{(t)}\right\}_{t=1}^{\infty}$ satisfy $\eta^{(t)}\rightarrow0$, $\sum_{t=1}^{\infty}\eta^{(t)}=\infty$ and $\sum_{t=1}^{\infty}{\left[\eta^{(t)}\right]}^2\leq\infty$, then updates given in Eq.~(\ref{eq:up_1}) and~(\ref{eq:up_2}) will derive a dual variable that converges to the optimal dual variable,~\ie, $\lambda^{(t)}\rightarrow\tilde{\lambda}_j^{\alpha}$.
\end{proposition}

\begin{algorithm}[t]
	\caption{Dual variable update for SFs transfer}
	\label{algo:d_update}
	\begin{algorithmic}[1]
		\REQUIRE The number of iterations T, A sequence of step size $\eta^{(t)}$.
		\INPUT Value functions of a set of source policies evaluated on $M_j^c$: $\left\{ \hat{V}_{r,j}^{\tilde{\pi}_i}(s), \hat{V}_{c,j}^{\tilde{\pi}_i}(s) \right\}_{i=1}^N$.
		\OUTPUT Approximate optimal dual variable $\lambda^{(T)}$.
		\STATE \textbf{Initialization} $\lambda^{(1)}=0$, $t=1$
		\WHILE{$t \leq T$}
		\STATE $i^{(t+1)} = \argmax_{i\in\left[N\right]} \hat{V}_{r,j}^{\tilde{\pi}_i}\left(s\right) + \lambda^{(t)}\left( \hat{V}_{c,j}^{\tilde{\pi}_i}\left(s\right) - \tau \right)$
		\STATE $\lambda^{(t+1)} = \lambda^{(t)} - \eta^{(t)}\left( \hat{V}_{r,j}^{\tilde{\pi}_{i^{\left(t+1\right)}}}(s)-\tau \right)$
		\STATE Project the dual variable to non-negative orthant: $\lambda^{(t+1)}=\text{P}_{\mathbb{R}_{\geq0}}\left(\lambda^{\left(t+1\right)}\right)$.
		\STATE $t \leftarrow t + 1$
		\ENDWHILE
	\end{algorithmic}
\end{algorithm}

\mypara{Algorithm and practical use.} Our dual estimation method is summarized in Algorithm~\ref{algo:d_update}. The true value function of a source policy $\tilde{\pi}_i$ typically is not available in practice and has to be estimated, e.g.,  with a sample average by evaluating $\tilde{\pi}_i$'s SFs with multiple trajectories on the source task and using the target task's reward vectors. If the estimate $\hat{V}_{r,j}^{\tilde{\pi}_i}(s)=\nicefrac{1}{K_i}\sum_{k=1}^{K_i} R_{r,j}\left(s,\omega_{i,k}\right)$ is computed with a random sample of trajectories $\{\omega_{i,k}\}_{k=1}^{K_i}$ such that $\mathbb{E}\left[R_{r,j}\left(s,\omega_{i,k}\right)\right]=V_{r,j}^{\tilde{\pi}_i}(s)\,\forall k$ for both $r$ and $c$, then $\hat{\lambda}_j^{\alpha}$ can be shown to be a consistent estimator,
\begin{proposition}
	Denote by $\Lambda_j$ the set of all optimal dual variable $\tilde{\lambda}_j^{\alpha}$. Let the estimator $\hat{\lambda}_j^{\alpha} \in \argmin_{\lambda_j\geq 0} \max_{\pi_{\alpha}\in\Pi_c} \hat{V}_{r,j}^{\pi_{\alpha}}\left(s\right) + \lambda_j \left( \hat{V}_{c,j}^{\pi_{\alpha}}\left(s\right) - \tau \right)$. Then $\mathrm{dist}\left(\hat{\lambda}_j^{\alpha},\Lambda_j\right) \rightarrow 0$ with probability $1$ as the sample size $K_i$, $i=1,\cdots,N$, tends to infinity.
\end{proposition}

Once we have $\lambda^{(T)}$, we can select actions on the target task by $\pi(s)\in\argmax_a\max_{i\in[N]} \tilde{Q}_{r,j}^{\pi_i}(s,a) + \lambda^{(T)} \tilde{Q}_{c,j}^{\pi_i}(s,a)$, where the transfer policy is given according to Eq. (\ref{eq:gpi}). Note that the optimal dual variable defined in Eq.~(\ref{eq:opt_dual}) is for a specific state $s$. Therefore, dual estimation is repeatedly executed during roll-out on the target task. However, using Algorithm~\ref{algo:d_update} on every state is computationally inefficient if the action-value functions for adjacent states are similar. In such cases, dual variable estimation can be performed periodically, which we found to work well in our experiments.

%% file: sections/relatedwork.tex
\newlength\squareheight
\setlength\squareheight{5pt}
\newcommand\squareslash{\tikz{\draw (0,0) rectangle (\squareheight,\squareheight);
\draw(0,0.5 * \squareheight) -- (0.5 * \squareheight, \squareheight);
\draw(0.5 * \squareheight, 0) -- ( \squareheight, 0.5 * \squareheight);
\draw(0,0) -- (\squareheight,\squareheight)}}
\definecolor{gold}{rgb}{1.0, 0.84, 0.0}

\begin{figure*}[t]
    \centering
	\begin{subfigure}{.23\textwidth}
	    \centering
	    \includegraphics[width=0.8\textwidth]{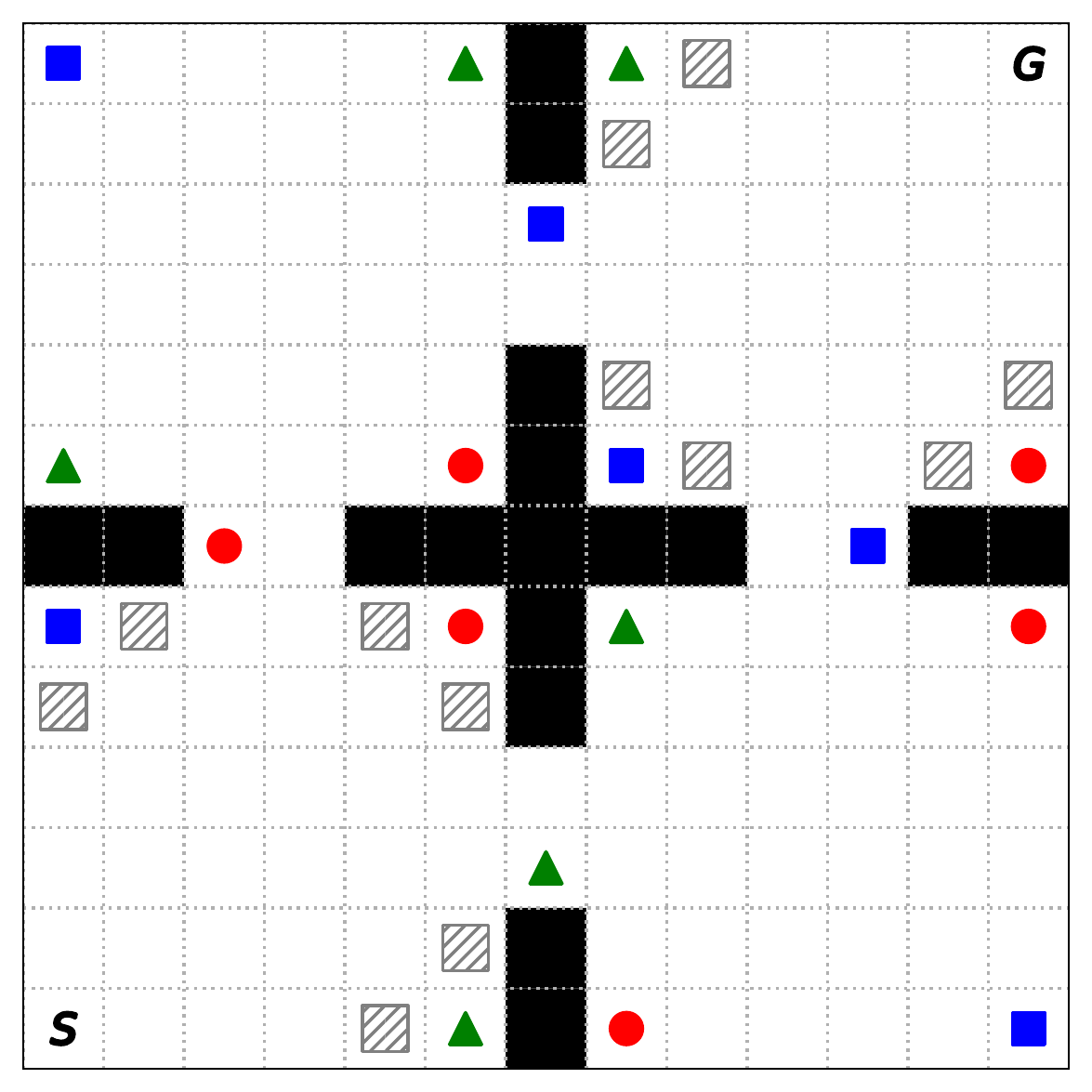}
    	\caption{Four-Room: `\textit{S}'/`\textit{G}' is the start/goal location. \textcolor{blue}{$\blacksquare$}, \textcolor{green}{$\blacktriangle$} and \textcolor{red}{$\CIRCLE$} are 3 types of objects with varying reward values on different tasks.~\squareslash~are traps.}
		\label{fig:fourroom}
	\end{subfigure}
	\hfill
	\begin{subfigure}{.43\textwidth}
	    \centering
	    \begin{minipage}{.42\textwidth}
	        \includegraphics[width=1.\textwidth]{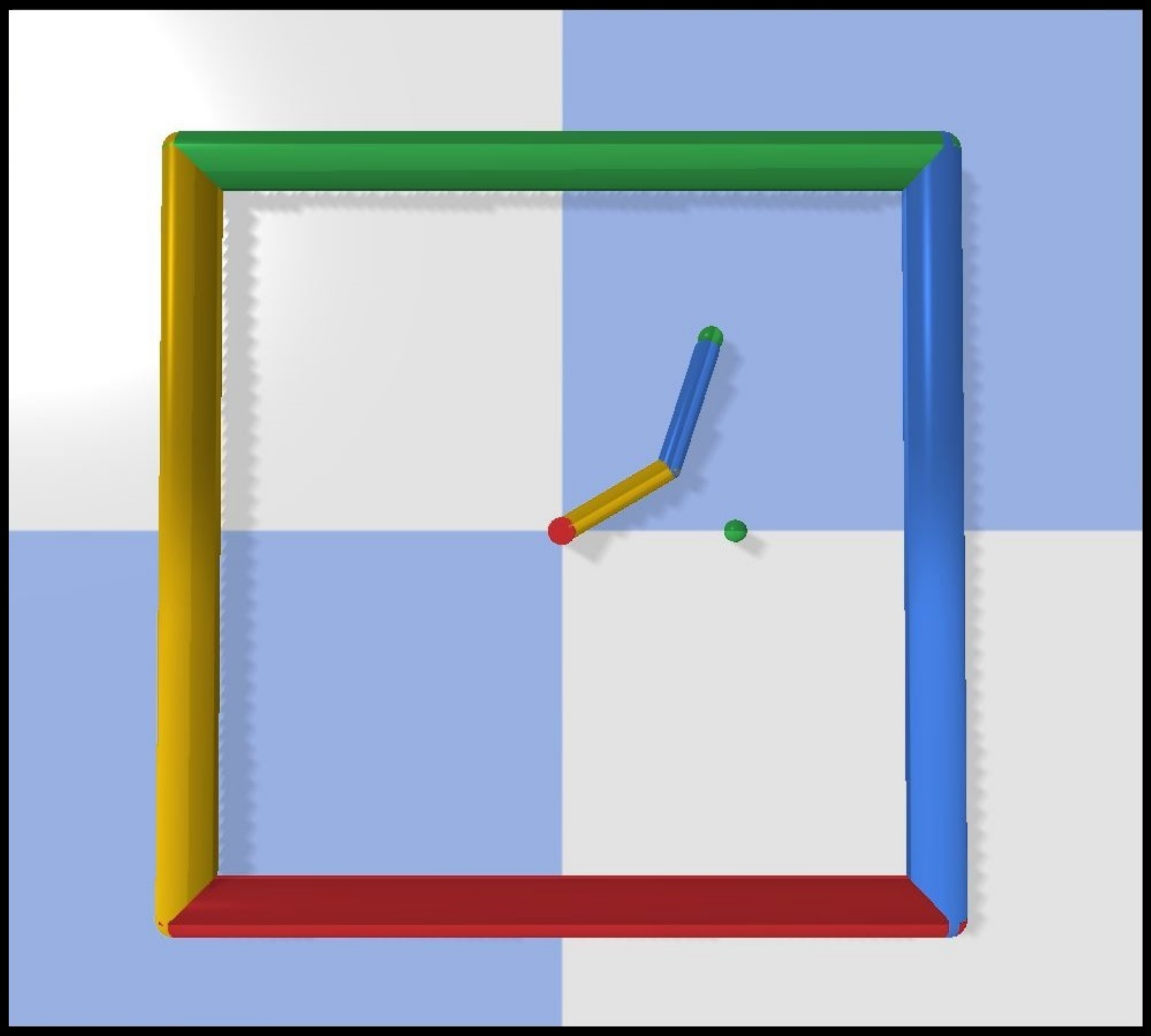}
		\end{minipage}
		\begin{minipage}{.4\textwidth}
		    \includegraphics[width=1.\textwidth]{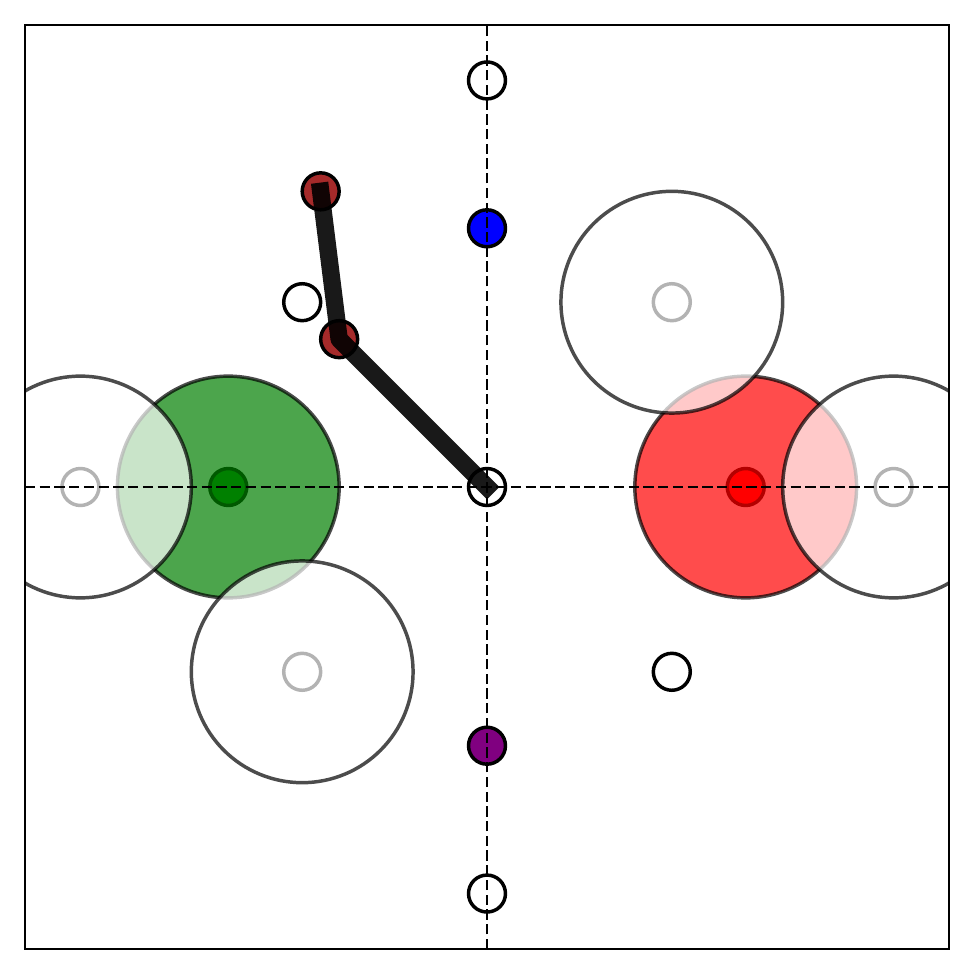}
		\end{minipage}
        \caption{Reacher. Left: a simulated two-link robotic arm based on MuJoCo physics engine. Right: four goal locations for training (colored filled) and eight for testing (white hollow) are represented by dots. Unsafe regions are plot in shaded circles.}
		\label{fig:reacher}
	\end{subfigure}
	\hfill
	\begin{subfigure}{.3\textwidth}
	    \centering
	    \includegraphics[width=0.8\textwidth]{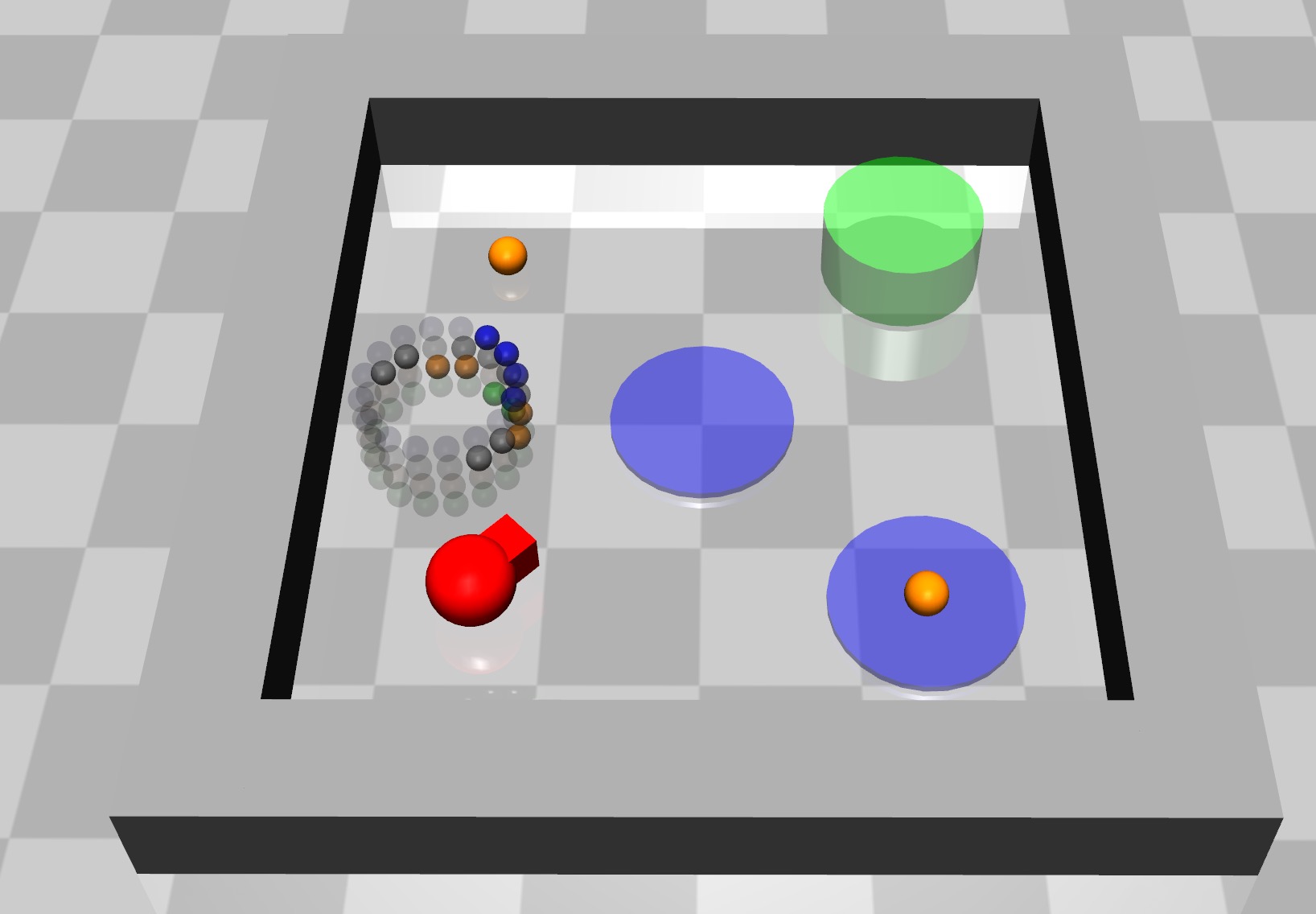}
    	\caption{SafetyGym: The red agent has to reach the goal (green cylinder) while avoiding the blue unsafe regions. Orange points are buttons with varying reward values on different tasks.}
		\label{fig:safetygym}
	\end{subfigure}
	\caption{\small Benchmark domains used in our experiments.}
	\label{fig:benchmarks}
\end{figure*}

\section{Related work} \label{sec:2}

Our work is related to existing research in transfer and safety in reinforcement learning. We give a brief overview of recent work and refer readers desiring more comprehensive information to excellent survey articles~\cite{JMLR:v10:taylor09a,zhu2020transfer,Garcia2015}. Safety is a consideration not restricted to RL; various methods have been proposed in the machine learning, control, and robotics communities~\cite{brunke2021safe}. For example, popular control approaches include the design of Lyapunov functions~\cite{NIPS2017_766ebcd5} and barrier functions~\cite{Cheng_Orosz_Murray_Burdick_2019}. However, these techniques often assume prior knowledge of the environment dynamics and thus, are unsuitable for complex RL tasks where a known model of the environment is unavailable. 

\mypara{Transfer learning in RL}. Knowledge transfer in RL can be achieved through various techniques, \eg, via instances~\cite{10.1145/1390156.1390225}, parameters~\cite{rajendran2015attend}, skill compositions for new tasks~\cite{pmlr-v139-vaezipoor21a,pmlr-v139-araki21a,NIPS2009_3eb71f62,NEURIPS2020_6ba3af5d}, and representations. Of particular relevance to our work are state representations that are informative of the reward~\cite{Konidaris2006AutonomousSK}. The successor features~\cite{NIPS2017_350db081} employed in our work is an example of such state representation transfer. Successor features (SFs) generalize the successor representation~\cite{DayanSR} and decouple environmental dynamics from rewards. This enables policy transfer across tasks with theoretical performance guarantees (provided by GPI). Recently, there has been significant work on extending SFs, \eg, to handle  environments with different dynamics~\cite{pmlr-v139-abdolshah21a} and exploit task descriptions~\cite{ma2020universal}. These advances, along with improvements in the construction of reward features~\cite{alver2021constructing}, can potentially be used to enhance \method's performance. 

\mypara{Safety in RL}. The term ``safety'' is overloaded in the field of RL and here, we focus on learning in a hazardous environment in an online fashion with transfer learning. One popular approach is to minimize risk during agent's interaction with the environment where the occurrence of rare dangerous events can cause irreparable harm to the agent and/or others. Methods that are designed to capture this risk include policy optimization over the worst case return~\cite{HEGER1994105}, variance of the return~\cite{10.2307/2629352} and the conditional value at risk~\cite{tamar2014policy}. However, since these techniques optimize a scalar return, practical usage typically entails careful design to trade-off reward and risk. 

An alternative approach---the one adopted in our work---is to model safety requirements as constraints. This results in a CMDP~\cite{altman1999constrained} for which various solvers have been developed. For example, with primal-dual optimization, \cite{BORKAR2005207} and \cite{tessler2018reward} incorporate actor-critic for policy learning. Other methods, such as CPO~\cite{achiam2017constrained}, PCPO~\cite{yang2020projectionbased} and~\cite{NEURIPS2018_4fe51490}, aim to achieve near-constraint satisfaction at each training iteration. CMDPs are also related to multi-objective learning, but emphasize avoidance of constraint violations~\cite{NEURIPS2019_873be070,gattami2019reinforcement}. In our work, we leverage the primal-dual approach for CMDPs, which enables SFs based policy transfer through the dual variable. For optimizing the policy, \method{} employs standard $Q$-learning; recent advances~\cite{pmlr-v97-le19a, NEURIPS2020_5f7695de, wei2021provably} are complementary and may further improve upon our results.

\mypara{Safety and transfer in RL.} Methods that combine both transfer and safety in RL have only been explored in very recent work. For example, recent work~\cite{Ethan2020} fine-tunes pre-trained policies on a CMDP using a combination of primal-dual optimization and Soft Actor-Critic \cite{haarnoja2018soft}. However, this method lacks a means of efficient policy evaluation and therefore, has difficulties scaling to many tasks. The closest related work to ours is RaSFQL~\cite{gimelfarb2021riskaware}, which employs transfer via successor feature for risk-sensitive RL. RaSFQL learns robust policies that avoid high variance behavior, but suffers from the limitations (mentioned above) associated with methods that incorporate a risk-sensitive criterion to the objective. Despite limitations, the overall direction of these prior works is promising---risk is minimized via both a safety mechanism and sample complexity reduction. \method{} is a step along this direction and the first to investigate efficient task transfer in the constrained setting.

%% file: sections/results.tex
\section{Experiments}

In this section we report on experiments designed to evaluate our primary hypothesis that \method{} achieves positive transfer in terms of task accomplishment and, more importantly, constraints. We also sought to empirically evaluate the importance of the dual-variable during transfer. 

\mypara{Environments.} We evaluated methods using three simulated environments (Fig. \ref{fig:benchmarks}). The first two (Four-Room and Reacher) are benchmarks used in prior work~\cite{NIPS2017_350db081,gimelfarb2021riskaware} to evaluate transfer learning in RL and safety.
This last domain is a customized SafetyGym~\cite{Ray2019} environment where a robot (red agent in Fig. \ref{fig:safetygym}) has to navigate to a goal location in a room. Please see the Appendix for implementation details.

\mypara{Compared methods.} We compared \method{} against state-of-the-art transfer learning baselines:
(1) Successor Feature Q-Learning (\sfql)~\cite{NIPS2017_350db081}, which performs transfer using GPI and SFs, but without considering constraints; the cost at each time step is added to the reward and hence, part of the return.  
(2) Risk-Aware SF Q-Learning (\rasfql)~\cite{gimelfarb2021riskaware}, a state-of-the-art method safe policy transfer method that employs a risk-sensitive criterion. 
We also compared safe RL method \cpo{}, and Primal-Dual Q-Learning (\pdql) which can be used to solve CMDPs with safety constraints. Comparing \method{} against \pdql{}/CPO enable us to ascertain the value of transfer learning, while performance differences from \sfql{} and \rasfql{} would be informative of any safety gains afforded by the constraint formulation. For the tasks where states and actions are continuous, we adopt the approach in prior work~\cite{gimelfarb2021riskaware} by employing Deep Q-Networks (DQNs)~\cite{NIPS2017_350db081,mnih2015human} and discretizing the action space. 

\subsection{Results and analysis}

\begin{figure}
    \centering
    \begin{minipage}[t]{.32\columnwidth}
    \hspace{0.6cm}\textbf{\small Four-Room}
    \end{minipage}
    \begin{minipage}[t]{.32\columnwidth}
    \hspace{0.7cm}\textbf{\small Reacher}
    \end{minipage}
    \begin{minipage}[t]{.32\columnwidth}
    \hspace{0.6cm}\textbf{\small SafetyGym}
    \end{minipage}
    
	\begin{subfigure}[t]{.32\columnwidth}
		\includegraphics[width=1\textwidth]{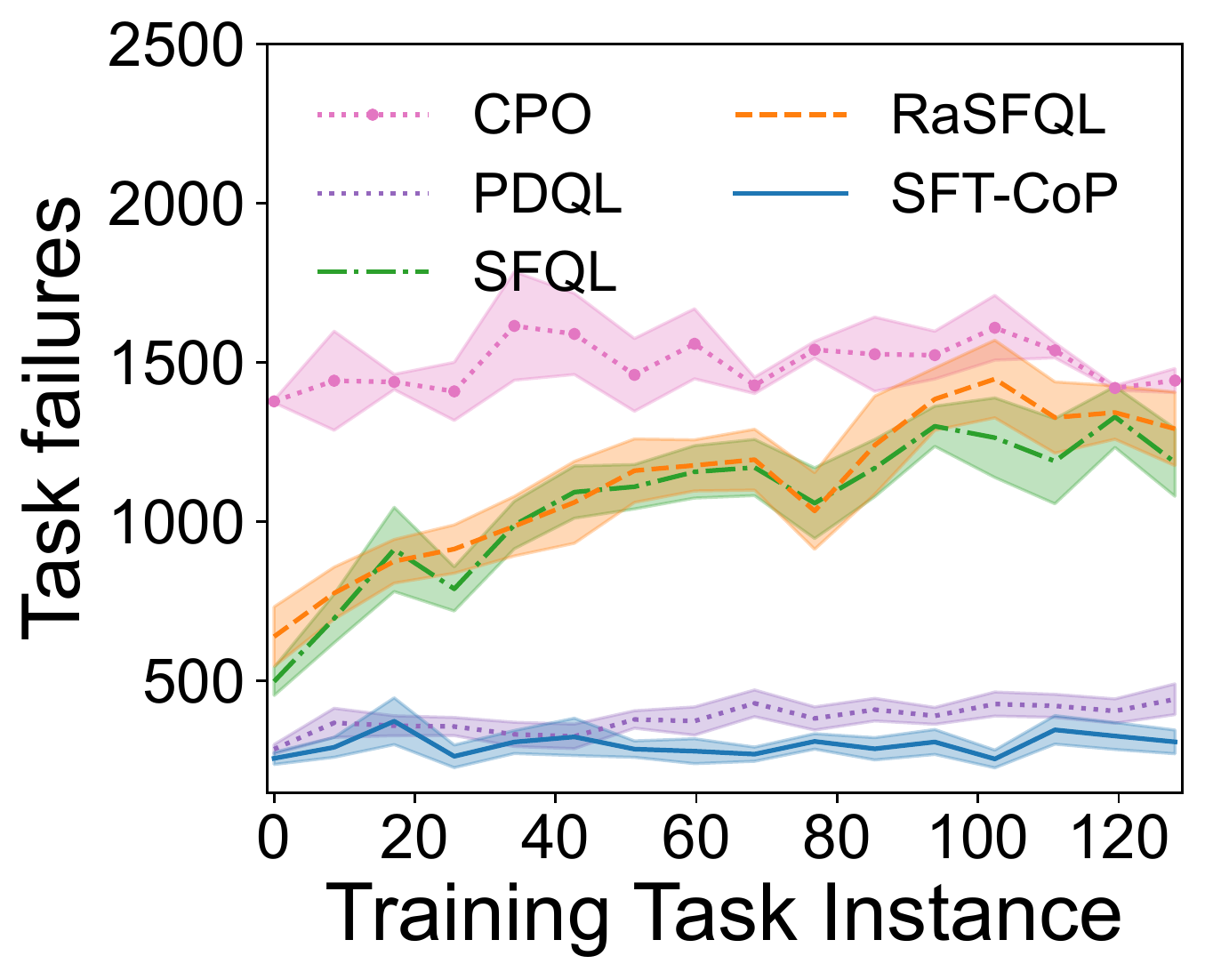}
		\label{fig:4room_failures}
	\end{subfigure}
	\begin{subfigure}[t]{.32\columnwidth}
	    \includegraphics[width=1\textwidth]{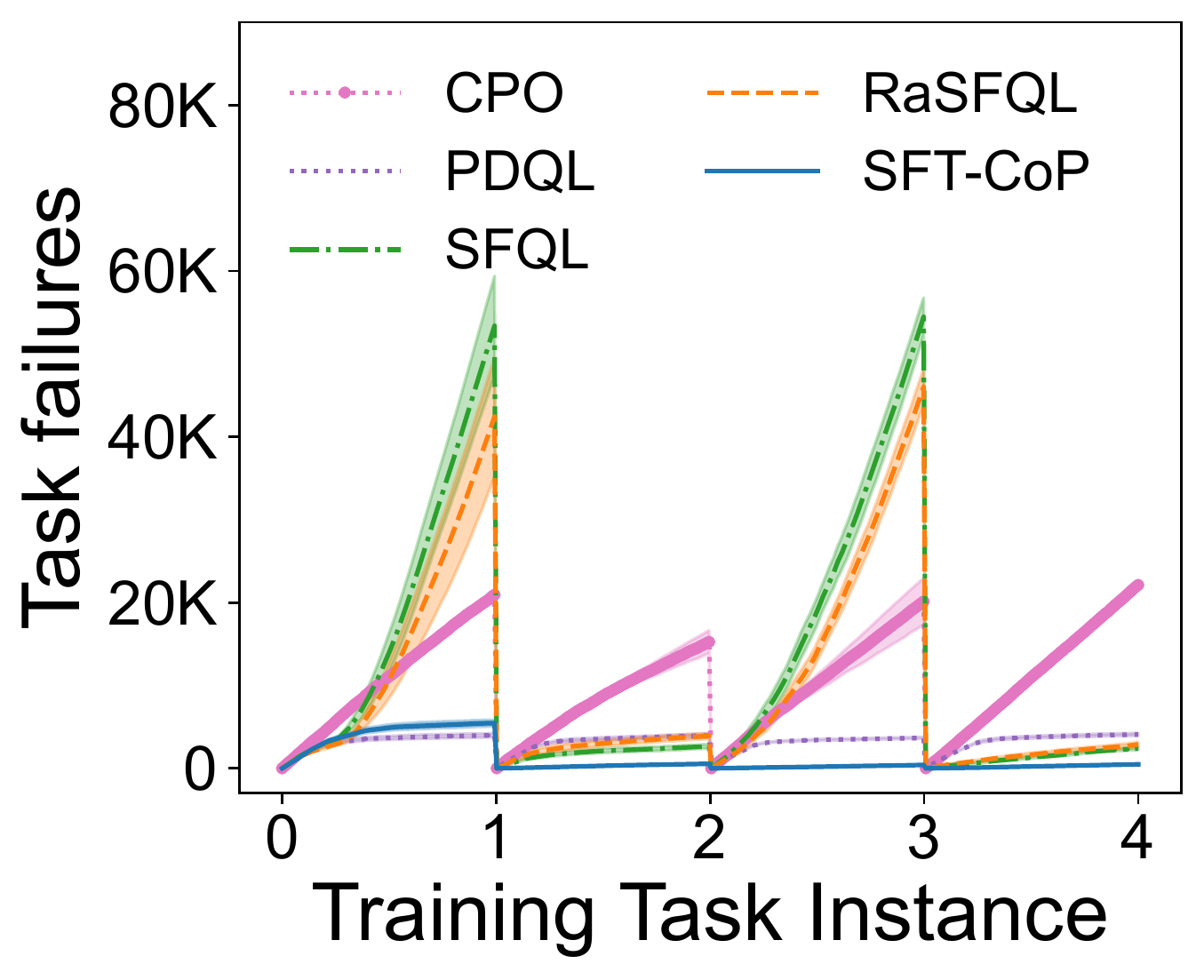}
		\label{fig:reacher_failures}
	\end{subfigure}
	\begin{subfigure}[t]{.32\columnwidth}
		\includegraphics[width=1\textwidth]{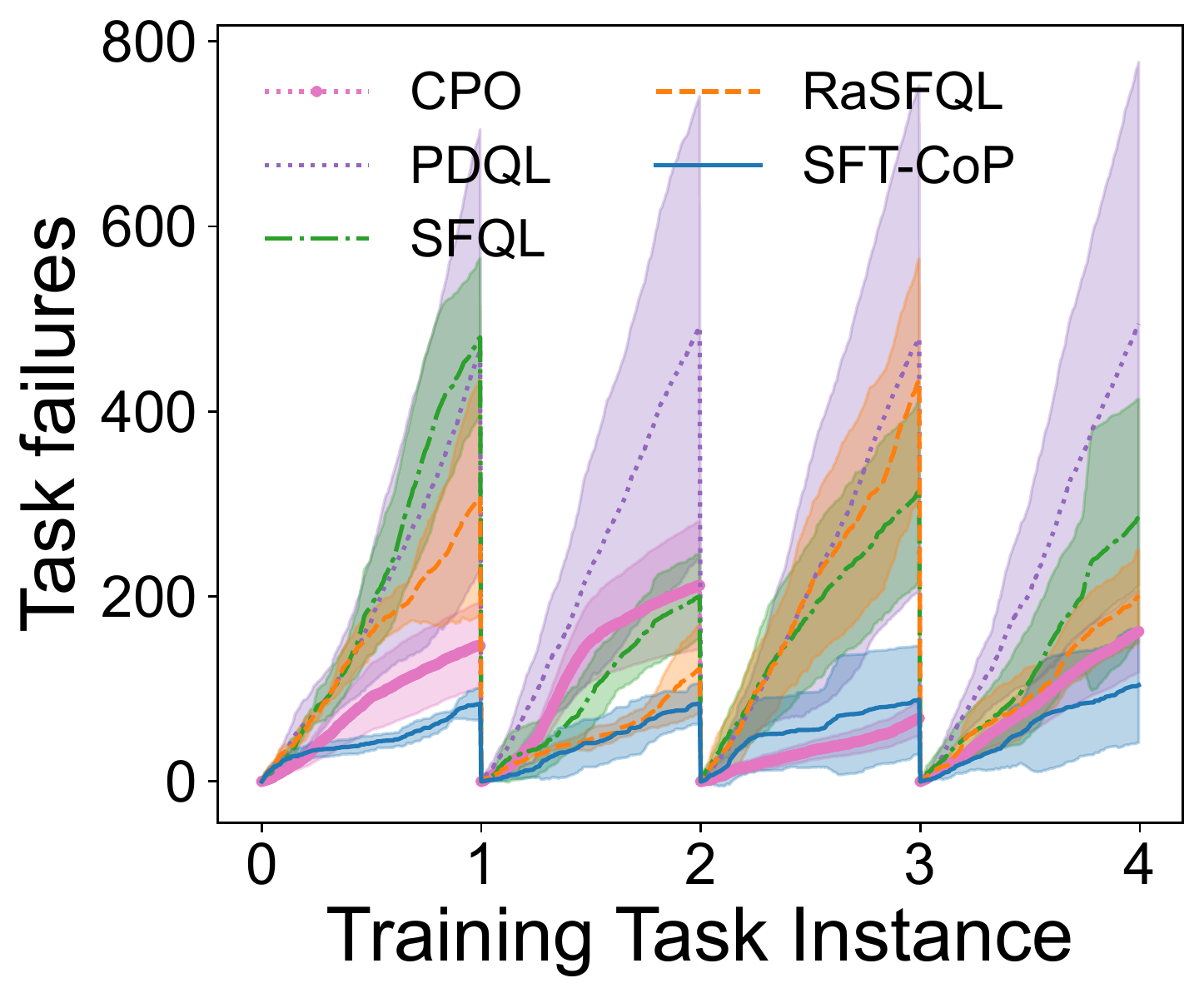}
		\label{fig:safetygym_failures}
	\end{subfigure}

	\begin{subfigure}[t]{.32\columnwidth}
		\includegraphics[width=1\textwidth]{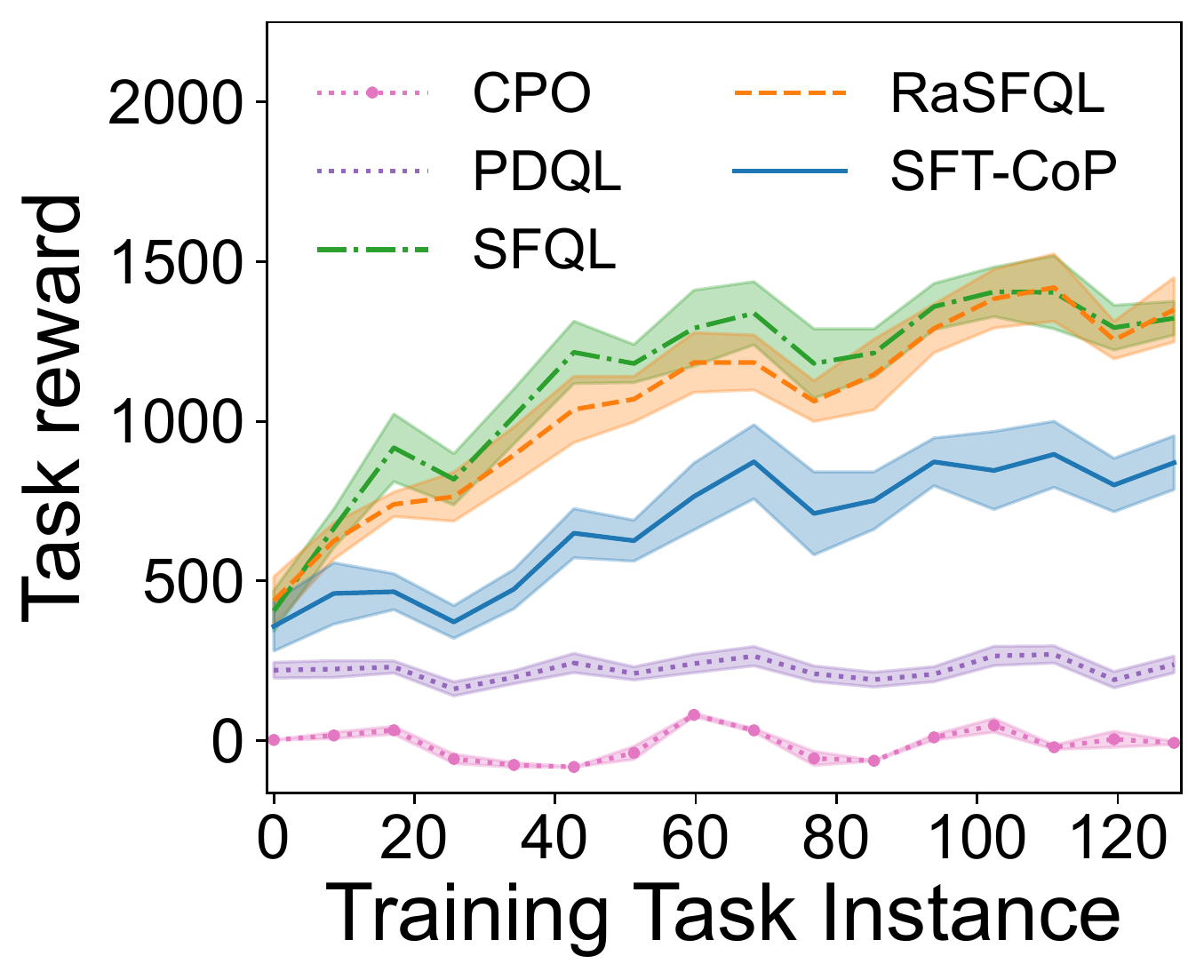}
		\label{fig:4room_rewards}
	\end{subfigure}
	\begin{subfigure}[t]{.32\columnwidth}
		\includegraphics[width=1\textwidth]{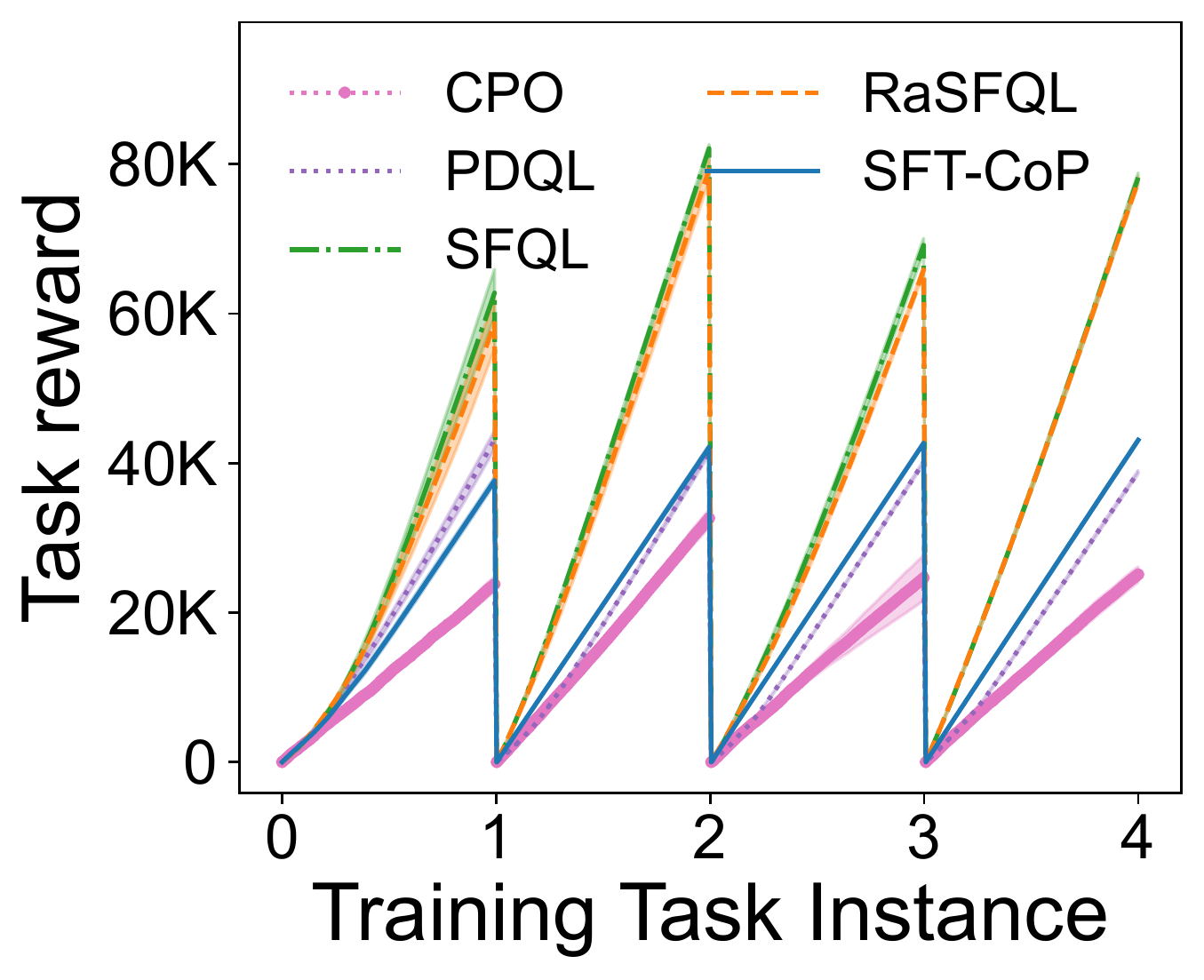}
		\label{fig:reacher_rewards}
	\end{subfigure}
	\begin{subfigure}[t]{.32\columnwidth}
		\includegraphics[width=1\textwidth]{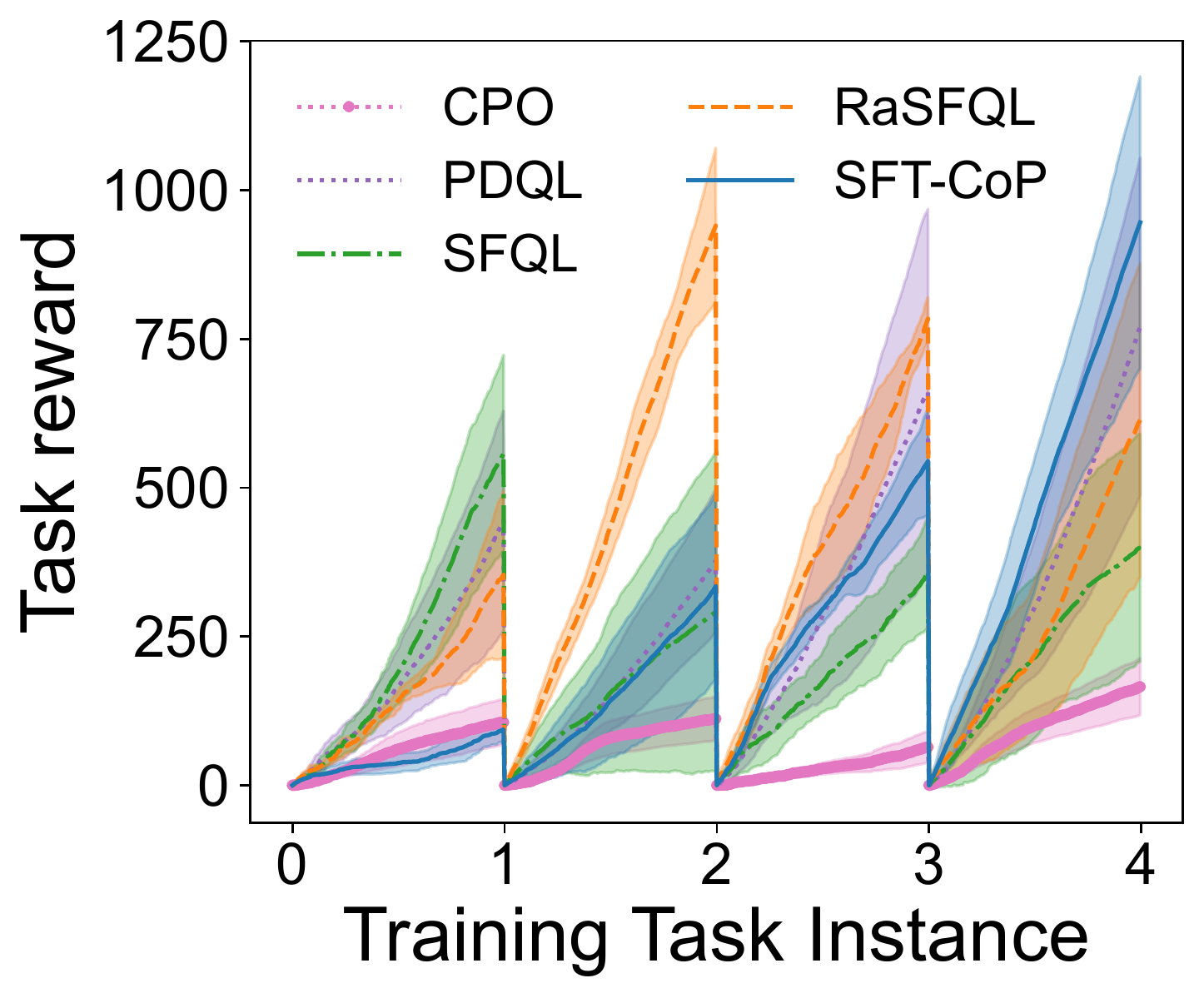}
		\label{fig:safetygym_rewards}
	\end{subfigure}
	\vspace{-1em}
	\caption{\small Performance of \cpo{}, \pdql{}, \sfql, \rasfql{} ($\beta=2$) and \method{} on the Four-Room, Reacher and SafetyGym domains. We compare task failures (top row) and rewards (bottom row) between the different transfer methods over the training task instances.}
	\label{fig:safe_transfer}
	\vspace{-1em}
\end{figure}

In the following section, we describe our key results, with extra results in the supplementary material. In general, we find that \method{} provides stricter adherence to safety constraints compared to the baselines. Our experiments further emphasize the importance of the dual variable $\lambda$ during transfer, which supports our theoretical findings. 

\mypara{Does \method{} achieve positive transfer (in terms of task objectives and constraints) in safety-constrained settings?} In brief, yes. Fig.~\ref{fig:safe_transfer} compares \method{} against non-transfer methods \cpo{} and \pdql{}, and \method{} significantly outperform them. Without transferring from previous tasks, these methods do not converge and perform poorly, especially in later tasks. In contrast, transfer methods such as \method{} leverage accumulated knowledge to achieve larger rewards and fewer failures in later tasks. 

\begin{wrapfigure}[13]{r}{0.22\textwidth}
    \centering
	\includegraphics[width=0.2\textwidth]{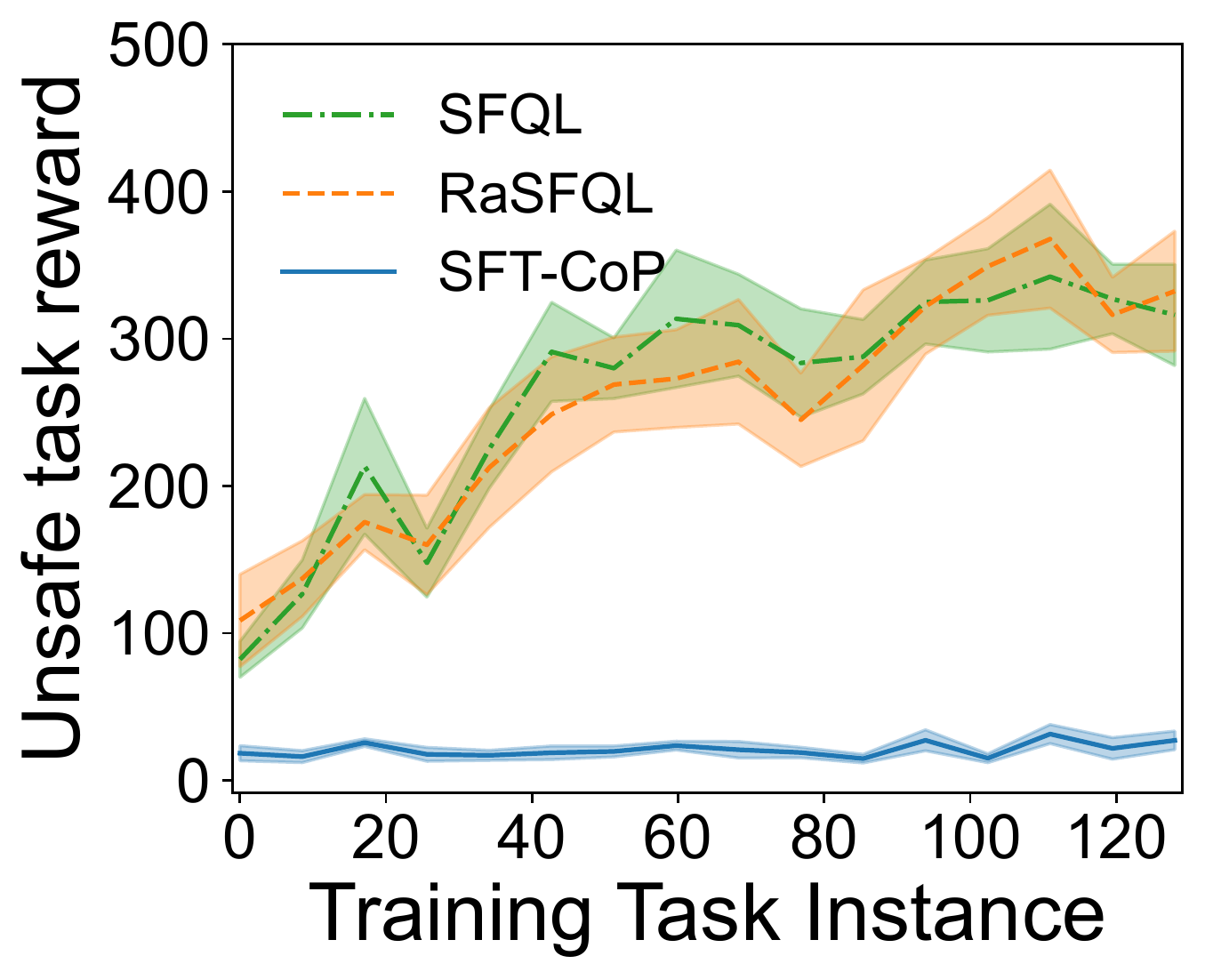}
	\caption{\small Unsafe rewards averaged over training tasks of \sfql, \rasfql{} ($\beta=2$) and \method{} on the Four-Room domain.}
	\label{fig:safe_unsafe_4room}
\end{wrapfigure}

\mypara{How does \method{} compare against transfer RL methods?} Fig. \ref{fig:safe_transfer} compares the \sfql{}, \rasfql{}, and \method{}. During learning, exploration is conducted so failures are not completely avoidable, even with transfer. All methods attain increasing rewards as more tasks are seen, but \method{} generally achieves the lowest failures.  Qualitatively, from Fig.~\ref{fig:safe_unsafe_4room} we observe that unlike \method{}, \sfql{} and \rasfql{} do not avoid the unsafe objects, e.g., in the bottom-left and top-right rooms in Foor-Rooms, or those within unsafe regions in Reacher/SafetyGym.

\begin{figure}[h]
    \begin{minipage}[t]{0.47\textwidth}
    \begin{center}
    \includegraphics[width=0.4\columnwidth]{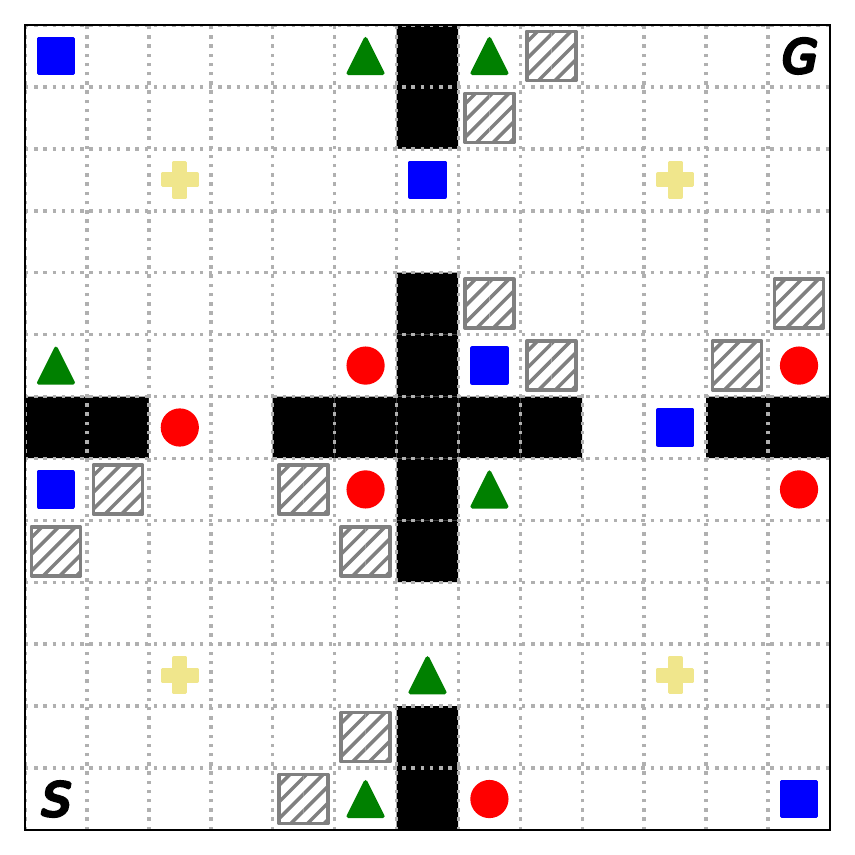}
    \includegraphics[width=0.49\columnwidth]{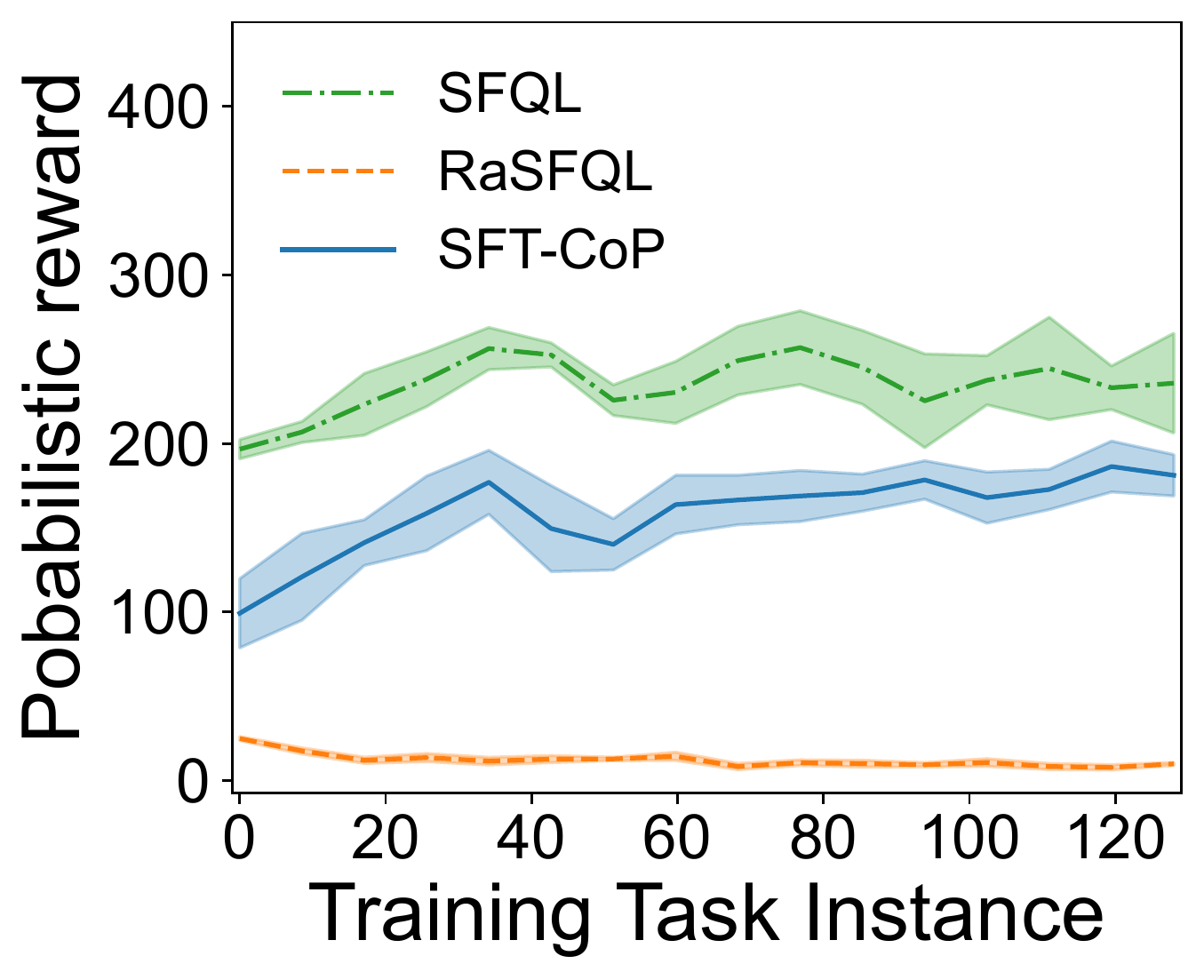}
    \end{center}
    \caption{\small Four-Room with probabilistic rewards from \textcolor{gold}{\ding{58}} which \rasfql{} does not collect.}
    \label{fig:probabilistic_reward}
    \end{minipage}
    \hfill
    \vspace{0.5em}
    \begin{minipage}[t]{0.49\textwidth}
    	\begin{minipage}[t]{.49\columnwidth}
	\includegraphics[width=0.9\columnwidth]{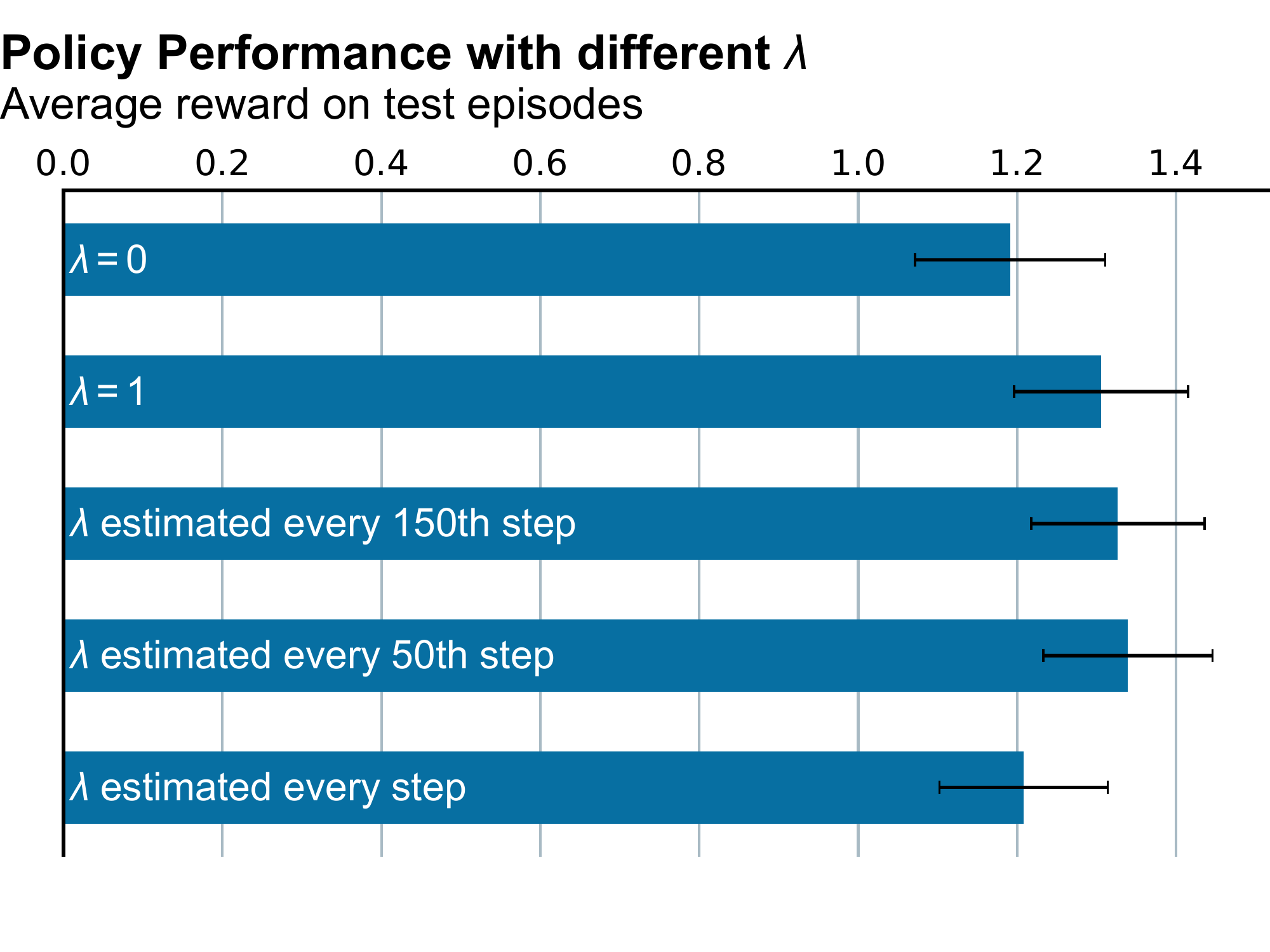}
    	\end{minipage}
    	\begin{minipage}[t]{.49\columnwidth}
    	\includegraphics[width=0.9\columnwidth]{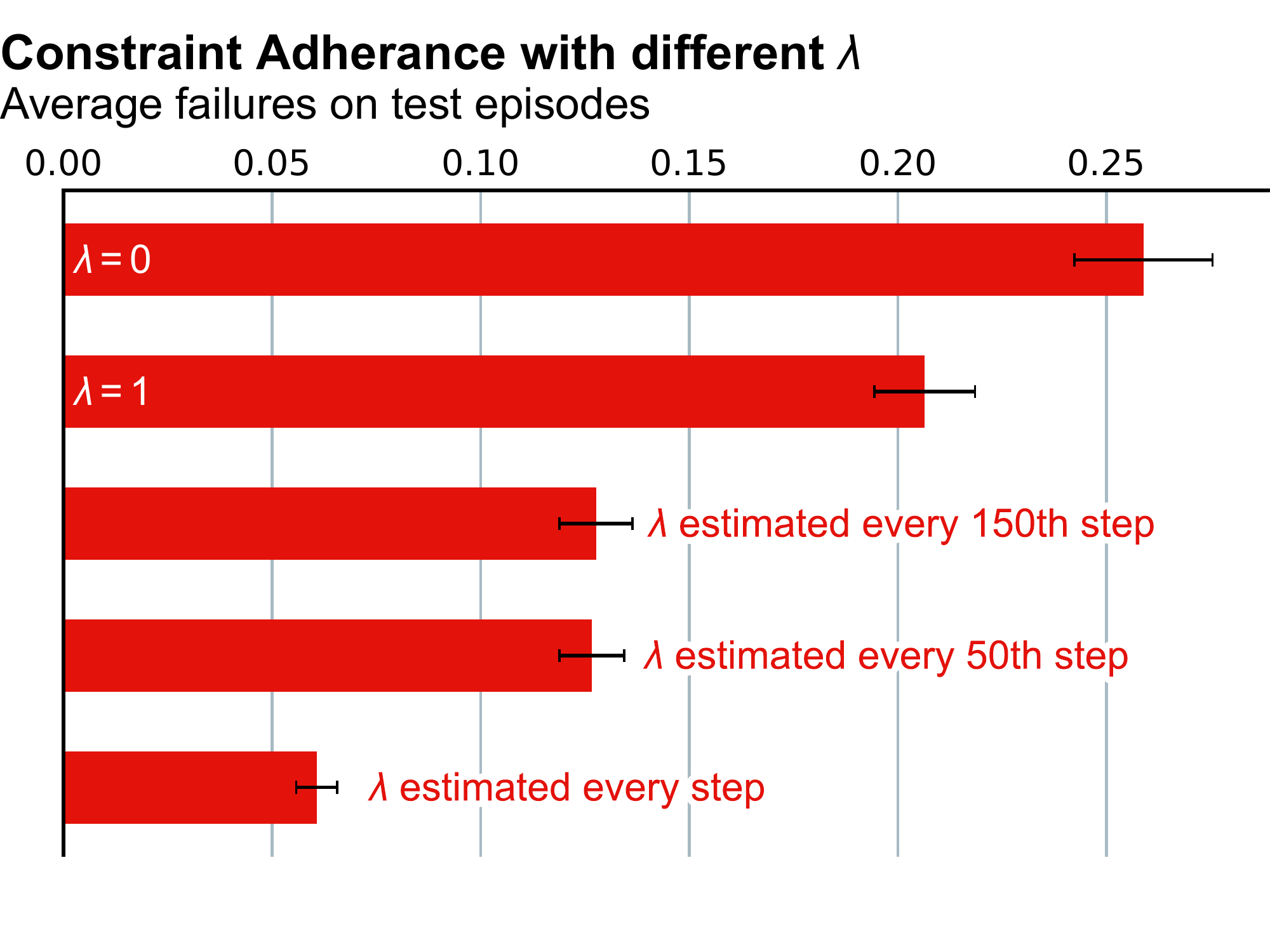}
    	\end{minipage}
    	\caption{\small Policy performance (left) and adherence to constraints (right) during transfer with different dual variables $\lambda$ in the Four-Room domain.}
    	\label{fig:lambda_var}
    \end{minipage}
\end{figure}

It should be noted that \method{} and \rasfql{} are motivated by different underlying principles---\method{} attempts to ensure compliance to constraints, while \rasfql{} seeks to reduce low-probability but high-cost failures. However, the standard reward variance formulation used in \rasfql{} tends to avoid not only uncertain cost, but also uncertain \emph{reward}; Fig. \ref{fig:probabilistic_reward} shows this phenomena empirically using the Four-Room domain modified with objects that provide positive reward with probability $5\%$. This problem does not occur with \method{}. Constraints can also be applied to avoid low probability unsafe events; when the unsafe regions have a low probability of failure ($5\%$ and $3.5\%$)~\cite{gimelfarb2021riskaware}, \method{} provides a similar level of safety as \rasfql{}.

\mypara{For \method, does accurate estimation of $\lambda$ matter in practice?} Our theoretical analysis suggests the importance of correctly estimating the optimal dual variable $\lambda$ when transferring to a new task. The bar charts in Fig.~\ref{fig:lambda_var} summarize the averaged rewards and failures of each episode with different methods of selecting $\tilde{\lambda}_j$. Setting $\tilde{\lambda}_j=0$ corresponds to the case where the cost is ignored, and $\tilde{\lambda}_j=1$ represents the setting where the cost is simply added to the reward. We also compare three different dual estimation update frequencies. As expected, the results show that estimating $\lambda$ via Alg. \ref{algo:d_update} resulted in fewer failures compared to fixed $\lambda = \{0,1\}$. Less frequent estimation led to more failures, but constraint adherence remains significantly better than the fixed $\lambda$ variants.

%% file: sections/conclusion.tex
\section{Conclusions and Future Work}

In this paper, we presented \method{}, a principled method for constrained policy transfer in reinforcement learning using successor features. Unlike prior work, our approach separates out safety considerations as constraints. To our knowledge, this is the first work showing efficient transfer learning of constrained RL policies. Moving forward, there is a rich literature on optimization and solvers for CMDPs and a study into which methods work best alongside successor features could lead to performance improvements. Moreover, the effectiveness of \method{} relies on useful successor features (that have to be learnt or provided) and is limited to a discrete action space; it remains an open question to learn a continuous action policy or a stochastic policy with a similar guarantee. Our bound in Prop. 1 brings new insights to the dual variable, but shares similar properties as in~\cite{NIPS2017_350db081} and~\cite{gimelfarb2021riskaware} in terms of approximation error and scaling. Improving this bound would be interesting future work.

%% file: sections/appendix.tex
\section{Proof of theoretical results} \label{sec:app}

\subsection{Proof of Proposition 1} \label{sec:app1}

Before the formal proof, we first introduce the following revised lemmas of GPI~\cite{pmlr-v80-barreto18a}.

\begin{lemma}
    \label{th:lemmaA1}
    \textnormal{(Generalized Policy Improvement.)} Let $\pi_1^*$, $\pi_2^*$, $\dots$, $\pi_N^*$ be $N$ optimal decision policies of $N$ tasks $\{M_i\in\mathcal{M_{\phi}}\}_{i=1}^N$, respectively, and let $\tilde{Q}^{\pi_1}$, $\tilde{Q}^{\pi_2}$, $\dots$, $\tilde{Q}^{\pi_N}$ be approximations of their respective action-value functions such that
    \begin{equation}
        \left\vert Q^{\pi_i^*}(s,a) - \tilde{Q}^{\pi_i}(s,a) \right\vert \leq \epsilon \,\:\text{for all}\,\: s\in\mathcal{S}, a\in\mathcal{A} \,\:\text{and}\,\: i\in[N].
    \end{equation}
    Define
    \begin{equation}
        \pi(s)\in\argmax_a\max_i \tilde{Q}^{\pi_i}(s,a).
    \end{equation}
    Then,
    \begin{equation}
        Q^{\pi}(s,a) \geq \max_i Q^{\pi_i^*}(s,a) - \frac{2}{1-\gamma}\epsilon.
    \end{equation}
\end{lemma}

\begin{lemma}
    \label{th:lemmaA2}
    Let $\delta_{ij}=\max_{s,a}\left\vert r_i(s,a,s^{\prime})-r_j(s,a,s^{\prime})\right\vert$. Then,
    \begin{equation}
        Q_i^{\pi_i^*}(s,a) - Q_i^{\pi_j^*}(s,a) \leq \frac{2\delta_{ij}}{1-\gamma}.
    \end{equation}
\end{lemma}

Now we are ready to prove Proposition~\ref{th:gpi}.

\begin{proof}
	Let $Q_{j,\lambda}^{\pi_i}\left(s,a\right) = Q_{j,r}^{\pi_i}\left(s,a\right) + \lambda\left(Q_{j,c}^{\pi_i}\left(s,a\right) - \tau\right)$ be the balanced action-value function of a parameter $\lambda$ and a policy $\pi_i$ on task $M_j^c\in\mathcal{M}_{\phi}^c$. As previously mentioned, it equals to the $Q$ function of policy $\pi_i$ with a reward function in the form of $r_j^{\lambda}(s,a,s^{\prime}) = r_j\left(s,a,s^{\prime}\right) + \lambda\left( c_j\left(s,a,s^{\prime}\right) - \left(1-\gamma\right)\tau \right)$. Then we can have
	\begin{align}
        & Q_{j,\lambda_j^*}^{\pi_j^*}\left(s,a\right) - Q_{j,\tilde{\lambda}_j}^{\pi}\left(s,a\right) \allowdisplaybreaks \\
        \leq \; & Q_{j,\lambda_j^*}^{\pi_j^*}\left(s,a\right) - Q_{j,\tilde{\lambda}_j}^{\pi_i^*}\left(s,a\right) + \frac{2\epsilon}{1-\gamma}\left(1+\tilde{\lambda}_j\right) \label{eq:p1} \allowdisplaybreaks \\
        = \; & Q_{j,\lambda_j^*}^{\pi_j^*}\left(s,a\right) - Q_{j,\lambda_j^*}^{\pi_i^*}\left(s,a\right) - \tilde{\lambda}_j\left(Q_{j,c}^{\pi_i^*}\left(s,a\right)-\tau\right) \notag \allowdisplaybreaks \\
        & + \lambda_j^*\left(Q_{j,c}^{\pi_i^*}\left(s,a\right)-\tau\right)  + \frac{2\epsilon}{1-\gamma}\left(1+\tilde{\lambda}_j\right) \allowdisplaybreaks \\
        \leq \; & \frac{2}{1-\gamma}\max_{s,a} \left\vert r_j^{\lambda_j^*}\left(s,a,s^{\prime}\right)-r_i^{\lambda_i^*}\left(s,a,s^{\prime}\right)\right\vert \notag \allowdisplaybreaks \\
        & + \left\vert\lambda_j^*-\tilde{\lambda}_j\right\vert\frac{1}{1-\gamma} + \frac{2\epsilon}{1-\gamma}\left(1+\tilde{\lambda}_j\right) \label{eq:p2} \allowdisplaybreaks \\
        \leq \; & \frac{2}{1-\gamma} \Big( \max_{s,a} \big\vert \phi\left(s,a,s^{\prime}\right)^{\intercal}\left(w_{r,j}-w_{r,i}\right) \notag \allowdisplaybreaks \\
        & + \phi\left(s,a,s^{\prime}\right)^{\intercal}\left(\lambda_j^*w_{c,j}-\lambda_i^*w_{c,i}\right) \big\vert + \tau\left(1-\gamma\right)\left\vert\lambda_j^*-\lambda_i^*\right\vert \Big) \notag \allowdisplaybreaks \\
        & + \left\vert\lambda_j^*-\tilde{\lambda}_j\right\vert\frac{1}{1-\gamma} + \frac{2\epsilon}{1-\gamma}\left(1+\tilde{\lambda}_j\right) \label{eq:p3} \allowdisplaybreaks \\
        \leq \; & \frac{2}{1-\gamma} \Big(\phi_{\max}\lVert w_{r,j}-w_{r,i} \rVert + \phi_{\max}\lVert \lambda_j^*w_{c,j}-\lambda_i^*w_{c,i}\rVert \notag \allowdisplaybreaks \\
        & + \left\vert\lambda_j^*-\tilde{\lambda}_j\right\vert + \epsilon\left(1+\tilde{\lambda}_j\right) \Big) + 2\tau\left\vert\lambda_j^*-\lambda_i^*\right\vert, \label{eq:p4}
	\end{align}
for any $i\in[N]$ and $\tilde{\lambda}_j>0$, where Eq.~(\ref{eq:p1}) is due to Lemma~\ref{th:lemmaA1} and Eq.~(\ref{eq:p2}) is due to applying Lemma~\ref{th:lemmaA2}, since there exists an optimal policy $\pi_i^*$ maximizing $Q_{i,\lambda_i^*}^{\pi_i}$ following the strong duality Lemma~\ref{th:duality}.

\end{proof}

\subsection{Proof of Proposition 2} \label{sec:app2}

We first restate a convergence lemma in sub-gradient optimization~\cite{anstreicher2009two}.
\begin{lemma}
    \label{th:sub_opt}
	Consider the optimization problem
	\begin{equation}
	\max_{x\in X} \quad f(x),
	\end{equation}
	where $X\subset\mathbb{R}^n$ is a convex set and $f(\cdot):\mathbb{R}^n\rightarrow\mathbb{R}$ is a concave function. Let $X^*$ denote the set of optimal solutions. Suppose that the update $x^{(t+1)}=\text{P}_{\mathcal{X}}\left(x^{(t)}+\eta^{(t)}\gamma^{(t)}\right)$ is applied, where $\gamma^{(t)}\in\partial f\left(x^{(t)}\right)$ is a subgradient. If $\eta^{(t)}\rightarrow0$, $\sum_{t=1}^{\infty}\eta^{(t)}=\infty$, $\sum_{t=1}^{\infty}{\left[\eta^{(t)}\right]}^2\leq\infty$ and the sequence $\left\{\gamma^{(t)}\right\}$ is bounded, then $x^{(t)}\rightarrow x^*\in X^*$.
\end{lemma}

Then the proof of Prop.~\ref{th:convergence} is given below.

\begin{proof}
Recall that the optimization problem for dual estimation is
\begin{equation}
    \label{eq:dual_approx_cvx_re}
	\tilde{\lambda}_j^{\alpha} \in \argmin_{\lambda_j\geq0}\max_{\pi_{\alpha}\in\Pi_c} V_{r,j}^{\pi_{\alpha}}\left(s\right) + \lambda_j\left( V_{c,j}^{\pi_{\alpha}}\left(s\right) - \tau \right).
\end{equation}
By assumption~\ref{th:slater_trans}, the primal problem is feasible and let $V_{r,j}^{\pi_{\alpha}^*}(s)$ denote the optimal value. Given the dual function according to the Eq. (\ref{eq:dual}) in the main paper
\begin{align}
    \label{eq:dual_cvx}
    d\left(\lambda_j\right) & = \max_{\pi_{\alpha}\in\Pi_c} L(\pi_{\alpha},\lambda_j) \notag \\
    & = \max_{\pi_{\alpha}\in\Pi_c} V_{r,j}^{\pi_{\alpha}}\left(s\right) + \lambda_j\left( V_{c,j}^{\pi_{\alpha}}\left(s\right) - \tau \right),
\end{align}
the dual optimization problem is then
\begin{equation}
    \label{eq:dual_prob}
    \min_{\lambda_j \geq 0} d\left(\lambda_j\right).
\end{equation}
By Lemma~\ref{th:duality},
strong duality holds for the problem in Eq.~(\ref{eq:dual_approx_cvx_re}). Therefore, $d(\lambda_j)\geq V_{r,j}^{\pi_{\alpha}^*}(s)$ for every $\lambda_j\geq 0$ and the set of optimal solutions in Eq.~(\ref{eq:dual_prob}) is nonempty. The dual function in Eq.~(\ref{eq:dual_cvx}) is also convex~\cite{boyd2004convex}.

Let $\pi_{\alpha,t}^* \in \argmax_{\pi_{\alpha}\in\Pi_c} L(\pi_{\alpha}, \lambda^{(t)})$. The output of the update step in Eq.~(\ref{eq:up_1})
\begin{equation}
    i^{(t+1)} = \argmax_{i\in\left[N\right]} V_{r,j}^{\tilde{\pi}_i}\left(s\right) + \lambda^{(t)}\left( V_{c,j}^{\tilde{\pi}_i}\left(s\right) - \tau \right),
\end{equation}
will give a value $L\left(\tilde{\pi}_{i^{\left(t+1\right)}}, \lambda^{(t)}\right)  \geq L\left(\pi_{\alpha,t}^*,\lambda^{(t)}\right)$ due to the linearity of expectation in $\Pi_{\alpha}$. Therefore, $L\left(\tilde{\pi}_{i^{\left(t+1\right)}}, \lambda^{(t)}\right) = L\left(\pi_{\alpha,t}^*,\lambda^{(t)}\right)$. Hence, the update step in Eq.~(\ref{eq:up_2})
\begin{equation}
    \lambda^{(t+1)} = \text{P}_{\mathbb{R}_{\geq 0}} \left( \lambda^{(t)} - \eta^{(t)}\left( V_{r,j}^{\tilde{\pi}_{i^{\left(t+1\right)}}}(s)-\tau \right) \right),
\end{equation}
is optimizing with a subgradient $V_{r,j}^{\tilde{\pi}_{i^{\left(t+1\right)}}}(s)-\tau \in \partial d(\lambda^{(t)})$, which is also bounded since the value function is bounded. Finally by the condition of the sequence of the step sizes and Lemma~\ref{th:sub_opt}, which can be adapted to minimization of a convex function, we conclude the proof.

\end{proof}

\subsection{Proof of Proposition 3} \label{sec:app3}

Prop. 3 can be proved by similar steps of the Prop. 4.1 in~\cite{shapiro2002}.

\section{Implementation details}

In this section we describe in detail of the environmental setup and training details of our empirical studies. Four-Room and Reacher are two benchmarks for  RL transfer and safety, and we adopted the similar configurations and hyper-parameters used in \cite{gimelfarb2021riskaware}. We also introduced a challenging task in SafetyGym, which has high state dimensions and complex physical dynamics.

\subsection{Four-Room}

The state in Four-Room consists of agent's $2$-dimensional location and a binary vector indicating whether an object has been picked up, thus, $\mathcal{S}=J^2\times\{0,1\}^{18}$ where $J=\{1,2,\dots,13\}\subset\mathbb{N}$. The agent has 4 actions of moving up, down, left or right. Each task has unique values of the object rewards sampled uniformly between $[-1,+1]$. The agent receives zero rewards in empty locations. Therefore, some objects (with positive rewards) are beneficial while objects with negative rewards should be avoided by the agent. The goal state has a reward of 2, and the cost of the unsafe traps is -0.1, which is consistent with the expected cost $\mathbb{E}\left[c\left(s,a,s^{\prime}\right)\right]=5\%\times(-2)$ used in the uncertainty environment of~\rasfql{}. Each task is trained for 20,000 interactions with the environment with an episode length of 200. All 128 tasks are sequentially trained. Thus, the total number of training iterations is 2.56 million. When started in a new task, the successor feature is initiated from the one learned in previous task. We then estimate the dual variable and use policy transfer during transfer learning on the new task. Some hyper-parameters of $Q$-learning in training are: discount factor $\gamma=0.95$, epsilon-greedy for exploration $\epsilon=0.12$, learning rates for successor feature, reward vector and $\lambda$ are all $\alpha=0.5$. For dual estimation during transfer, the learning rate $\eta^{(t)} = c\cdot\frac{1}{t}$, where $c=1000$ is a constant. The threshold $\tau=-0.000005$. Successor features are represented with tables and computation is performed with CPUs. We report results using means and standard deviations calculated from 10 independent runs. Plots of per-task performance are results averaged over 8 tasks.

\subsection{Reacher}

The state space $\mathcal{S}\subset\mathbb{R}^4$ for the 2-link robot arm. Since Reacher is an environment with continuous state and action spaces, we discretize the action space to $\{-1,0,1\}^2$ by following prior work (SFQL and RaSFQL). The goals in training tasks are located in $(0.14, 0)$, $(0, 0.14)$, $(-0.14, 0)$ and $(0, -0.14)$. The locations for 8 test tasks are $(0.22, 0)$, $(0, 0.22)$, $(-0.22, 0)$, $(0, -0.22)$, $(0.1, 0.1)$, $(-0.1, 0.1)$, $(-0.1, -0.1)$ and $(0.1, -0.1)$. There are 6 unsafe round regions centered in $(0.14, 0)$, $(-0.14, 0)$, $(0.22, 0)$, $(-0.22, 0)$, $(0.1, 0.1)$ and $(-0.1, -0.1)$, respectively, whose radii are all 0.06. The reward is proportional to the negative distance $d$ between the robot arm end effector and the goal location $r = 1-4d$. The cost of the unsafe region is -0.1. Each task is trained for 100,000 interactions with the environment with an episode length 500. We use a two-hidden-layer neural network for deep successor feature, where the size of the hidden layer is 256. The network is trained using stochastic gradient descent (SGD) with a learning rate of 0.001, a buffer size 400,000 and a batch size of 32. Other hyper-parameters are: discount factor $\gamma=0.9$, epsilon-greedy for exploration $\epsilon=0.1$, learning rate for the dual variable $\alpha=10$ and threshold $\tau=-0.1$. The reacher environment is provided by the open-source PyBullet Gymperium packages~\cite{benelot2018} for the MuJoCo physics engine~\cite{conf/iros/TodorovET12}. The model is trained with GPUs and each trail (sequential training over 4 tasks) takes around 7 hours. We report results using means and standard deviations calculated from 10 independent runs.

\subsection{SafetyGym}

This environment has a larger state dimensionality $\mathcal{S}\subset\mathbb{R}^{77}$. The state space consists of the following components: readings from motor sensors of dimension 12, lidar readings of all 4 types of objects (button, goal, trap and wall) in 16 angles surrounding the robot, which bring up to a dimension of 64 in total, and finally a single-dimension indicator variable to represent whether the agent is inside a trap. We also discretize the action space in a similar fashion to Reacher. The two traps are located in $(0, 0)$ and $(0.5, -0.5)$. The goal is located in $(0.5, 0.5)$. The two buttons are located in $(-0.5, 0.5)$ and $(0.5, -0.5)$, therefore the former button is in a safe region while the latter is inside a trap, we refer to the former as `safe button' and the latter `risky button'. The robot starts in $(-0.5, -0.5)$. The reward in this environment is sparse, it is only obtained upon touching the buttons or the goal. The trap has a cost of 1 upon entrance. We train 4 tasks in total and in each task we alter the rewards of the buttons. The rewards of the `safe button' and the `risky button' in the 4 tasks are $(1, 4)$, $(2, 1)$, $(2, 5)$ and $(3, 3)$ respectively. Every task is trained for 200,000 steps with an episode length 500. We use the same network used for Reacher. We use a buffer size 100,000 and a batch size of 64. Other hyper-parameters are: discount factor $\gamma = 0.99$, epsilon-greedy for exploration $\epsilon = 0.25$, learning rate for the dual variable $\alpha = 10$ and threshold $\tau = -0.001$. We report results using means and standard deviations calculated from 10 independent runs.

\section{Additional studies}

\subsection{Constraint satisfaction}\label{sec:c3}
\begin{figure*}
    \centering
	\begin{subfigure}[t]{.24\textwidth}
		\includegraphics[width=1\textwidth]{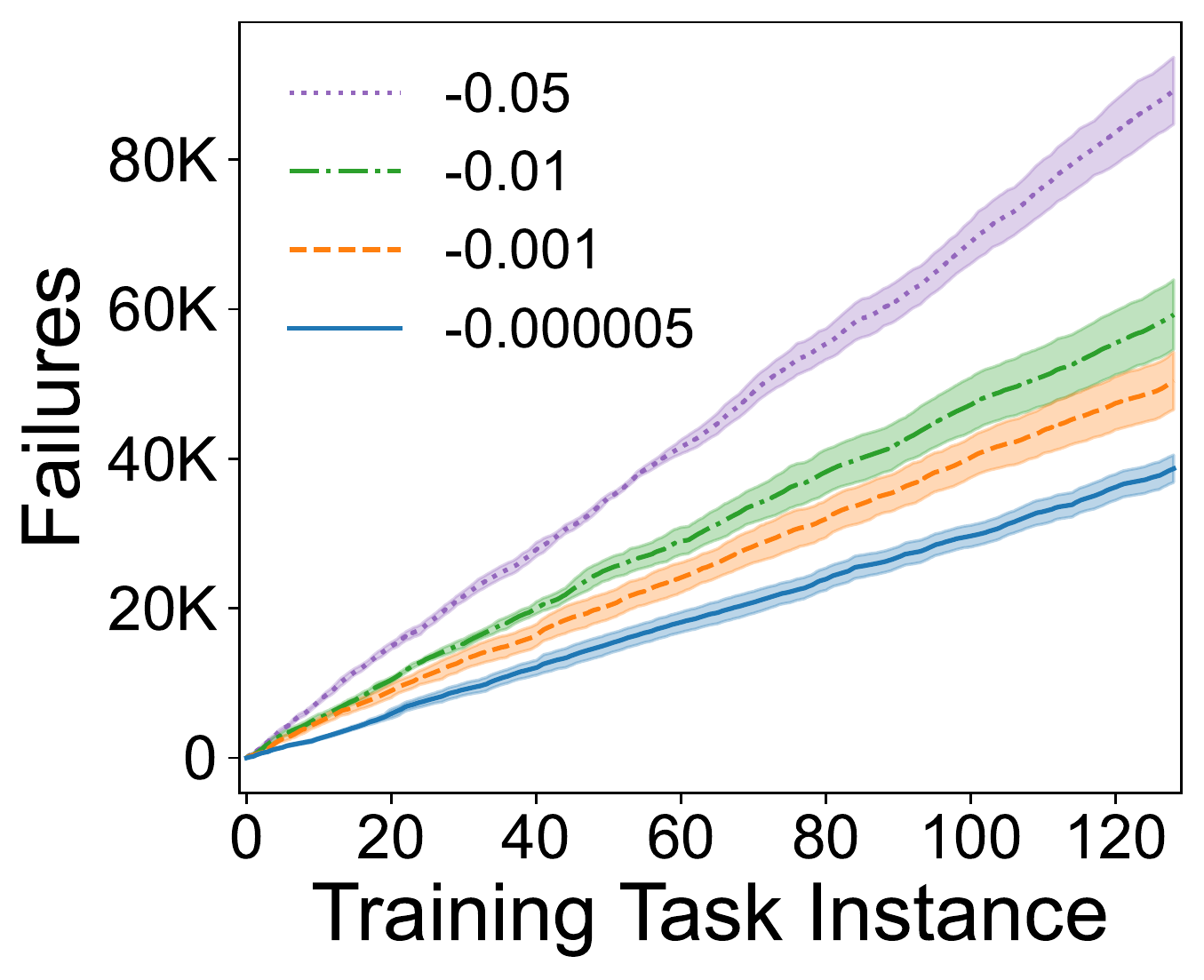}
		\caption{Accumulated failures.}
	\end{subfigure}
	\begin{subfigure}[t]{.24\textwidth}
		\includegraphics[width=1\textwidth]{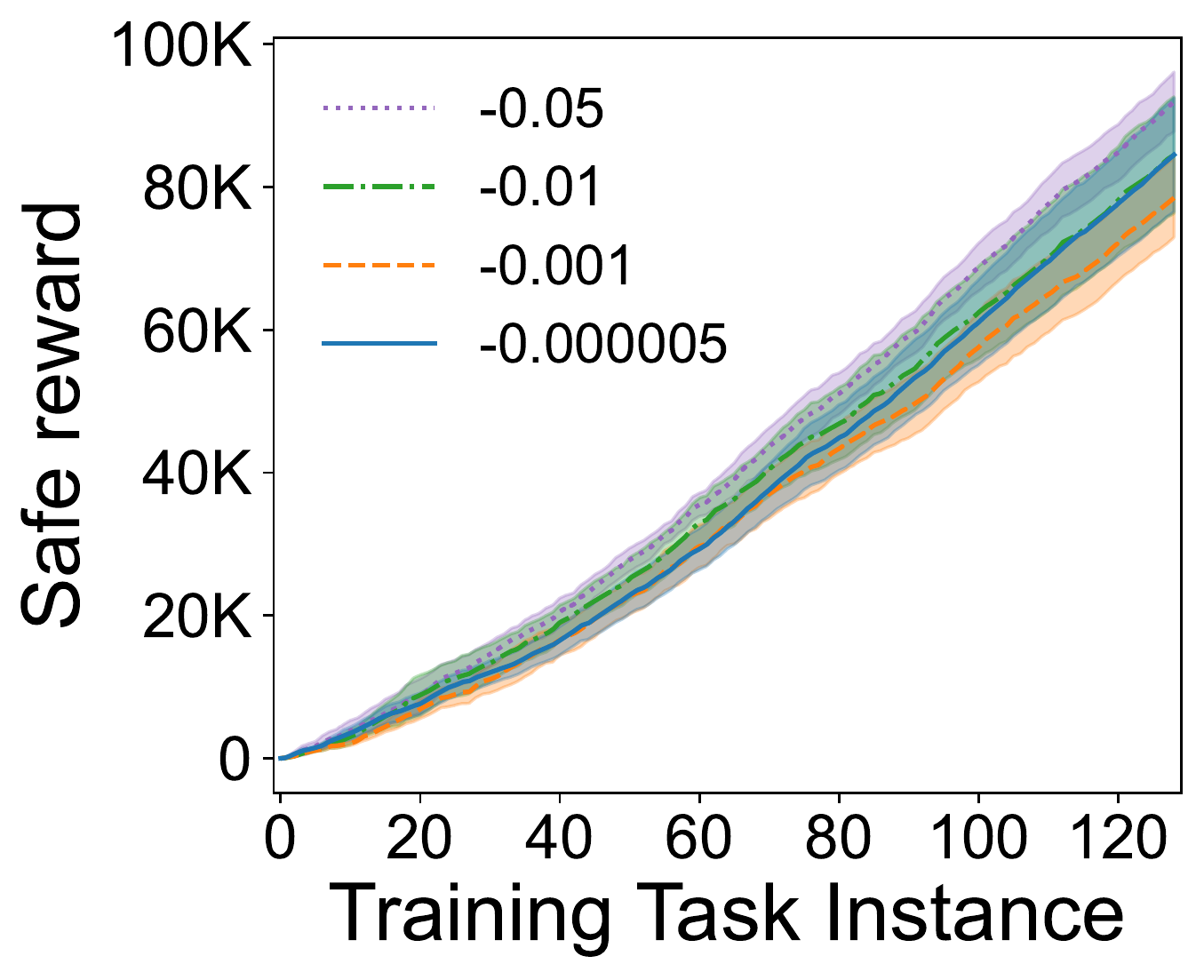}
		\caption{Accumulated rewards from safe objects.}
	\end{subfigure}
	\begin{subfigure}[t]{.24\textwidth}
		\includegraphics[width=1\textwidth]{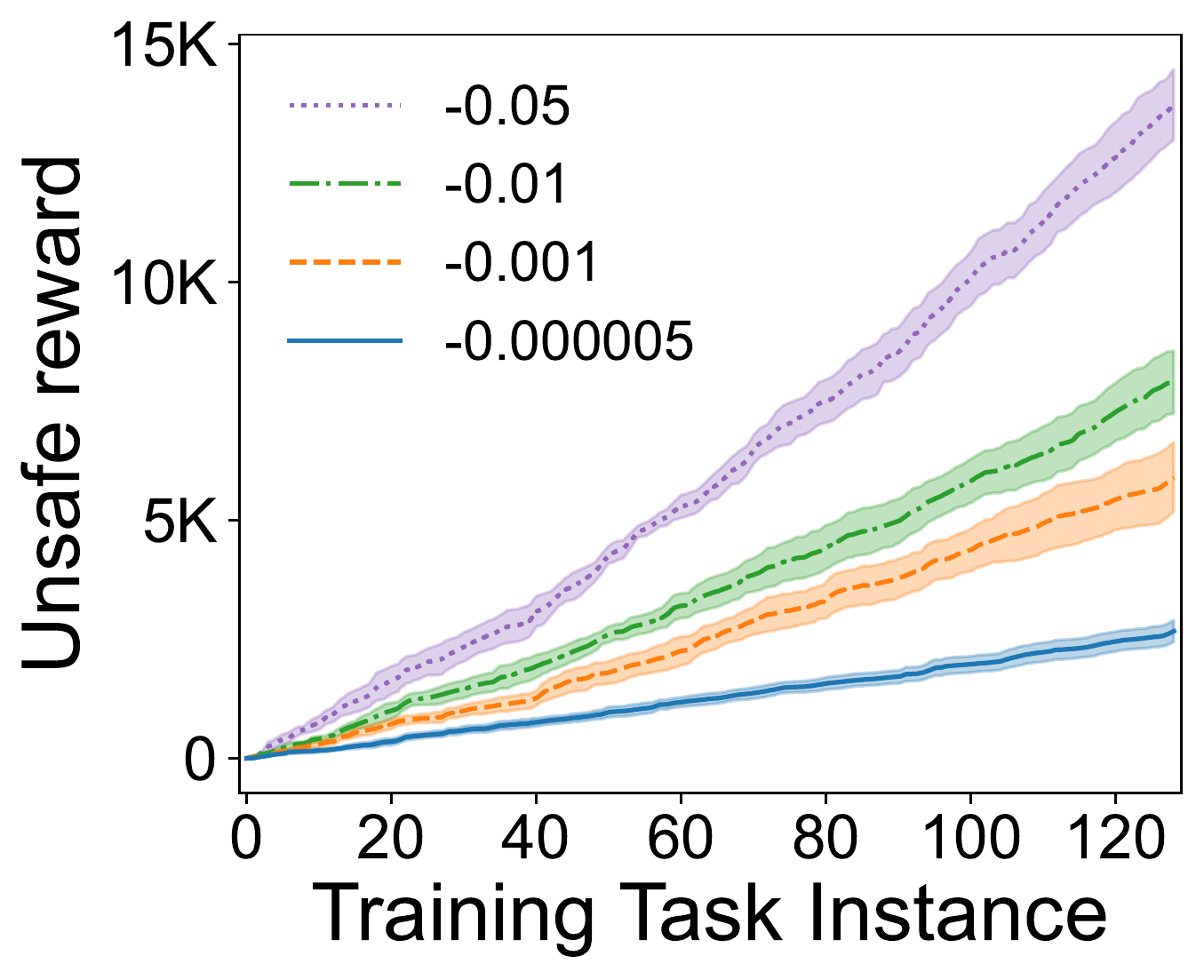}
		\caption{Accumulated rewards from unsafe objects.}
	\end{subfigure}
	\begin{subfigure}[t]{.24\textwidth}
	    \includegraphics[width=1\textwidth]{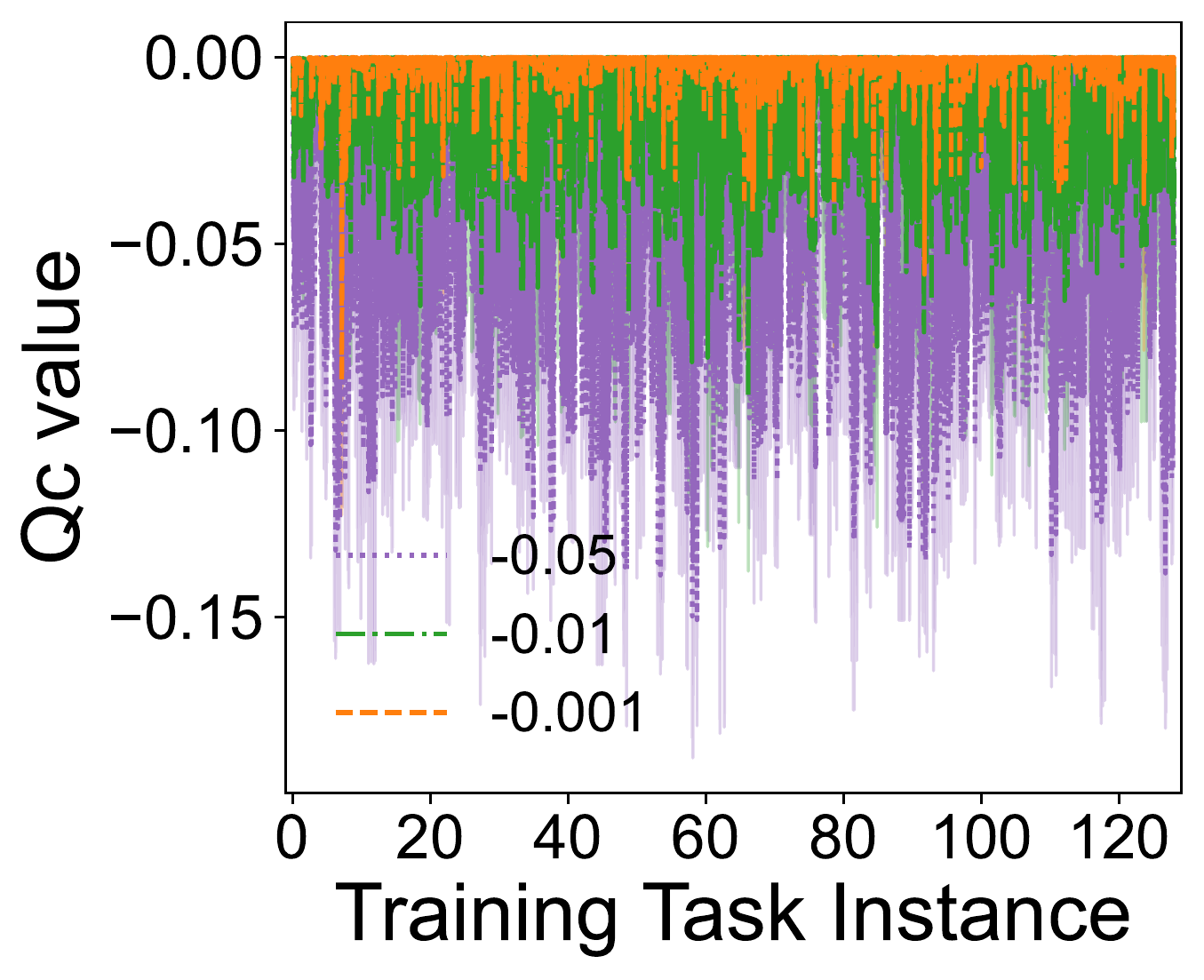}
		\caption{Value of $Q_{c,j}^{\pi}$.}
		\label{fig:4room_threshold_qc}
	\end{subfigure}
	\caption{Performance of \method{} with different threshold $\tau$ on the Four-Room domain. We compute accumulated (a) failures, (b) rewards from safe objects, (c) rewards from unsafe objects and (d) the value of $Q_{c,j}^{\pi}$, over the training task instances.}
	\label{fig:4room_threshold}
\end{figure*}
\begin{figure*}
    \centering
	\begin{subfigure}[t]{.24\textwidth}
		\includegraphics[width=1\textwidth]{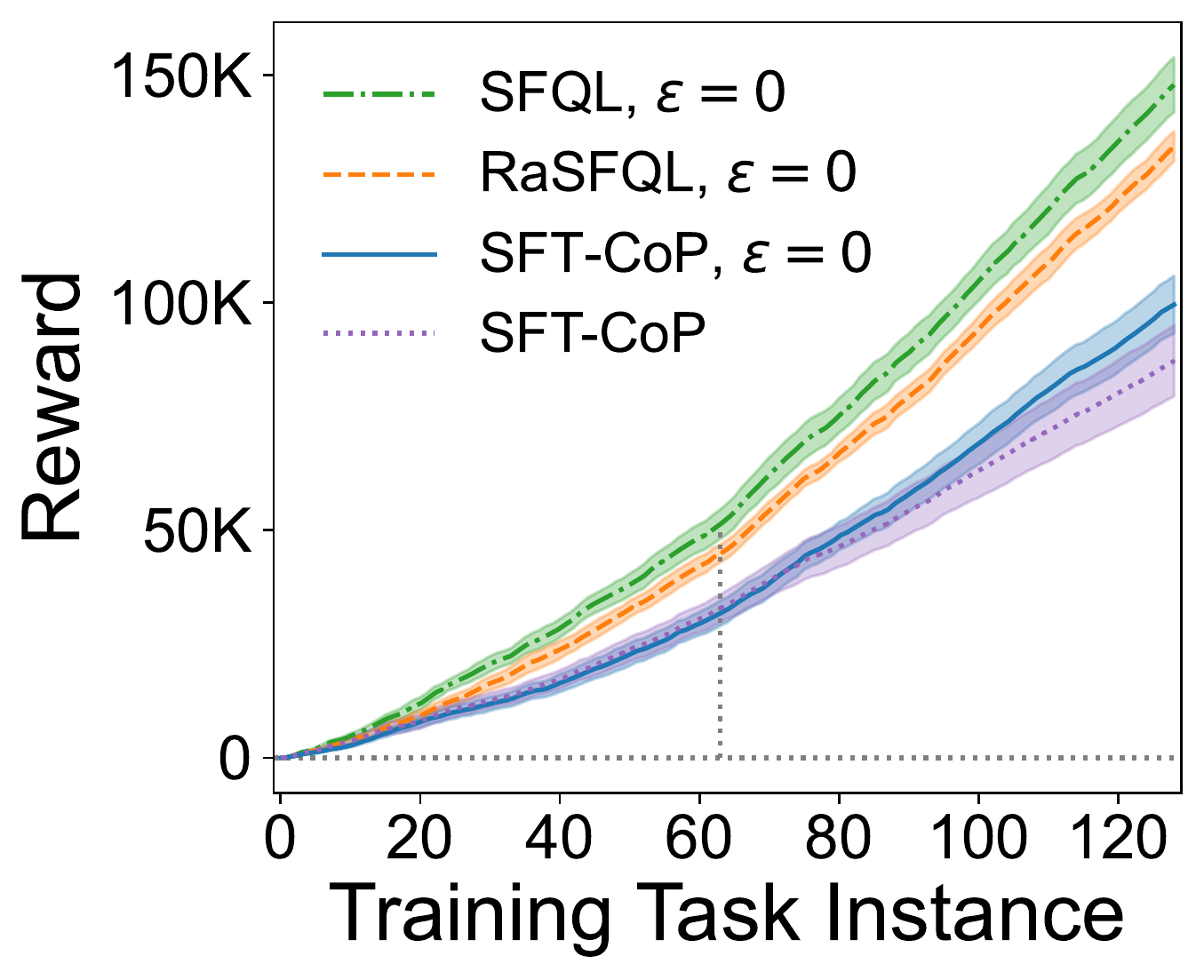}
		\caption{Accumulated rewards.}
	\end{subfigure}
	\begin{subfigure}[t]{.24\textwidth}
		\includegraphics[width=1\textwidth]{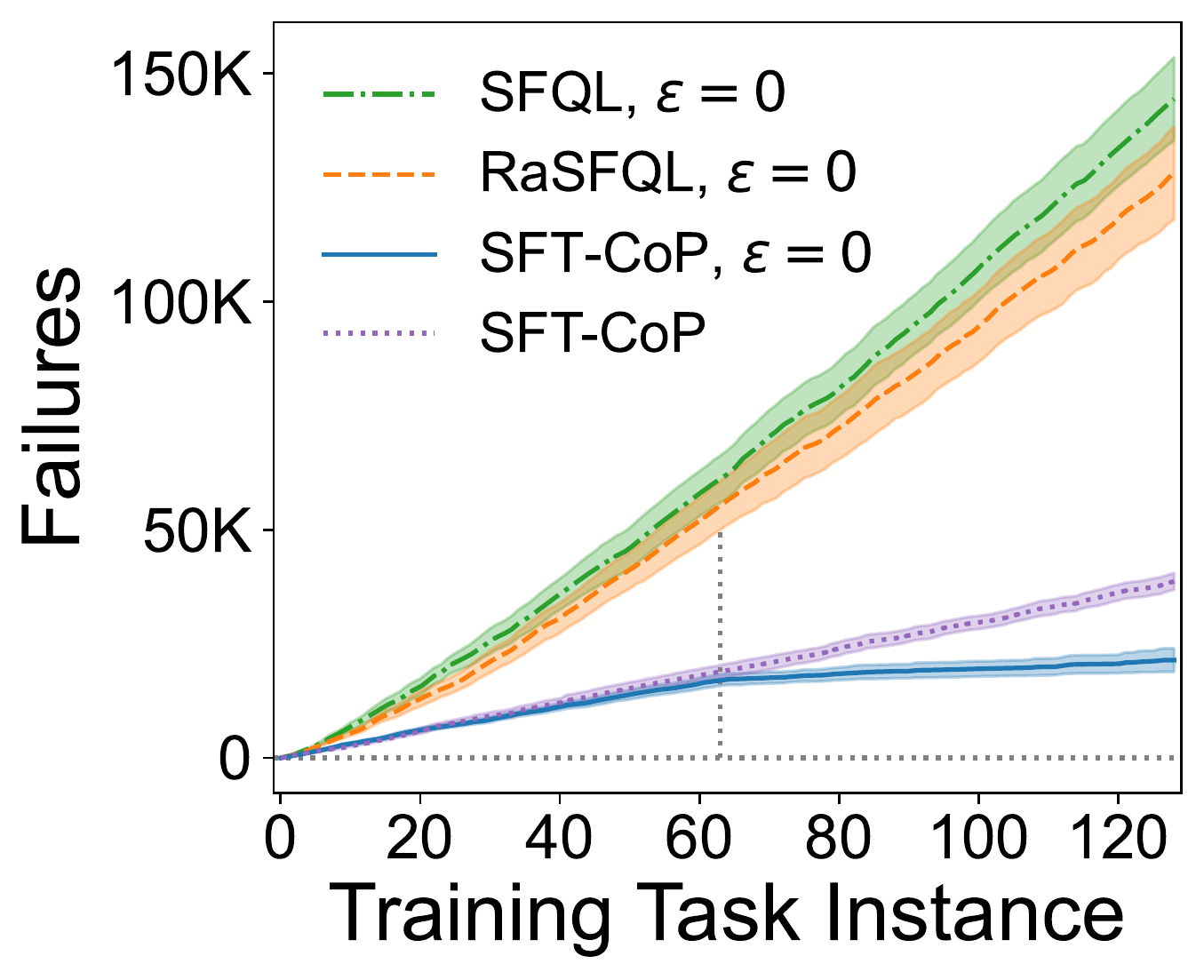}
		\caption{Accumulated failures.}
	\end{subfigure}
	\caption{Accumulated (a) rewards and (b) failures of \sfql, \rasfql{} ($\beta=2$) and \method{} on the Four-Room domain with the parameter of epsilon-greedy exploration $\epsilon$ set to $0$ after task 64.}
	\label{fig:4room_epsilon}
\end{figure*}

Unlike the risk-aware method that learns to avoid any return with variance, the constraint formulation is more flexible in handling varying tolerance for violations. Fig.~\ref{fig:4room_threshold} shows that by setting different thresholds, \method{} can satisfy the constraints during transfer learning (Fig.~\ref{fig:4room_threshold_qc} exhibits the $Q$ values are aligned with different thresholds). As expected, \method{} has fewer failures when the threshold is tighter.

The reported curves over training task instances in this paper are training results when sequentially transferred to new tasks, with epsilon-greedy based exploration. Note that \method{} can achieve almost zero constraint violation when the exploration parameter is set to $0$,~\eg, Fig.~\ref{fig:4room_epsilon} shows that after task 64 the failures curve becomes flat compared with continually increasing failures of epsilon-greedy based training.

\subsection{Ablation study}
\begin{figure*}
    \centering
	\begin{subfigure}[t]{.24\textwidth}
		\includegraphics[width=1\textwidth]{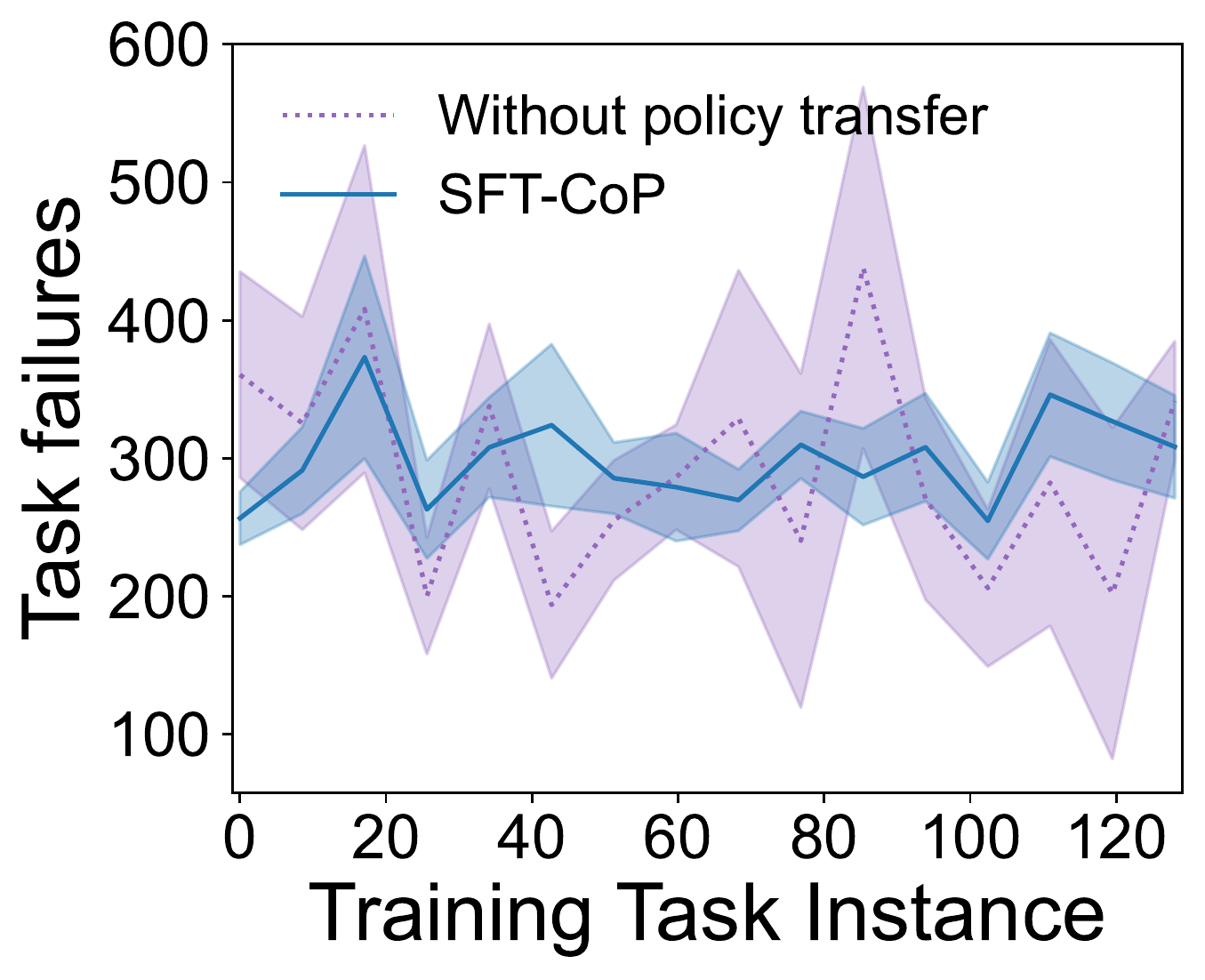}
		\caption{Task failures.}
	\end{subfigure}
	\begin{subfigure}[t]{.24\textwidth}
	    \includegraphics[width=1\textwidth]{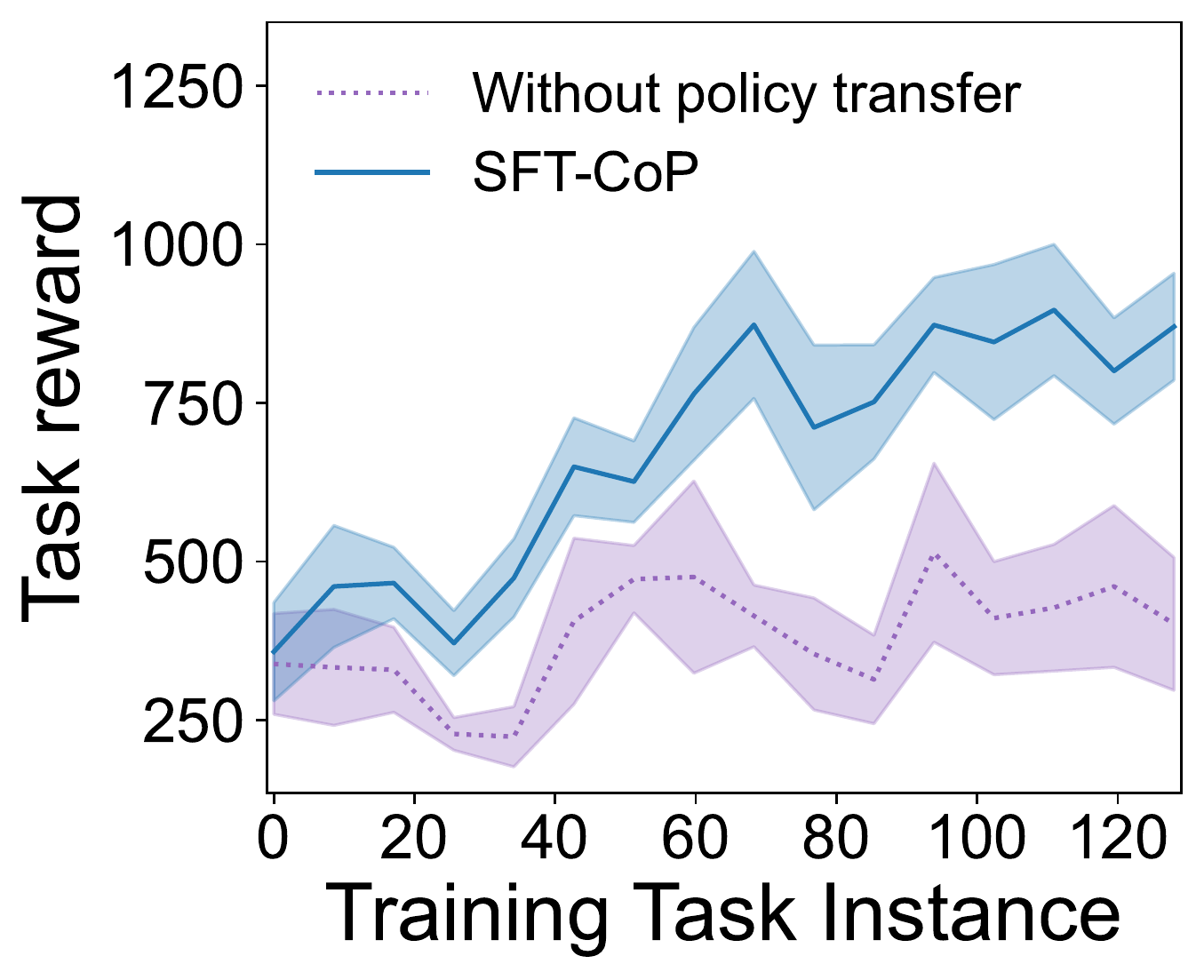}
		\caption{Task rewards.}
	\end{subfigure}
	\begin{subfigure}[t]{.24\textwidth}
		\includegraphics[width=1\textwidth]{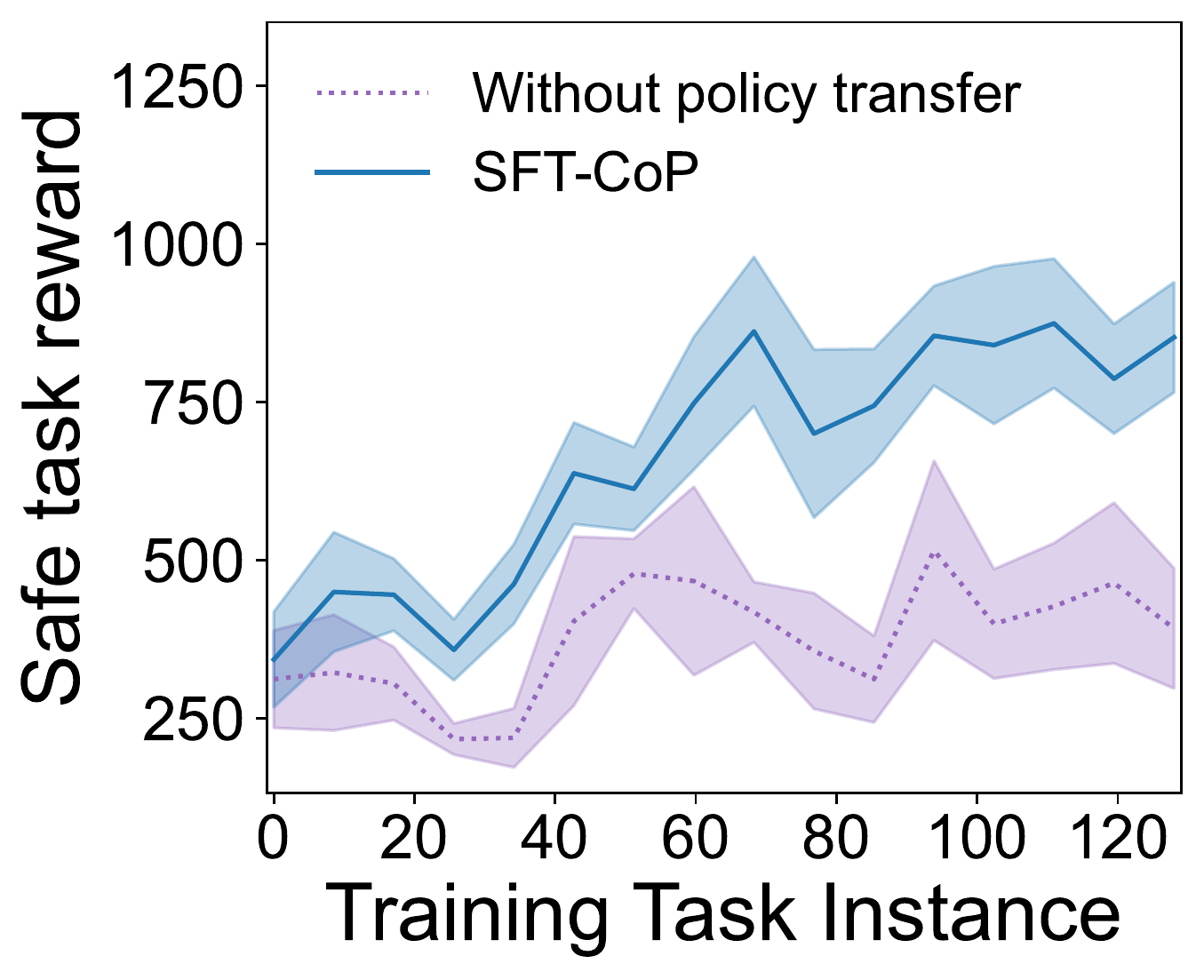}
		\caption{Task rewards from safe objects.}
	\end{subfigure}
	\begin{subfigure}[t]{.24\textwidth}
		\includegraphics[width=1\textwidth]{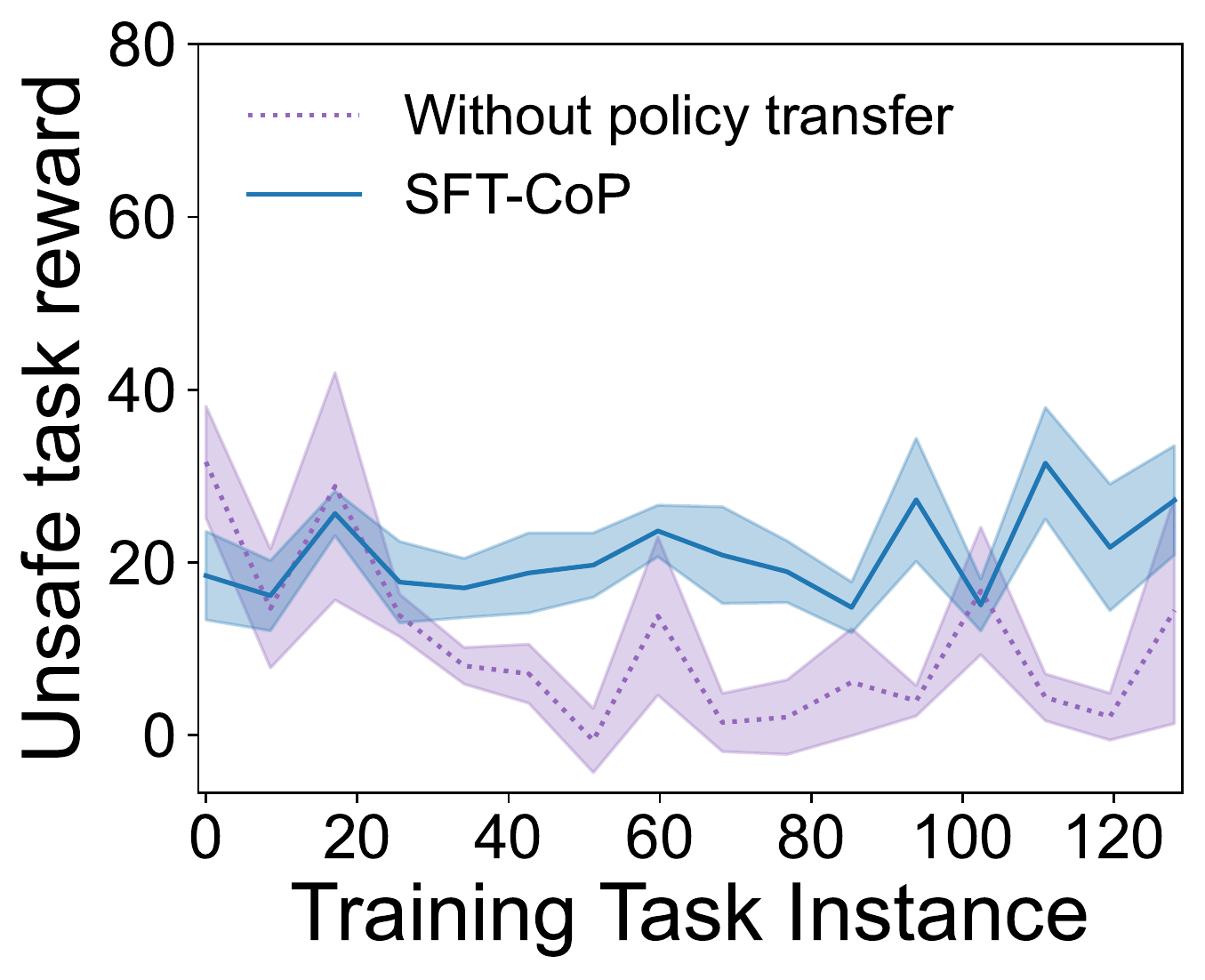}
		\caption{Task rewards from unsafe objects.}
	\end{subfigure}
	\caption{Performance of \method{} w. and w.o. multi-source policy transfer during sequential transfer learning on the Four-Room domain. We compute accumulated (a) failures, (b) total rewards, (c) rewards from safe objects and (d) rewards from unsafe objects, over the training task instances.}
	\label{fig:4room_ablation}
\end{figure*}

In this section, we show the effect of our multi-source policy transfer strategy in Eq. (5) by comparing \method{} against a transfer learning baseline that does not use this strategy. This baseline copies the successor feature from the previous task like \method{} and estimates the optimal dual variable when learning on a new task during transfer. However, it does not use Eq. (5) for policy transfer. Fig.~\ref{fig:4room_ablation} shows that although it can achieve a similar level of failures during the sequential transfer learning, it collects fewer rewards. The performance does not improve  in later tasks as shown by the flat curves. This suggests that without using policy transfer, the learned $Q$ function gives a poorer policy that does not fully explore states.

\subsection{Comparison with~\rasfql{}}
\begin{figure*}
    \centering
	\begin{subfigure}[t]{.24\textwidth}
		\includegraphics[width=1\textwidth]{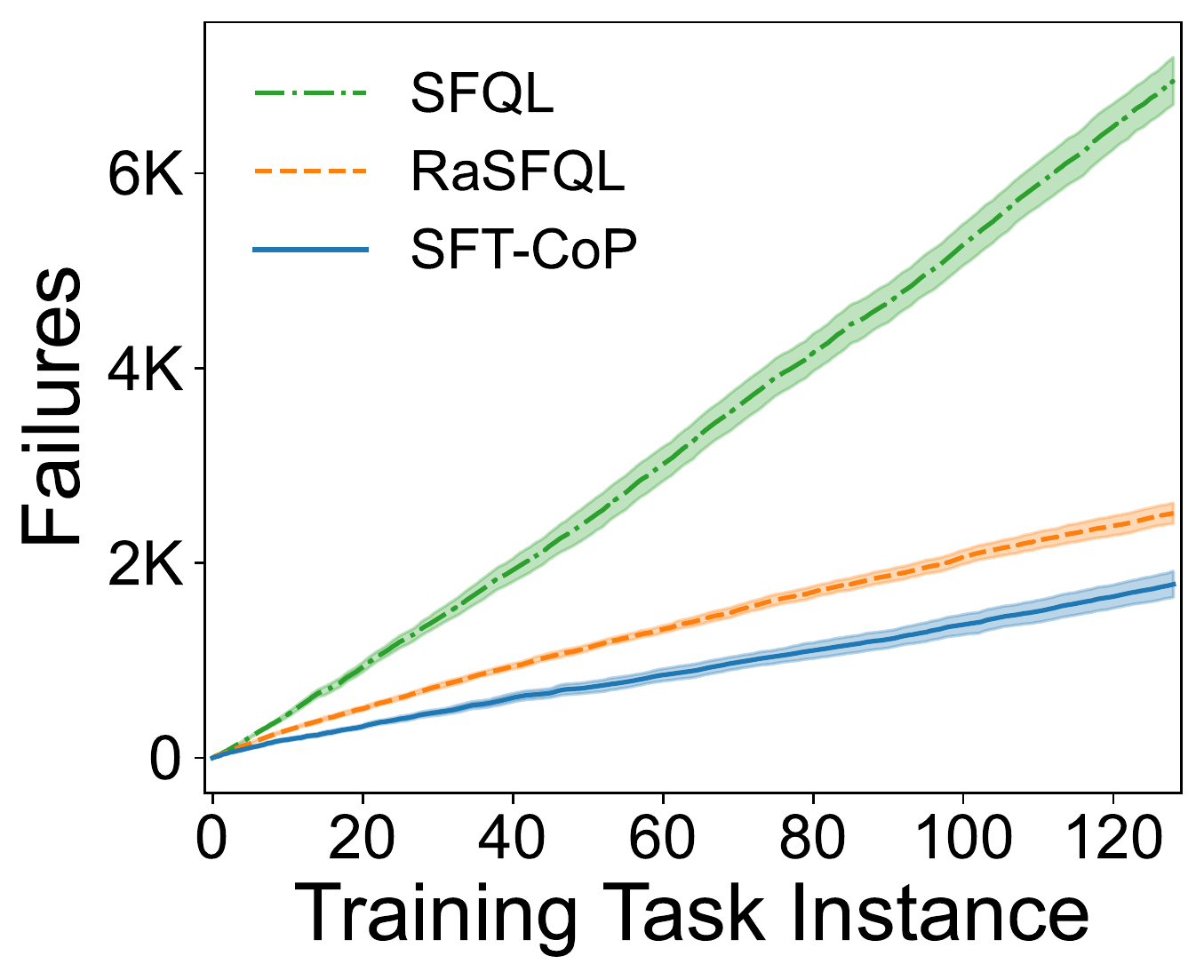}
		\caption{Accumulated failures.}
	\end{subfigure}
	\begin{subfigure}[t]{.24\textwidth}
	    \includegraphics[width=1\textwidth]{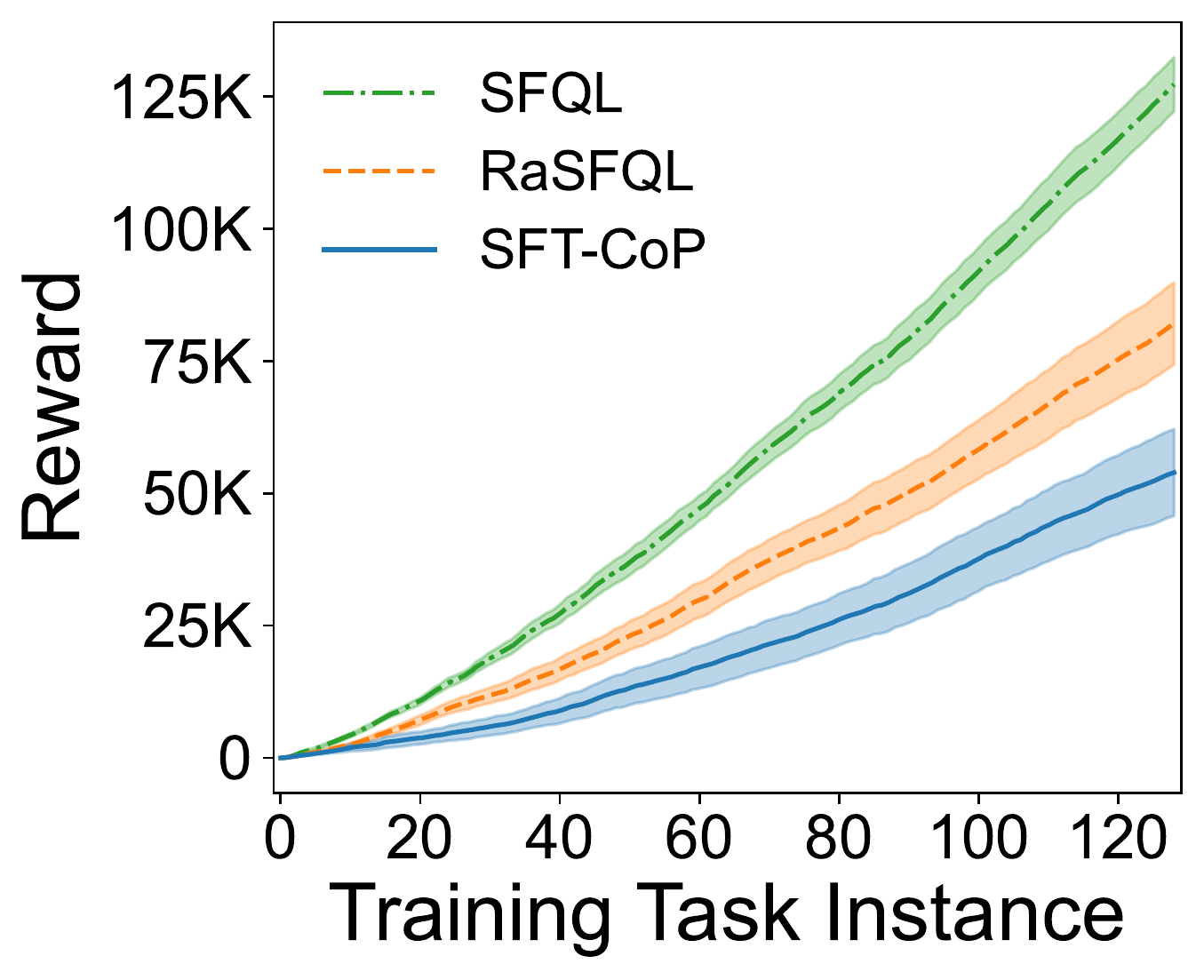}
		\caption{Accumulated rewards.}
	\end{subfigure}
	\begin{subfigure}[t]{.24\textwidth}
		\includegraphics[width=1\textwidth]{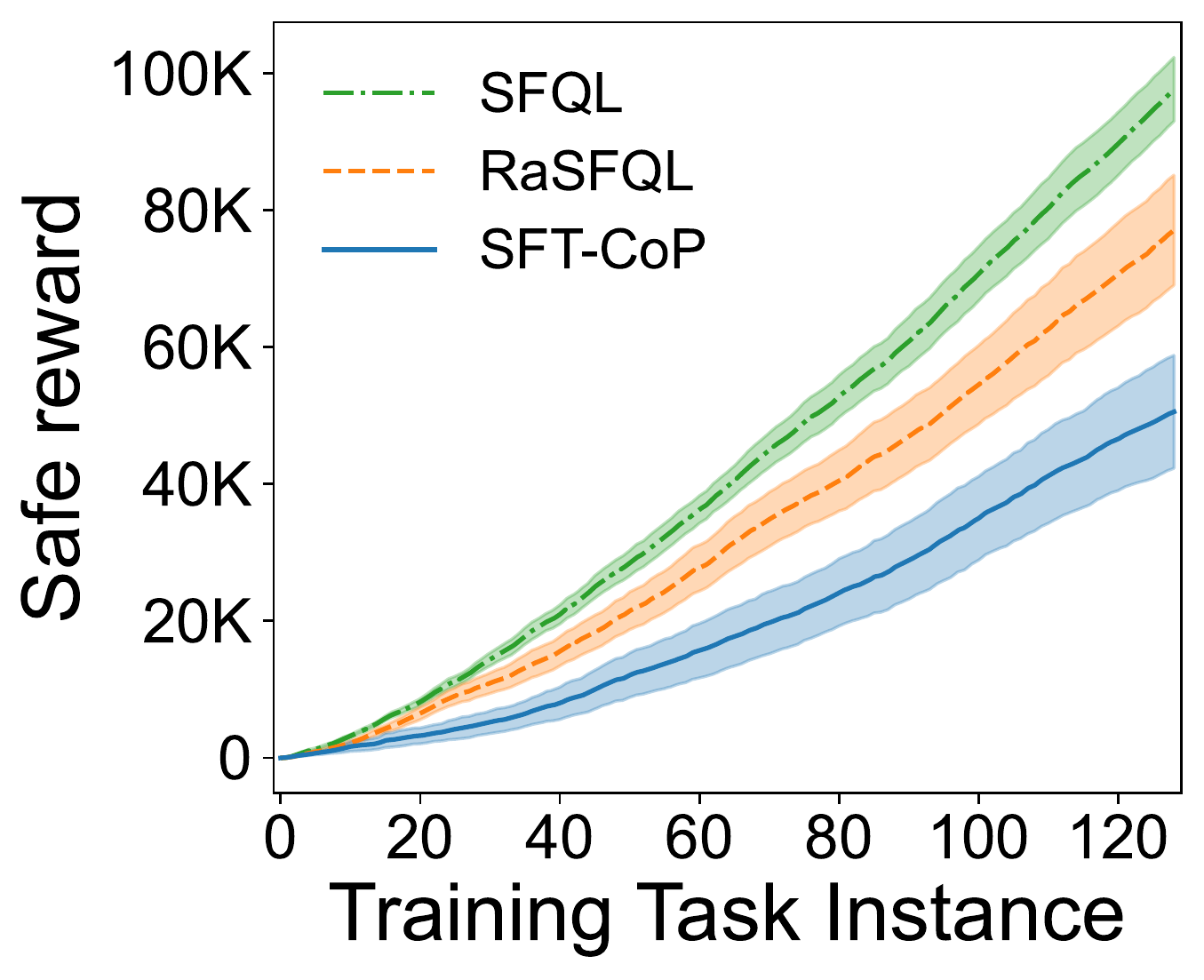}
		\caption{Accumulated rewards from safe objects.}
	\end{subfigure}
	\begin{subfigure}[t]{.24\textwidth}
		\includegraphics[width=1\textwidth]{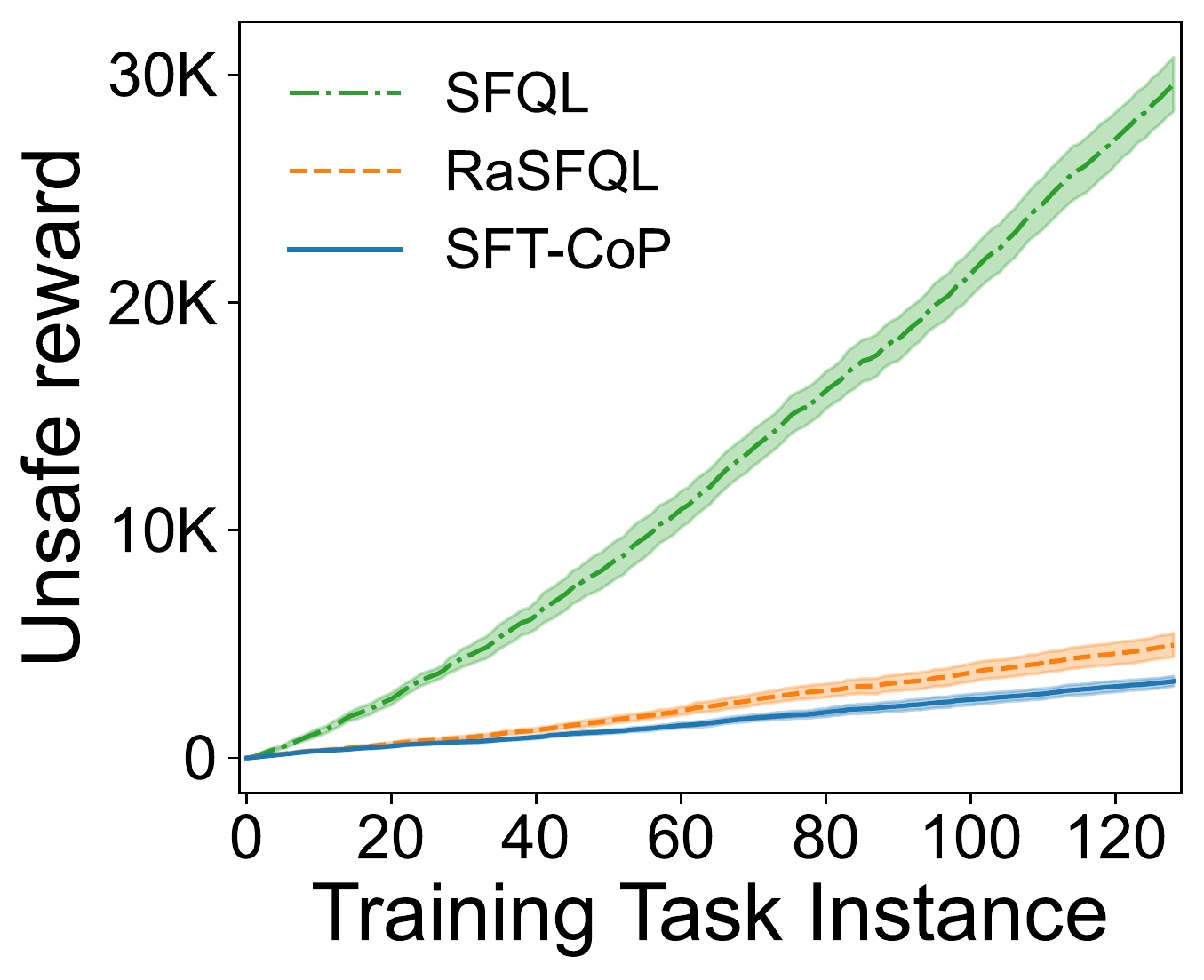}
		\caption{Accumulated rewards from unsafe objects.}
	\end{subfigure}
	\begin{subfigure}[t]{.24\textwidth}
		\includegraphics[width=1\textwidth]{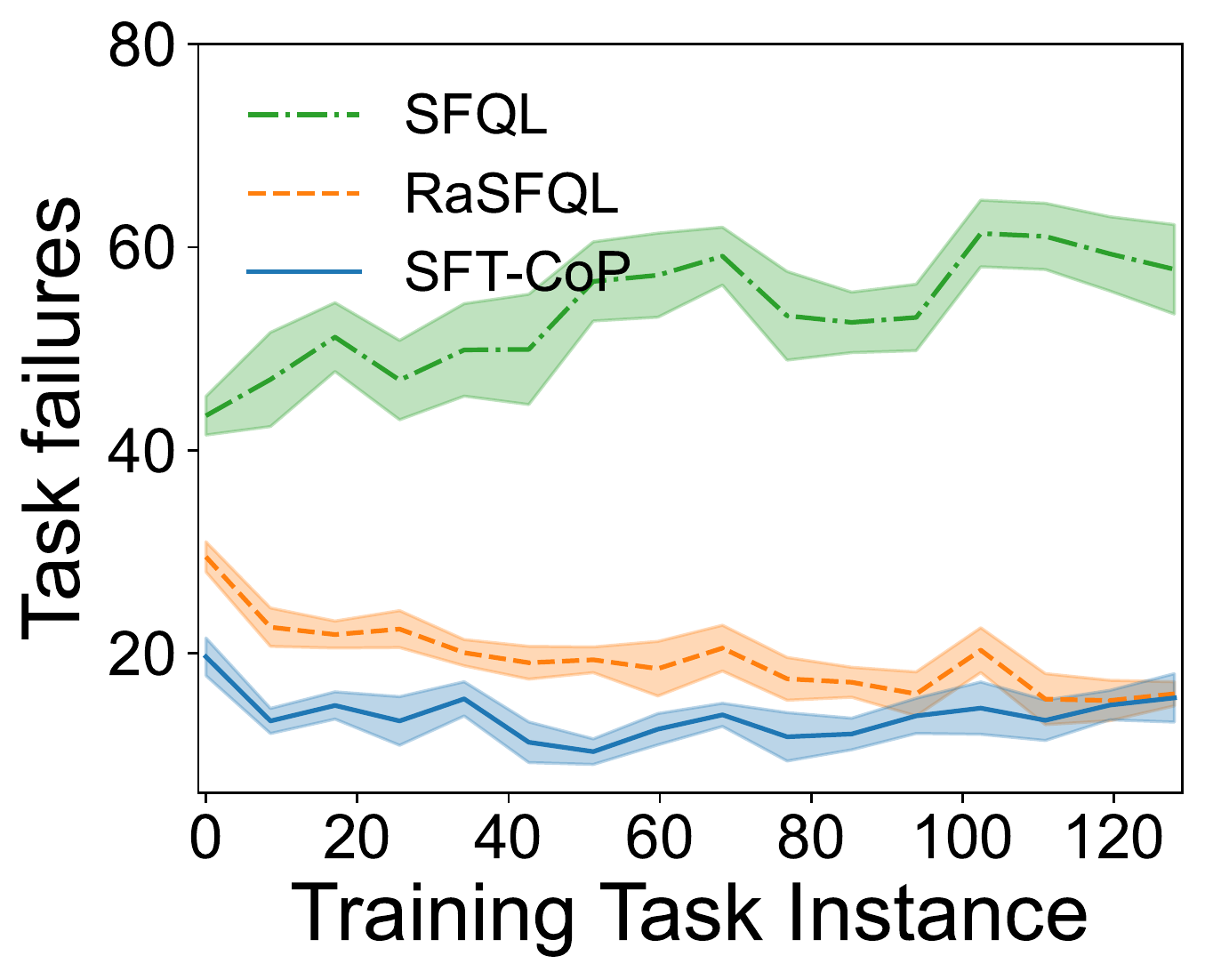}
		\caption{Task failures.}
	\end{subfigure}
	\begin{subfigure}[t]{.24\textwidth}
	    \includegraphics[width=1\textwidth]{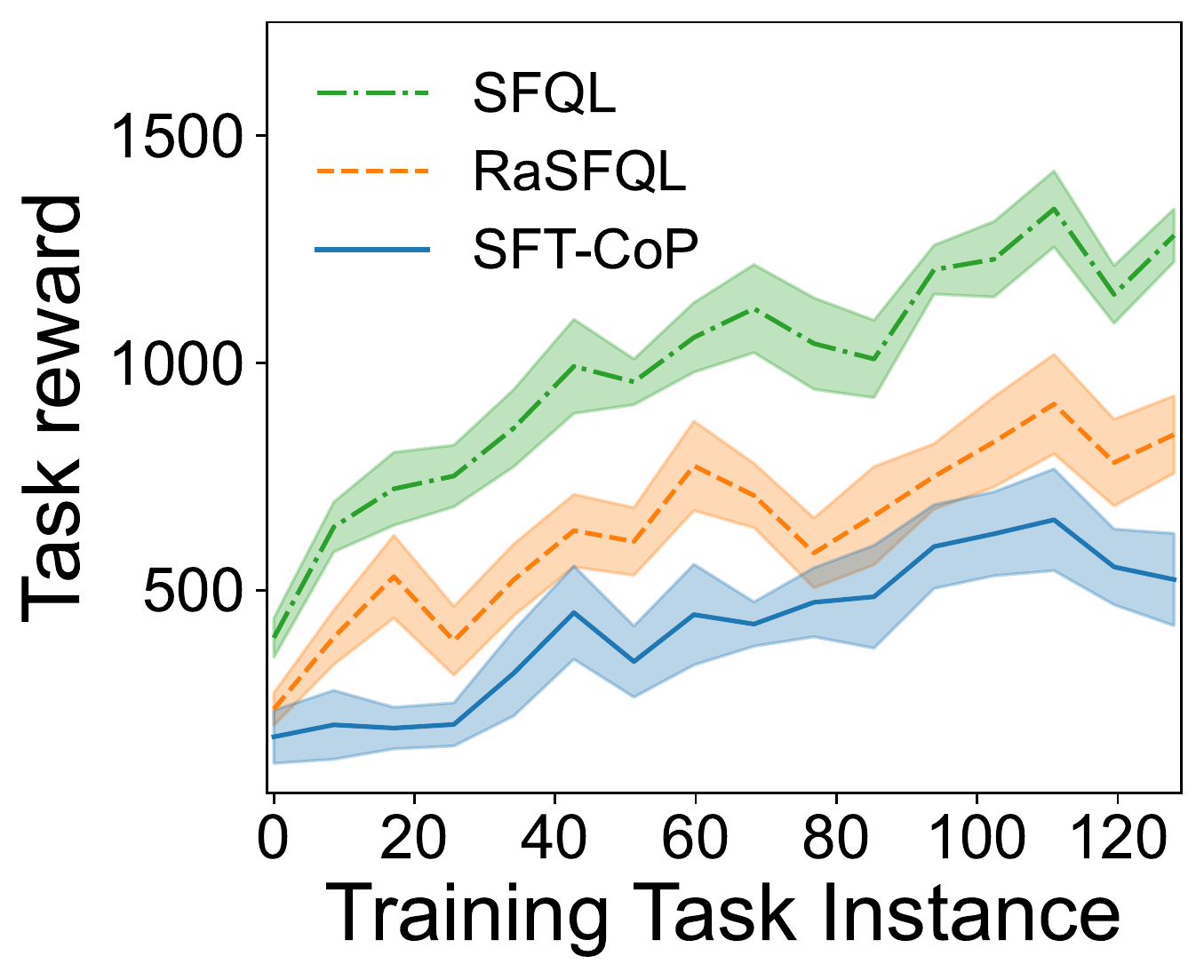}
		\caption{Task rewards.}
	\end{subfigure}
	\begin{subfigure}[t]{.24\textwidth}
		\includegraphics[width=1\textwidth]{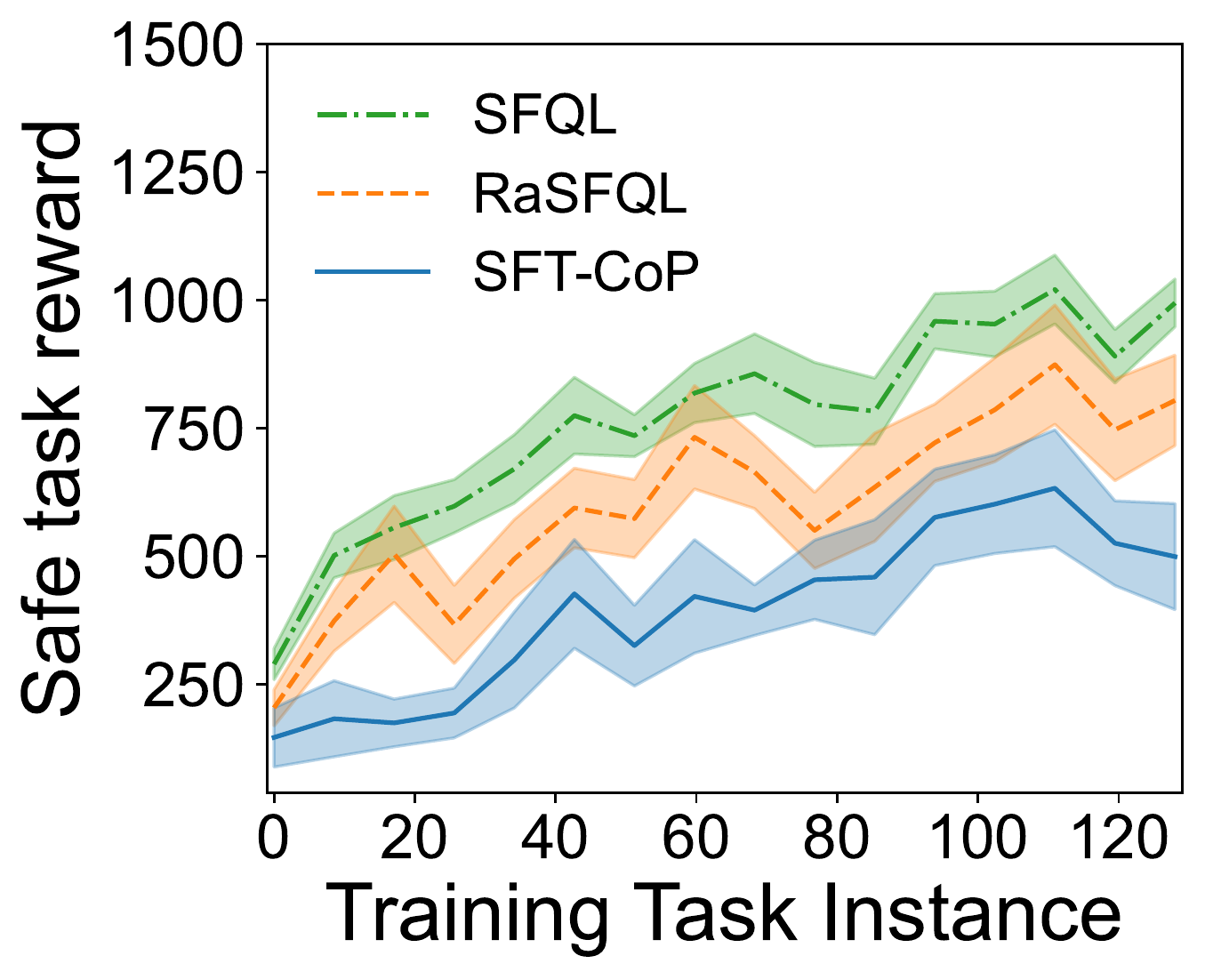}
		\caption{Task rewards from safe objects.}
	\end{subfigure}
	\begin{subfigure}[t]{.24\textwidth}
		\includegraphics[width=1\textwidth]{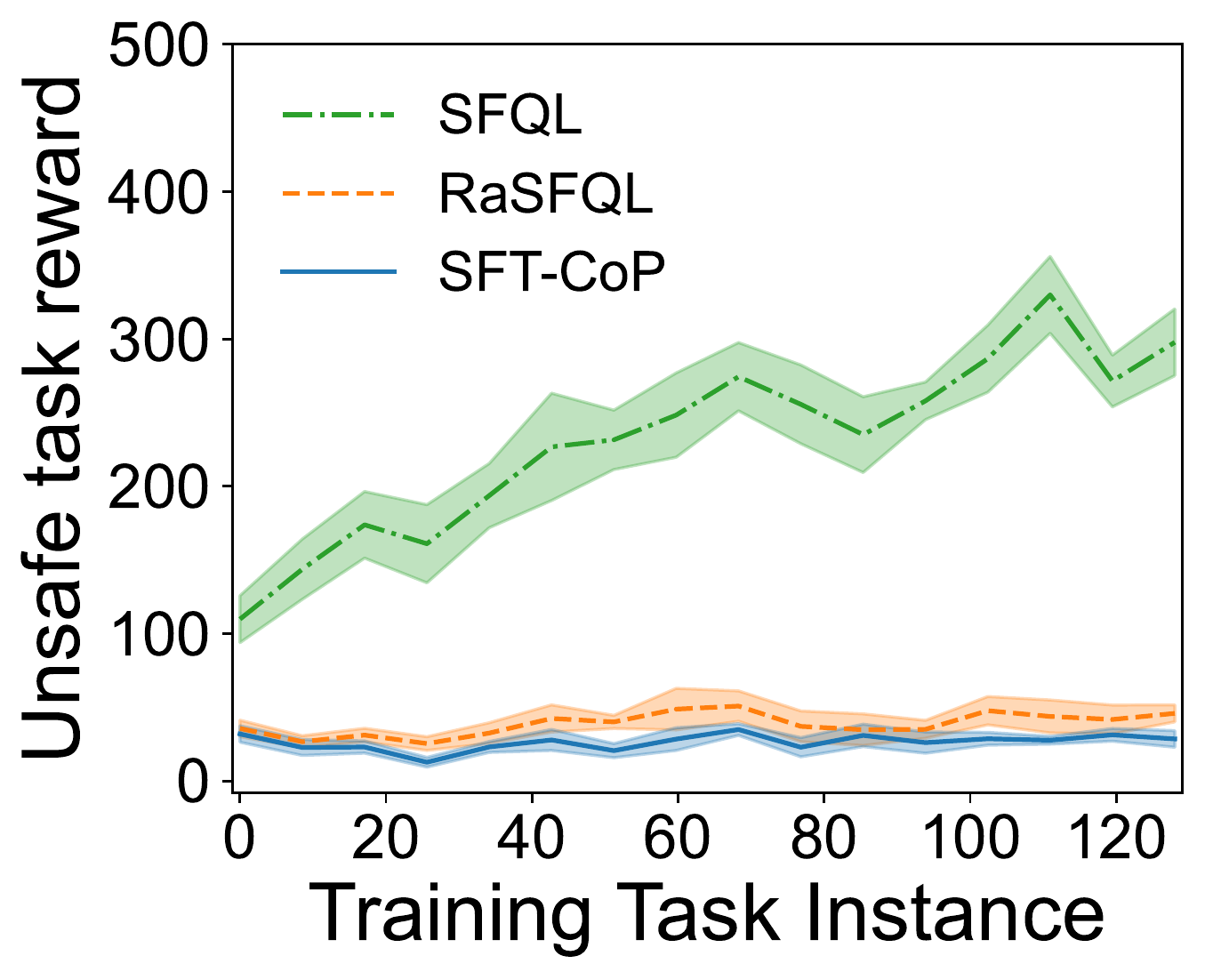}
		\caption{Task rewards from unsafe objects.}
	\end{subfigure}
	\caption{Performance of \sfql, \rasfql{} ($\beta=2$) and \method{} on the Four-Room domain with probabilistic traps. We compute accumulated and per-task (a, e) failures, (b, f) total rewards (c, g) rewards from safe objects and (d, h) rewards from unsafe objects, over the training task instances.}
	\label{fig:4room_lphc}
\end{figure*}
\begin{figure*}
    \centering
	\begin{subfigure}[t]{.24\textwidth}
		\includegraphics[width=1.\textwidth]{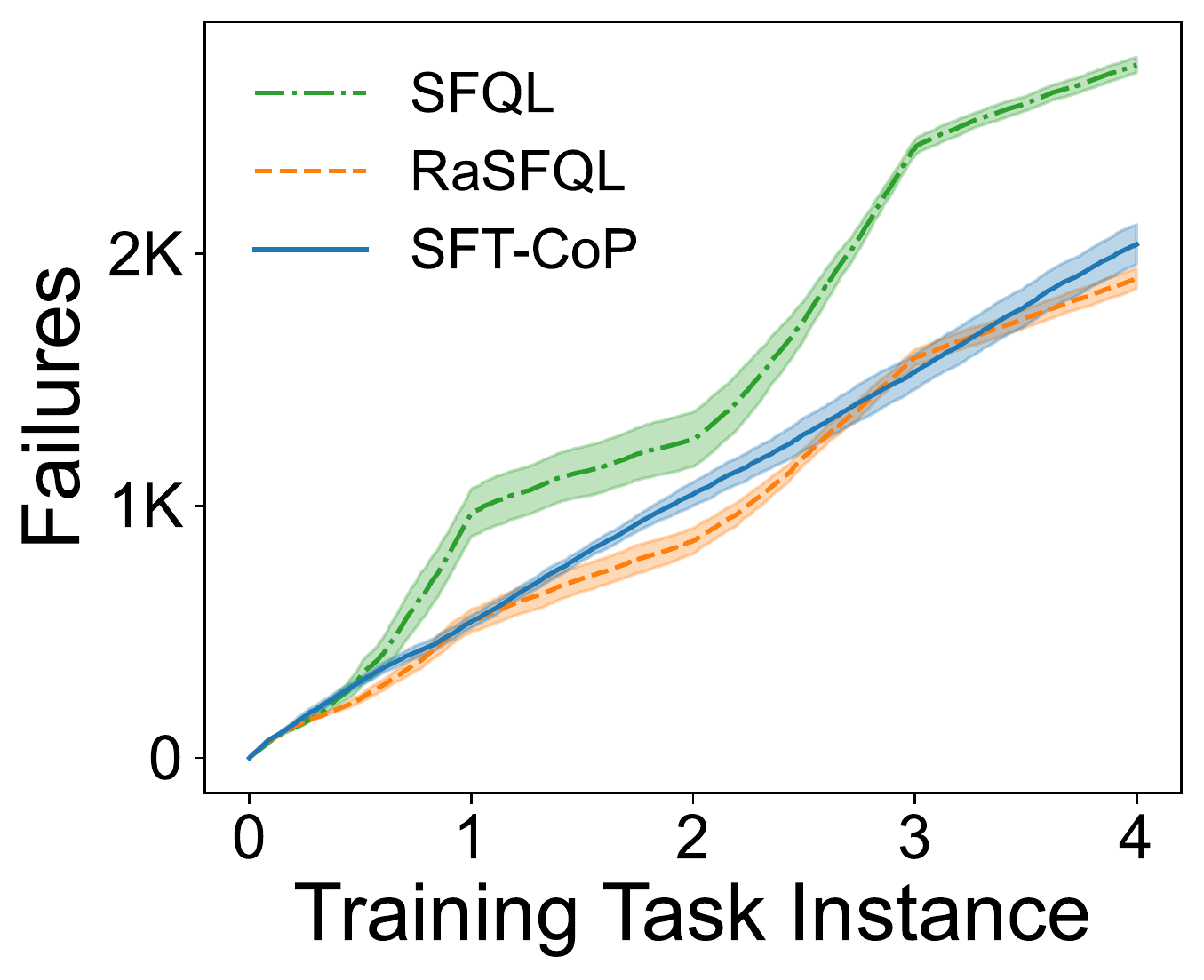}
		\caption{Accumulated failures.}
	\end{subfigure}
	\begin{subfigure}[t]{.24\textwidth}
	    \includegraphics[width=1.\textwidth]{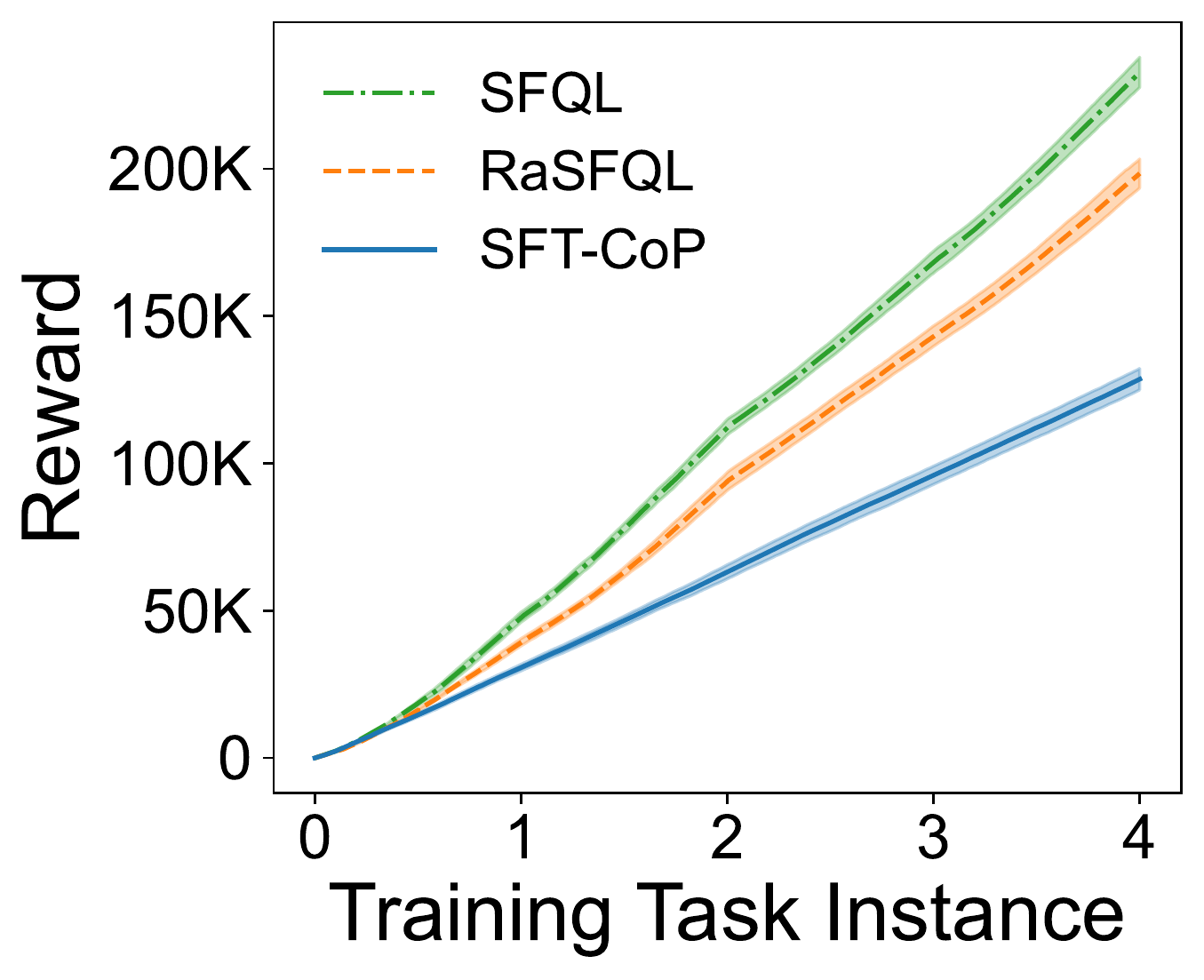}
		\caption{Accumulated rewards.}
	\end{subfigure}
	\begin{subfigure}[t]{.24\textwidth}
		\includegraphics[width=1.\textwidth]{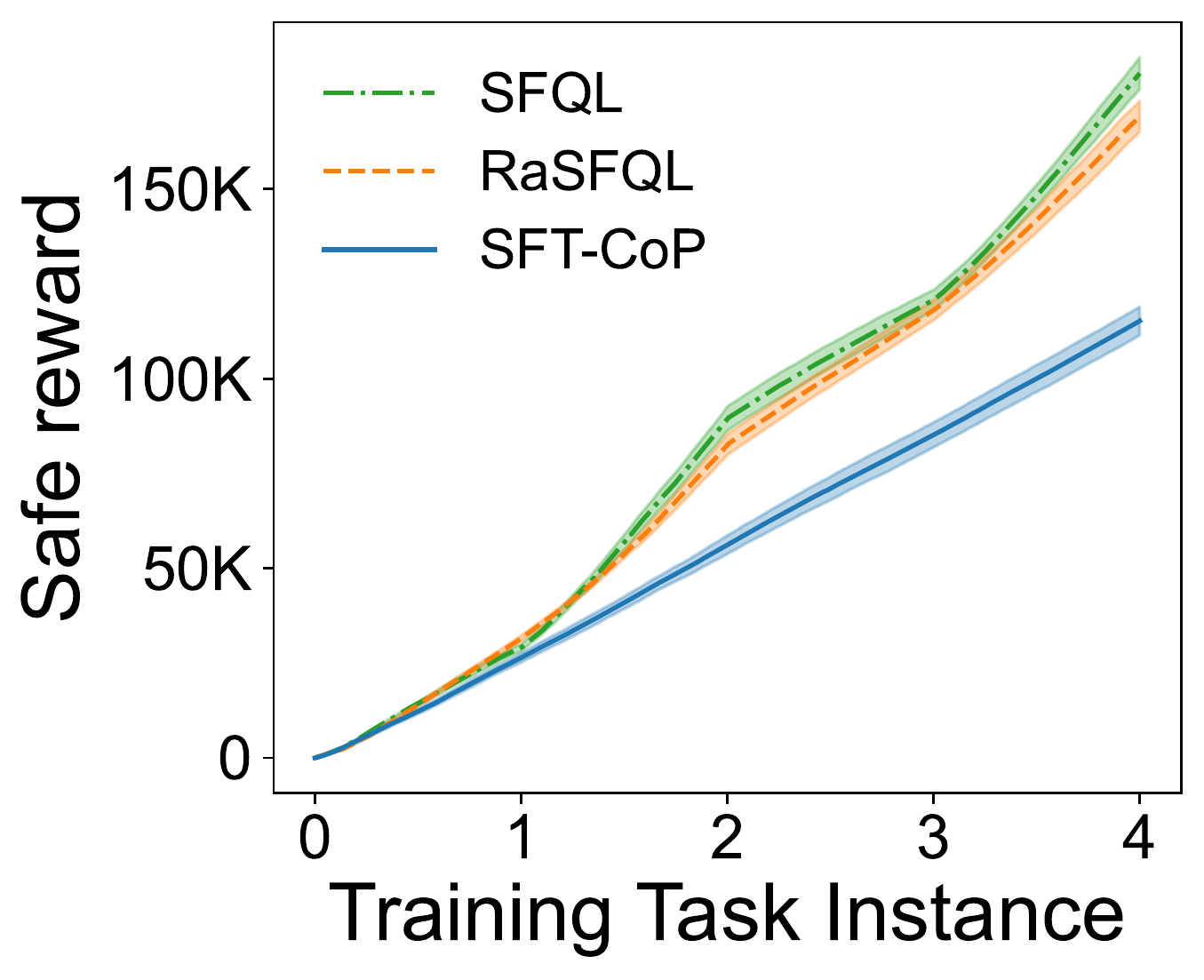}
		\caption{Accumulated rewards from safe regions.}
	\end{subfigure}
	\begin{subfigure}[t]{.24\textwidth}
	    \includegraphics[width=1.\textwidth]{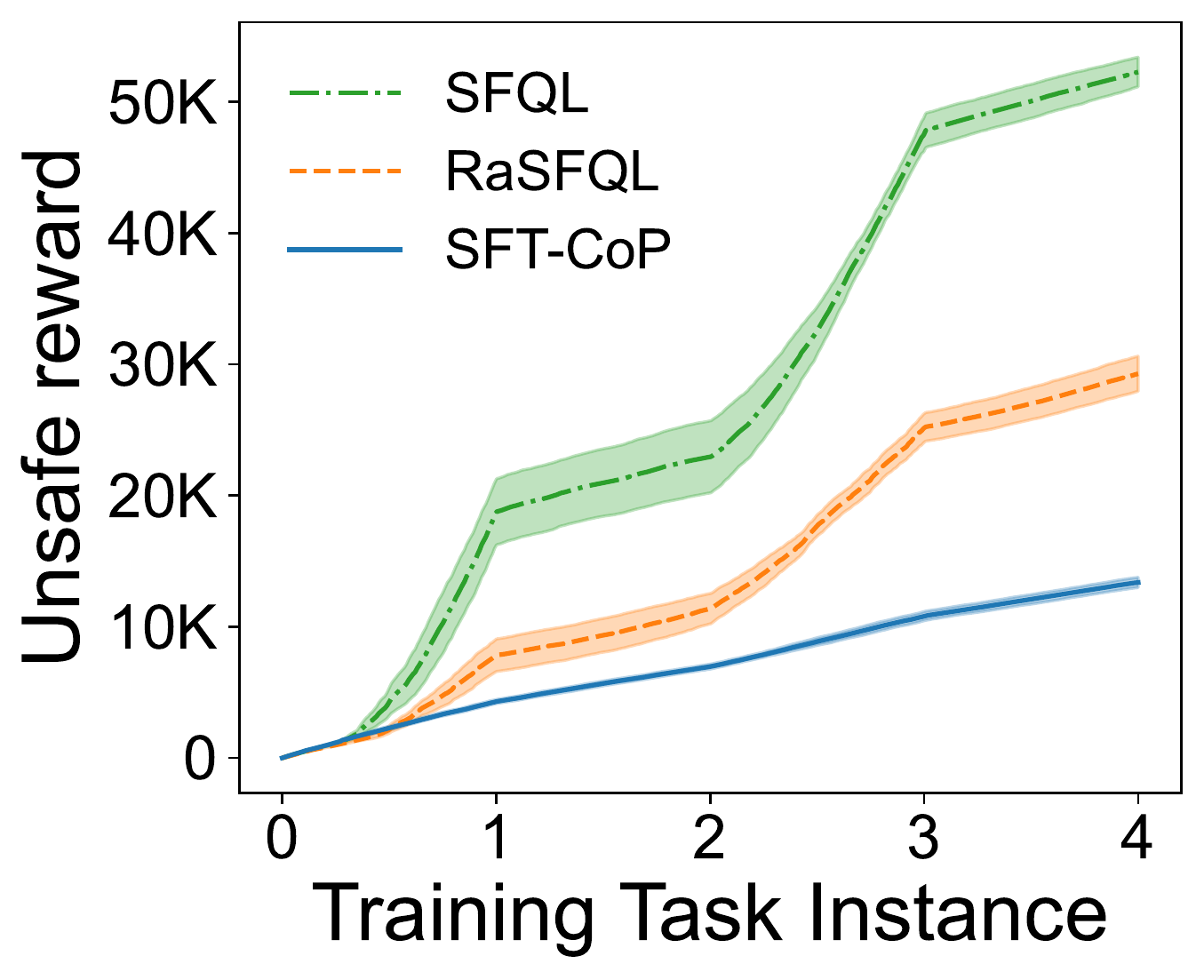}
		\caption{Accumulated rewards from unsafe regions.}
	\end{subfigure}
	\begin{subfigure}[t]{.24\textwidth}
		\includegraphics[width=1.\textwidth]{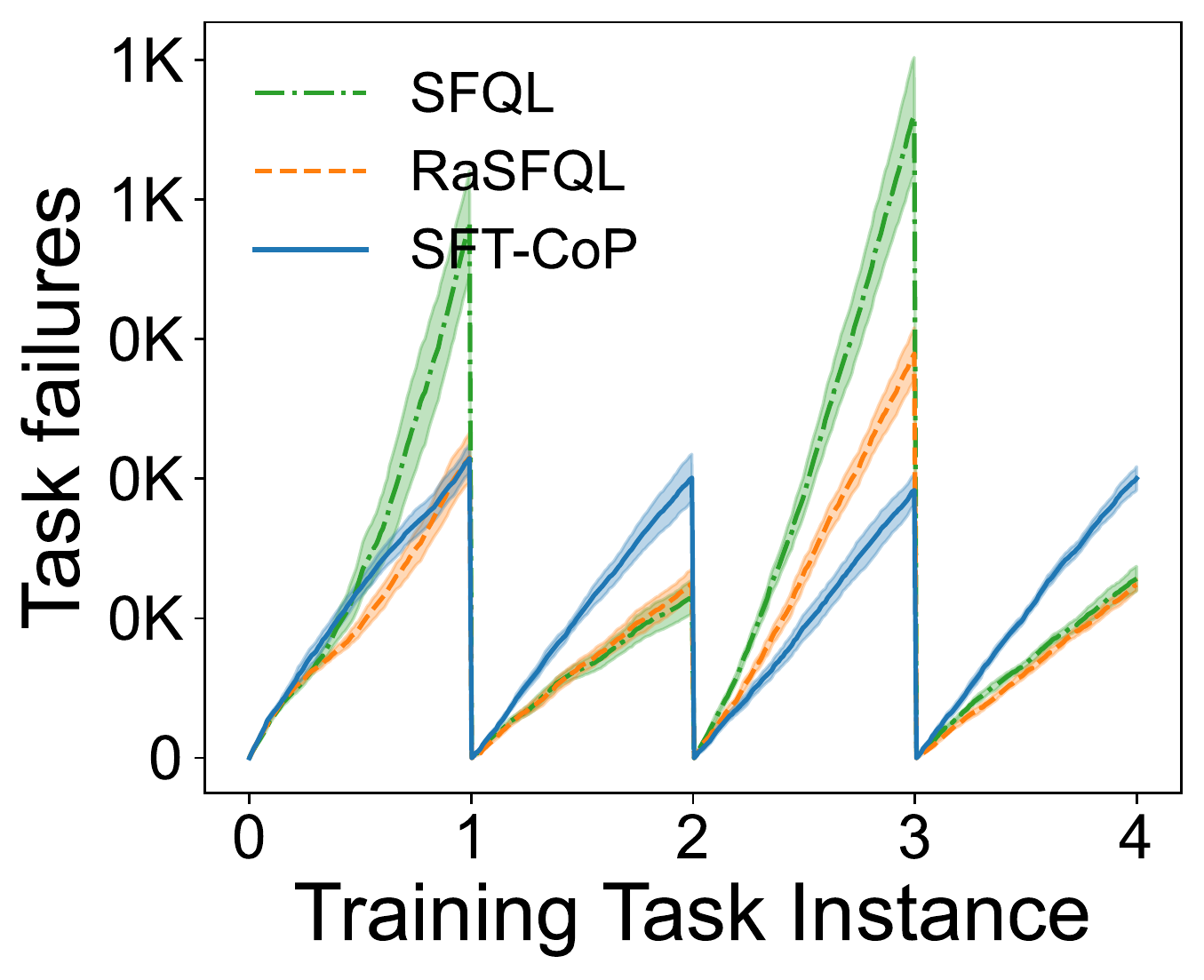}
		\caption{Task failures.}
	\end{subfigure}
	\begin{subfigure}[t]{.24\textwidth}
		\includegraphics[width=1.\textwidth]{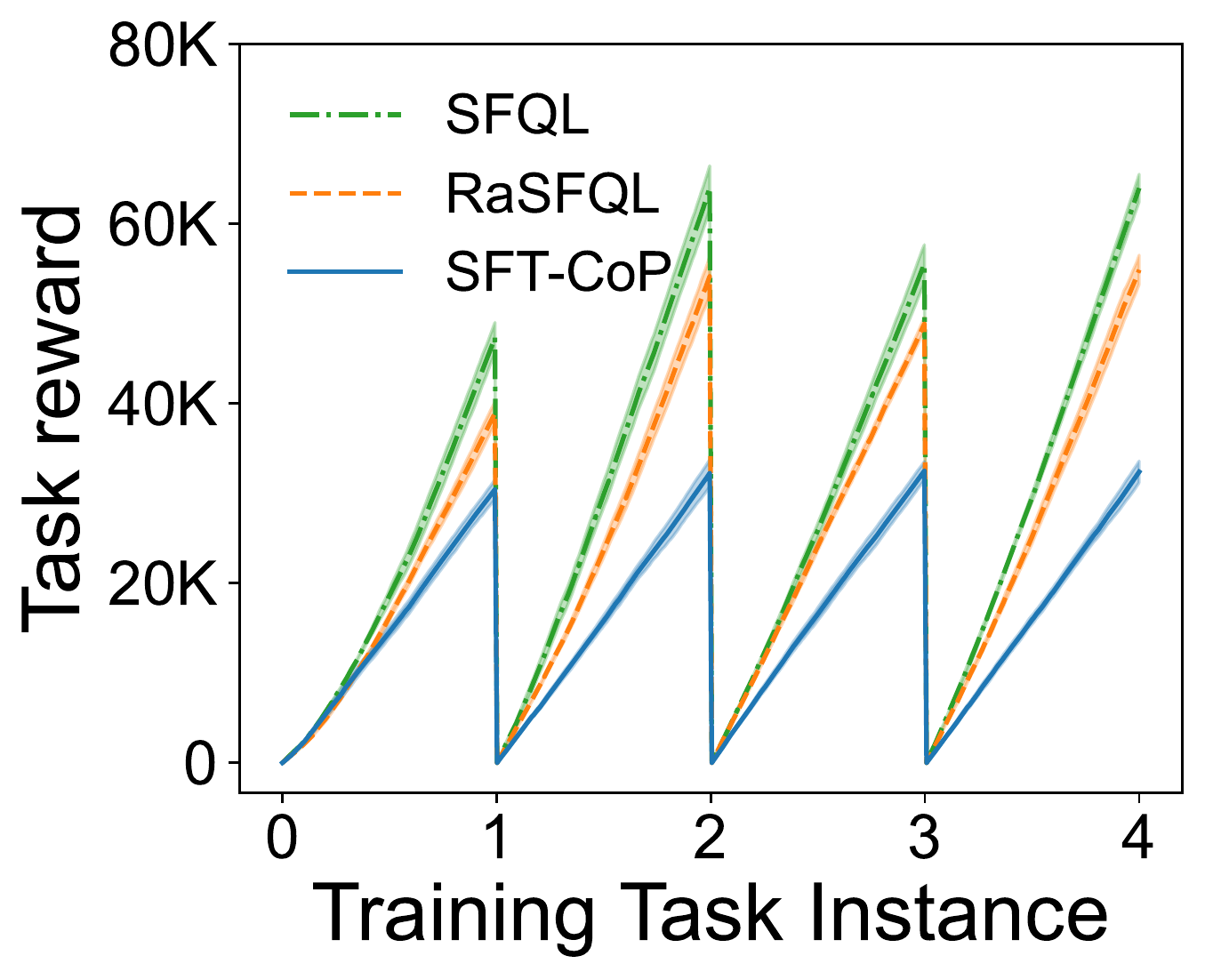}
		\caption{Task rewards.}
	\end{subfigure}
	\begin{subfigure}[t]{.24\textwidth}
		\includegraphics[width=1.\textwidth]{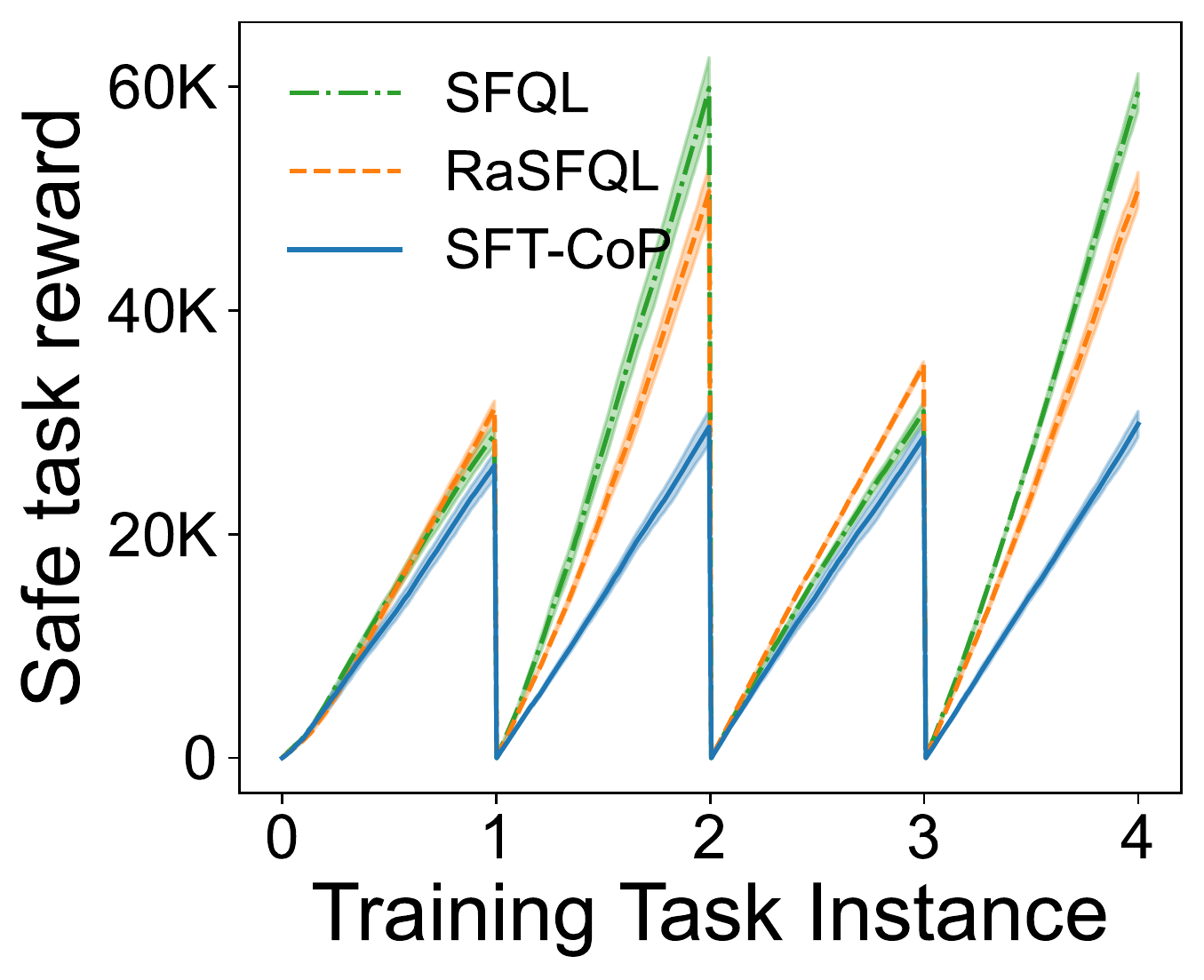}
		\caption{Task rewards from safe regions.}
	\end{subfigure}
	\begin{subfigure}[t]{.24\textwidth}
		\includegraphics[width=1.\textwidth]{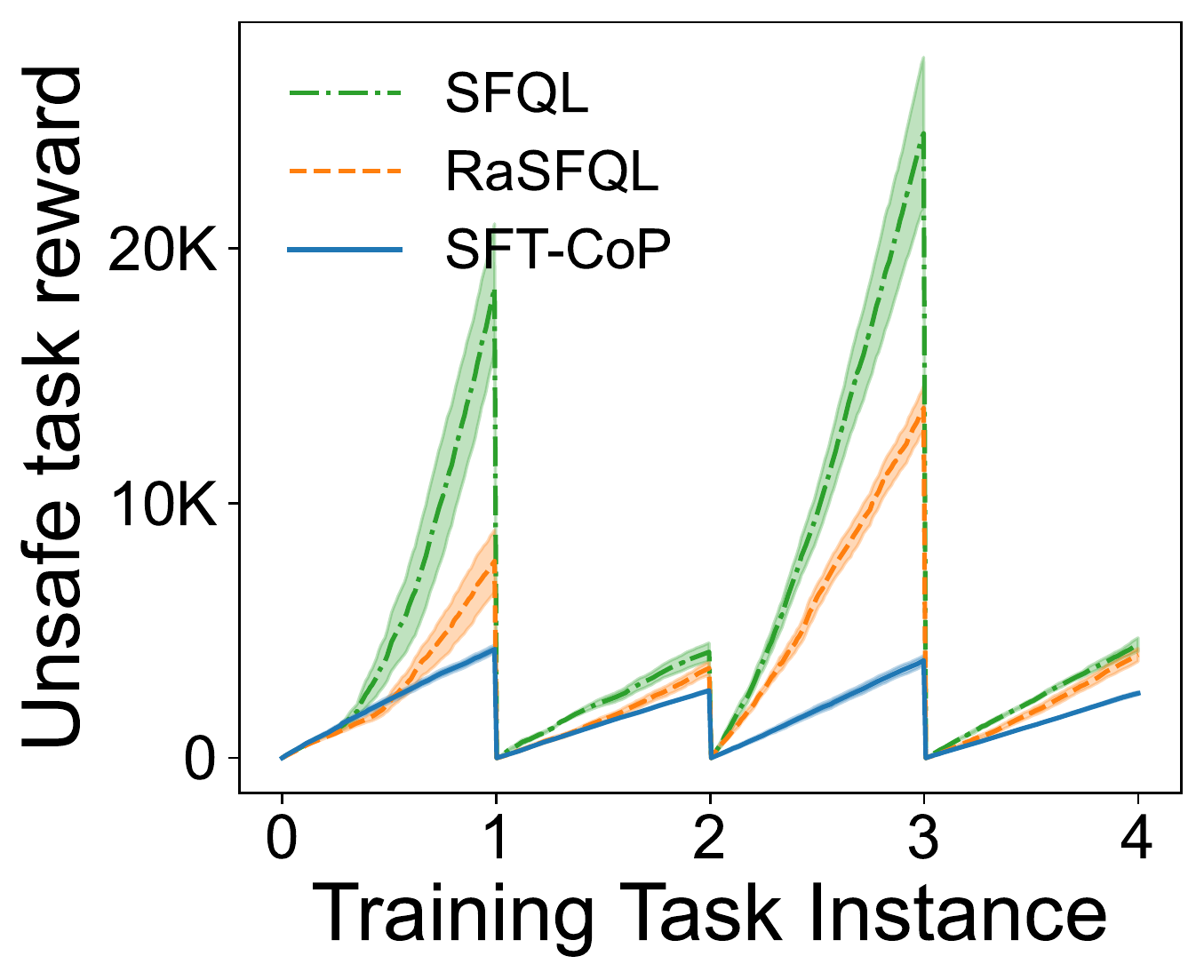}
		\caption{Task rewards from unsafe regions.}
	\end{subfigure}
	\caption{Performance of \sfql, \rasfql{} ($\beta=2$), and \method{} on the Reacher domain with probabilistic unsafe regions. We report accumulated and per-task (a, e) failures, (b, f) total rewards, (c, g) rewards from safe regions and (d, h) rewards from unsafe regions, over the training tasks.}
	\label{fig:reacher_lphc}
\end{figure*}

Fig.~\ref{fig:4room_lphc} and Fig.~\ref{fig:reacher_lphc} compare \method{} with \rasfql{} in the uncertain scenario on Four-Room and Reacher using the same risk configuration as in~\cite{gimelfarb2021riskaware}, where upon stepping into a trap or unsafe region, the agent only receives costs when they become activated (with a probability of $5$\% and $3.5$\% for these two environments, respectively). Results show that \method{} provides a similar level of safety as \rasfql{}.

\subsection{Additional results}
\begin{figure*}
    \centering
	\begin{subfigure}[t]{.24\textwidth}
		\includegraphics[width=1.\textwidth]{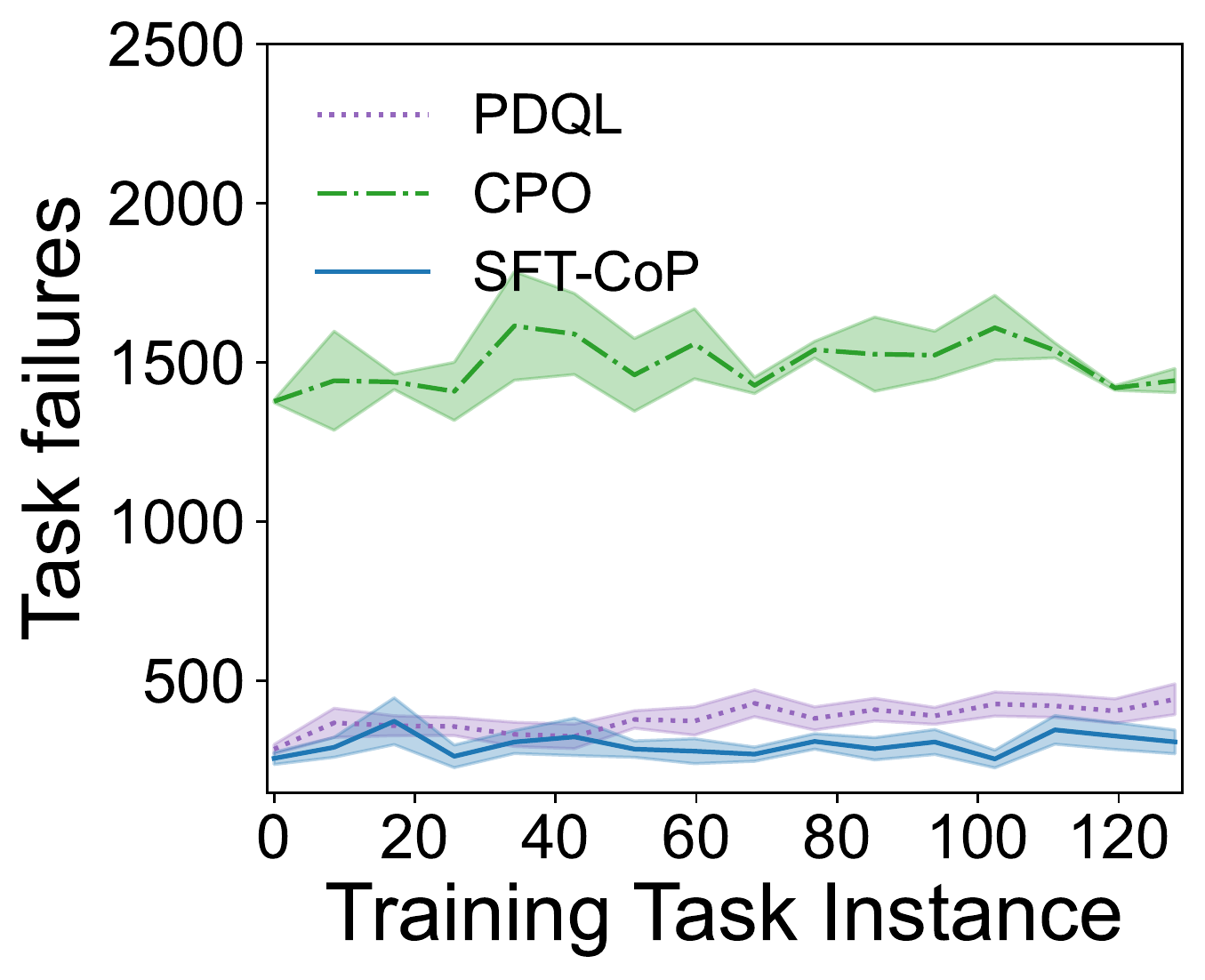}
		\caption{Task failures.}
	\end{subfigure}
	\begin{subfigure}[t]{.24\textwidth}
	    \includegraphics[width=1.\textwidth]{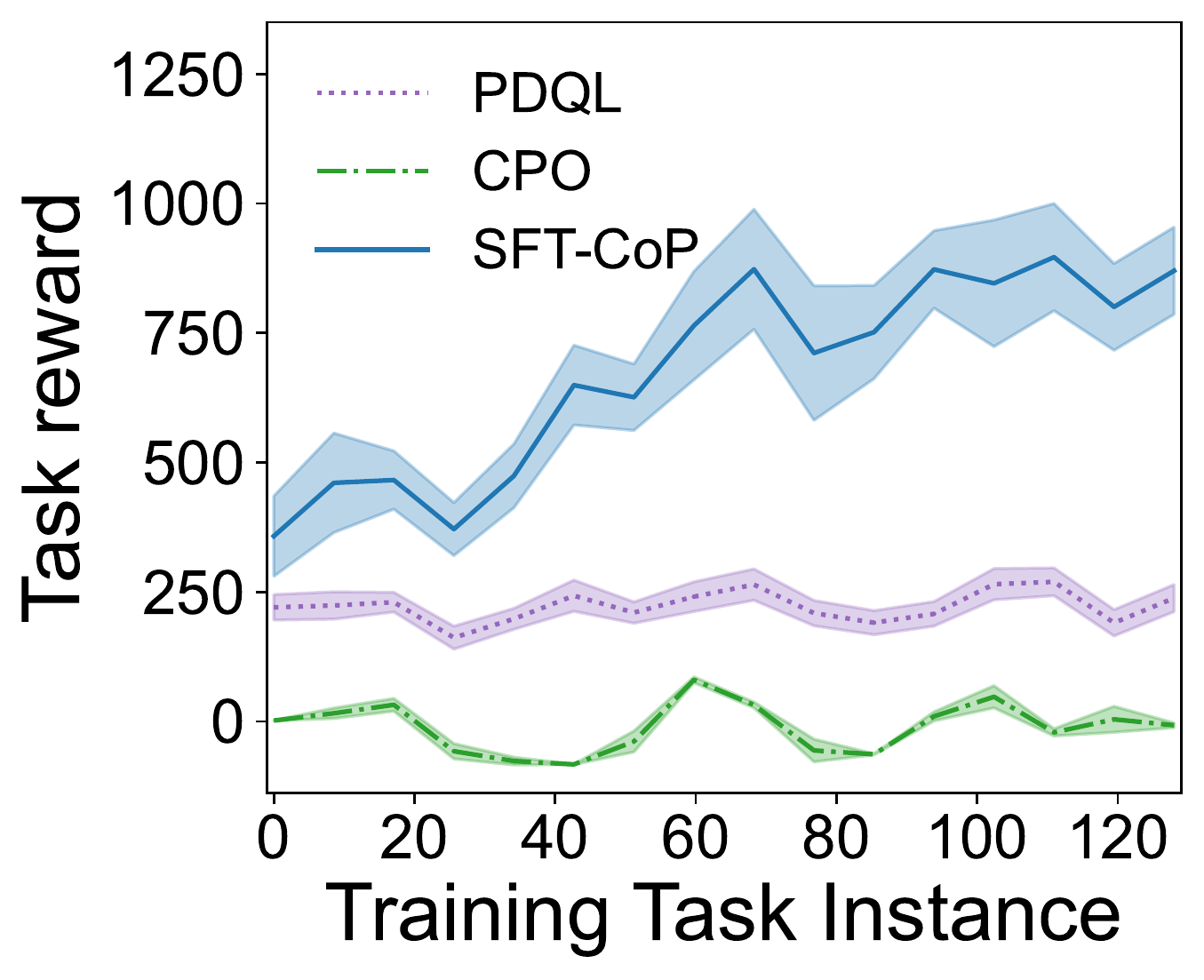}
		\caption{Task rewards.}
	\end{subfigure}
	\begin{subfigure}[t]{.24\textwidth}
		\includegraphics[width=1.\textwidth]{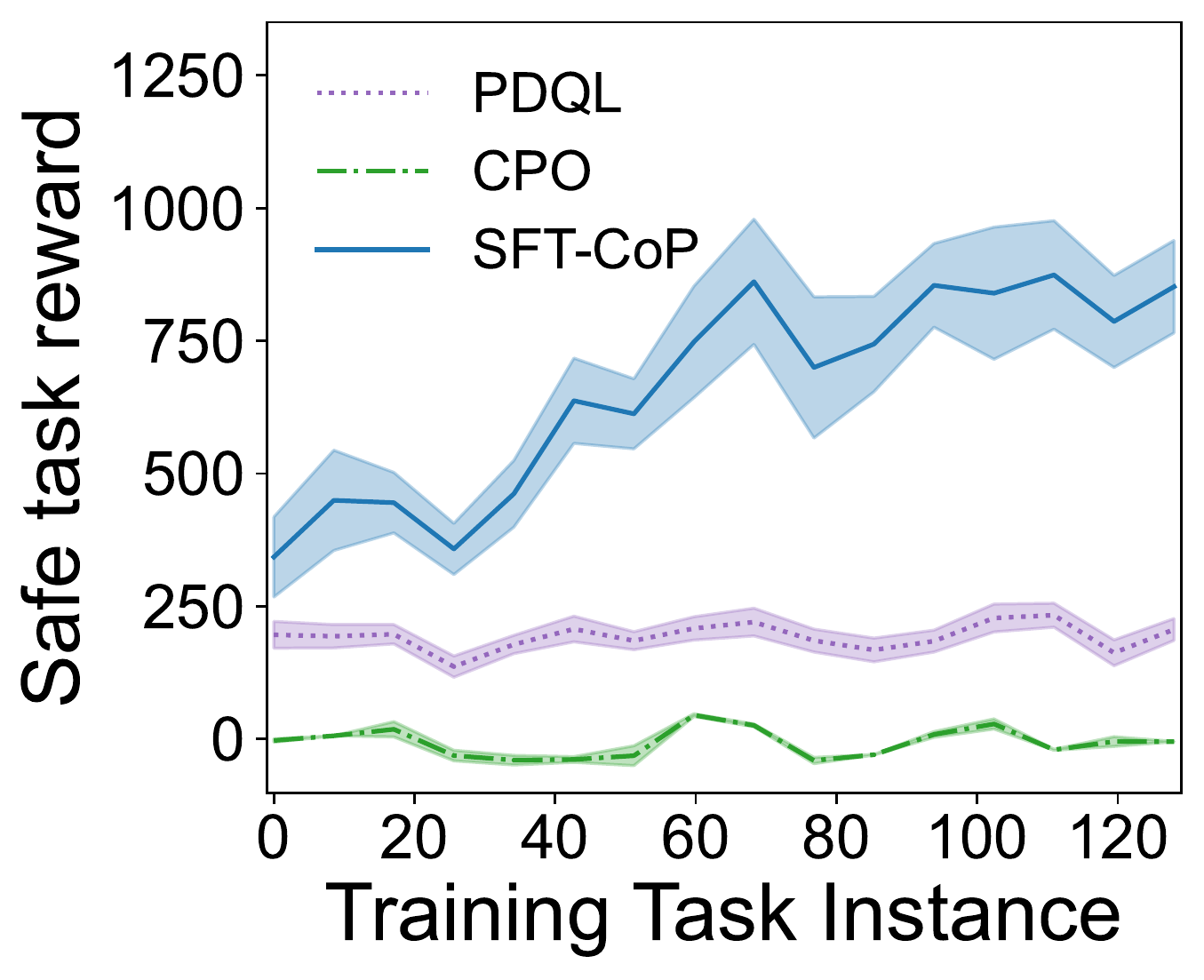}
		\caption{Task rewards from safe objects.}
	\end{subfigure}
	\begin{subfigure}[t]{.24\textwidth}
	    \includegraphics[width=1.\textwidth]{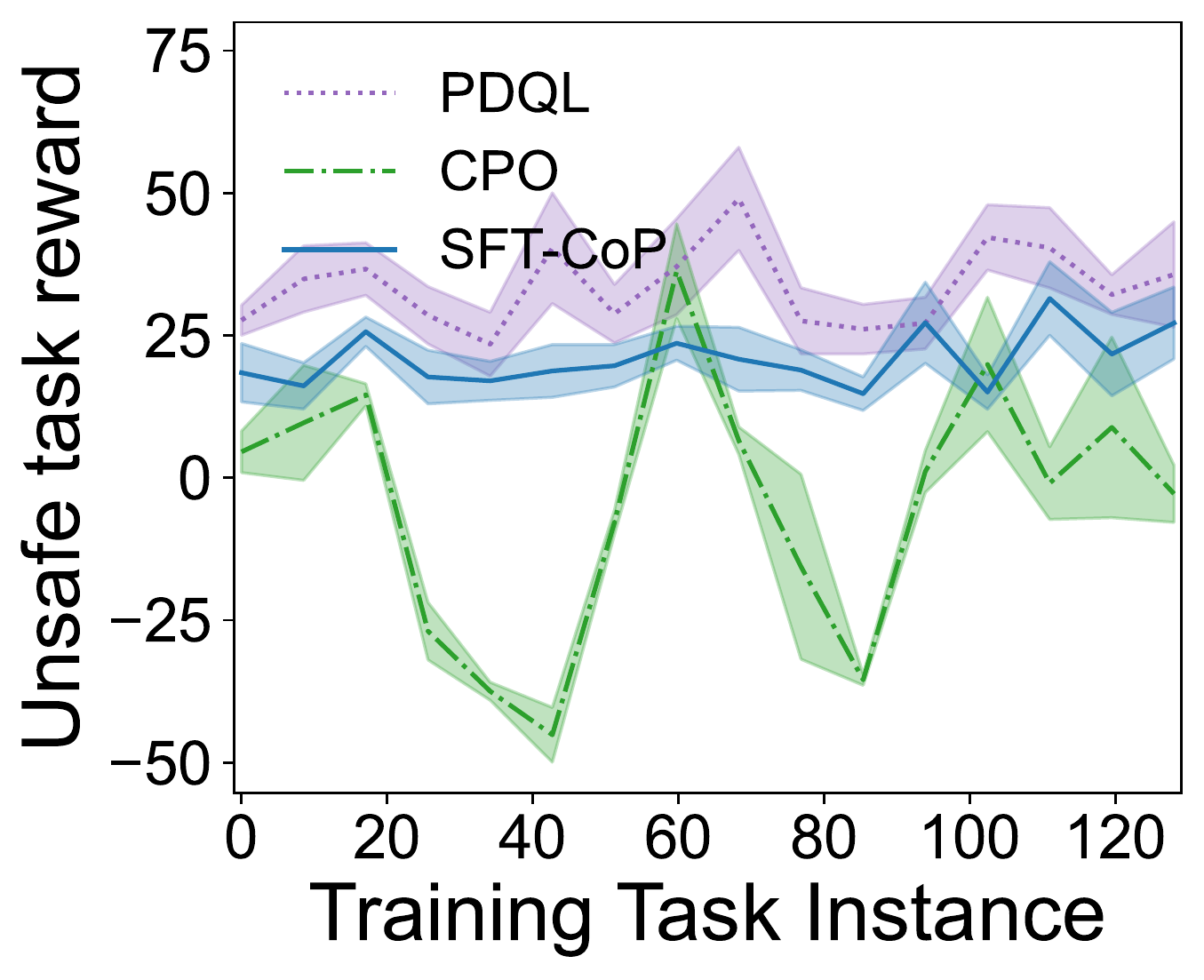}
		\caption{Task rewards from unsafe objects.}
	\end{subfigure}
	\begin{subfigure}[t]{.24\textwidth}
		\includegraphics[width=1.\textwidth]{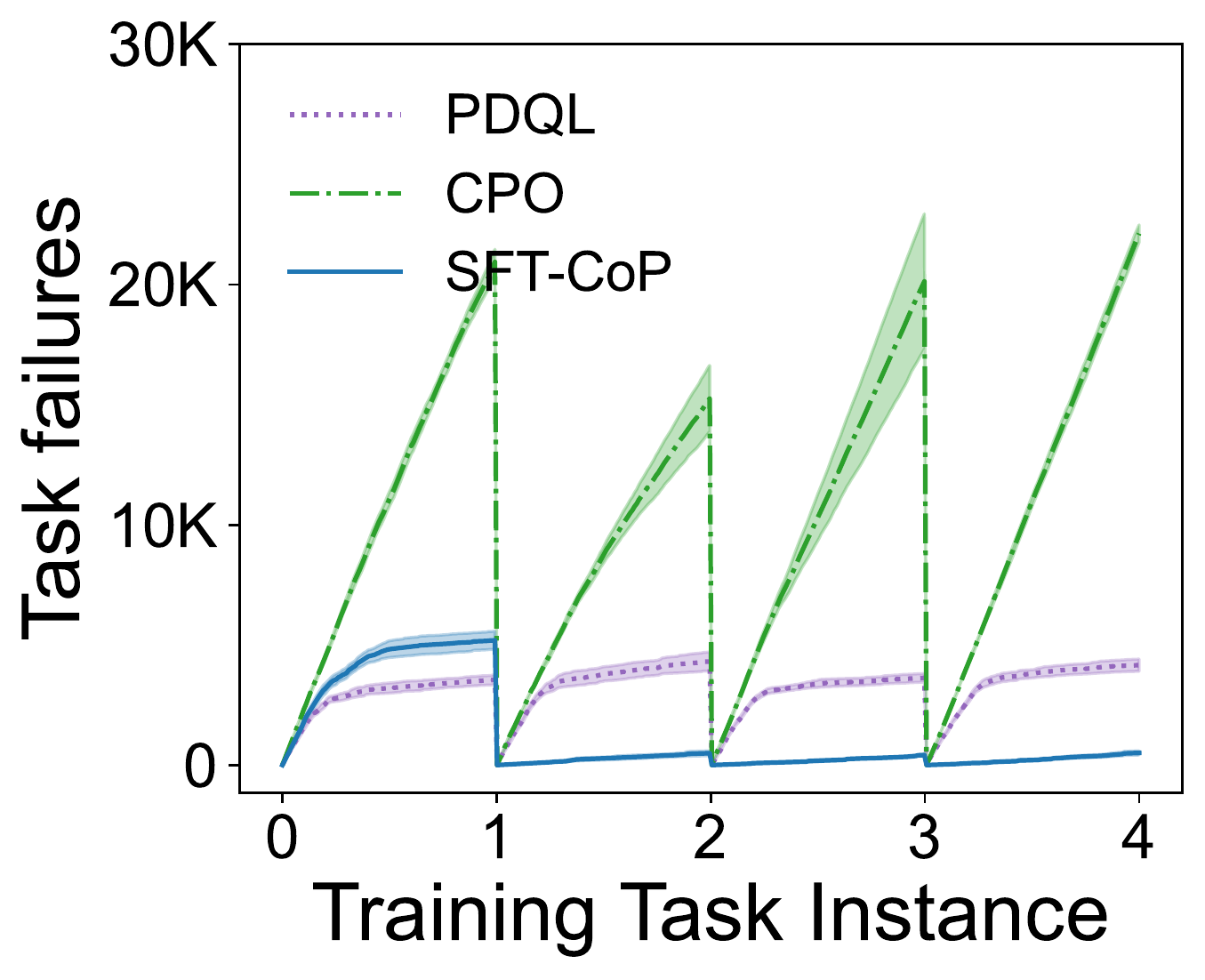}
		\caption{Task failures.}
	\end{subfigure}
	\begin{subfigure}[t]{.24\textwidth}
	    \includegraphics[width=1.\textwidth]{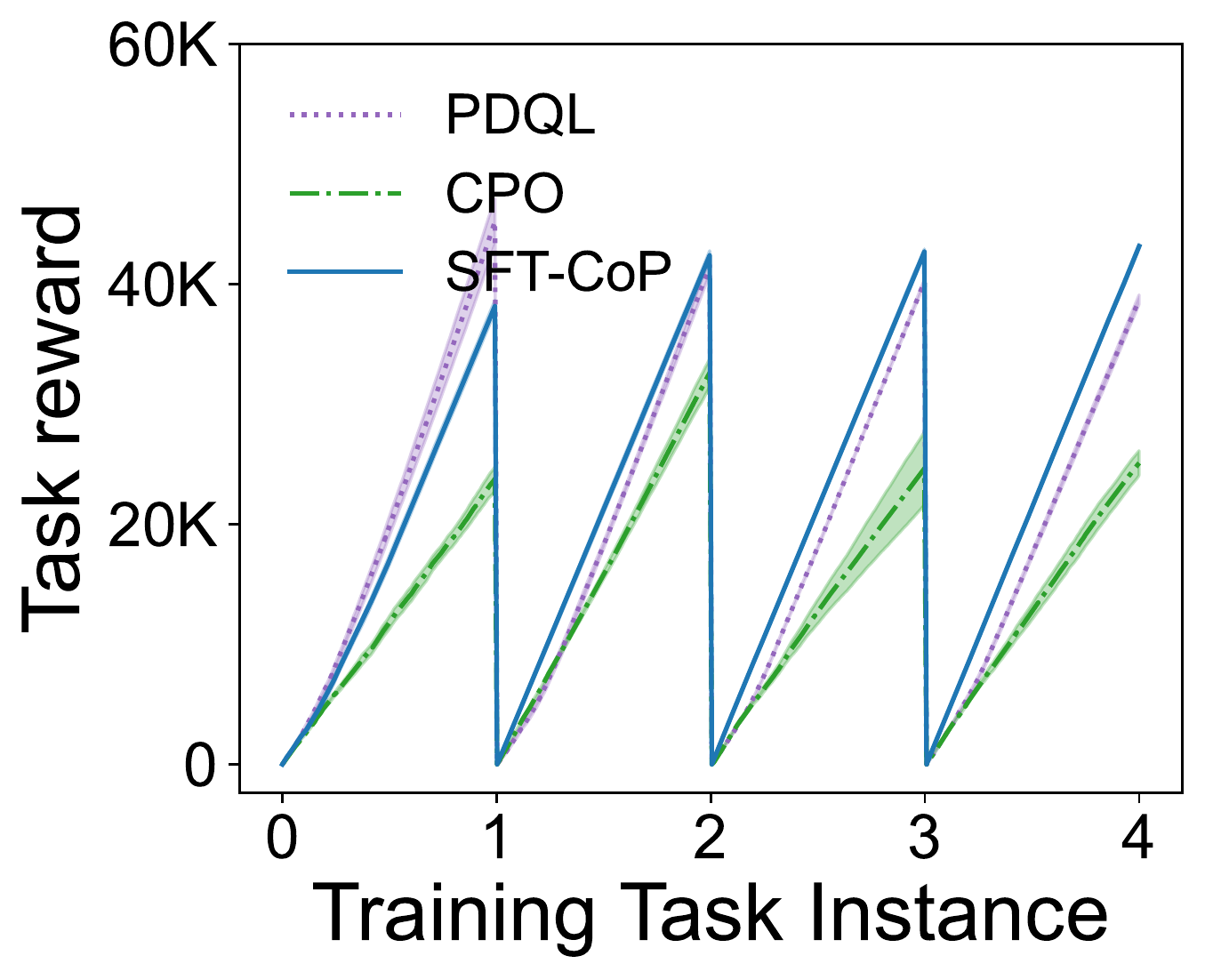}
		\caption{Task rewards.}
	\end{subfigure}
	\begin{subfigure}[t]{.24\textwidth}
		\includegraphics[width=1.\textwidth]{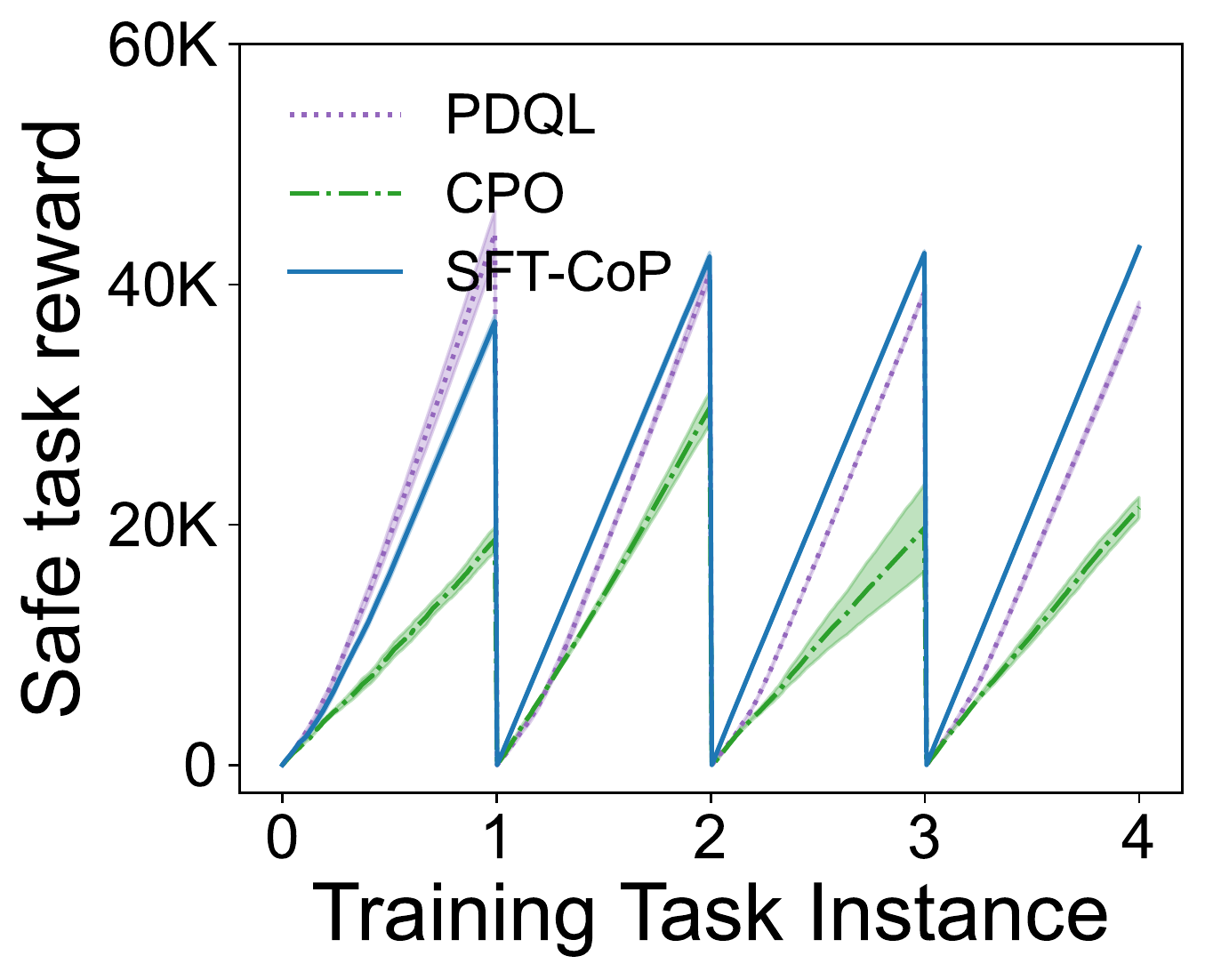}
		\caption{Task rewards from safe regions.}
	\end{subfigure}
	\begin{subfigure}[t]{.24\textwidth}
		\includegraphics[width=1.\textwidth]{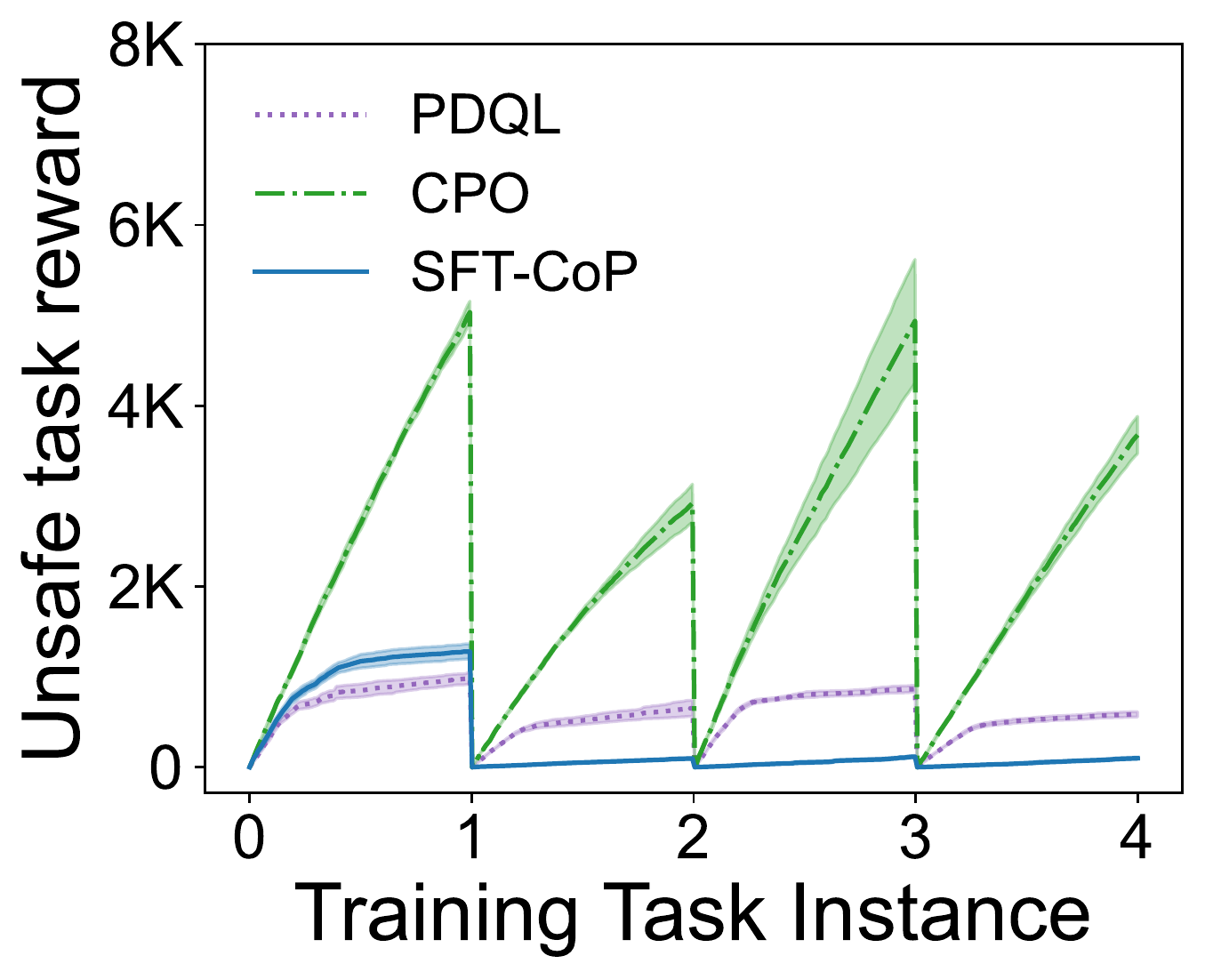}
		\caption{Task rewards from unsafe regions.}
	\end{subfigure}
	\caption{Performance of \pdql{} and \method{} on the Four-Room (top row) and Reacher (bottom row) domains. We report per-task (a, e) failures, (b, f) total rewards, (c, g) rewards from safe regions and (d, h) rewards from unsafe regions, over the training tasks.}
	\label{fig:non-transfer}
\end{figure*}
\begin{figure*}
    \centering
	\begin{subfigure}[t]{.24\textwidth}
		\includegraphics[width=1\textwidth]{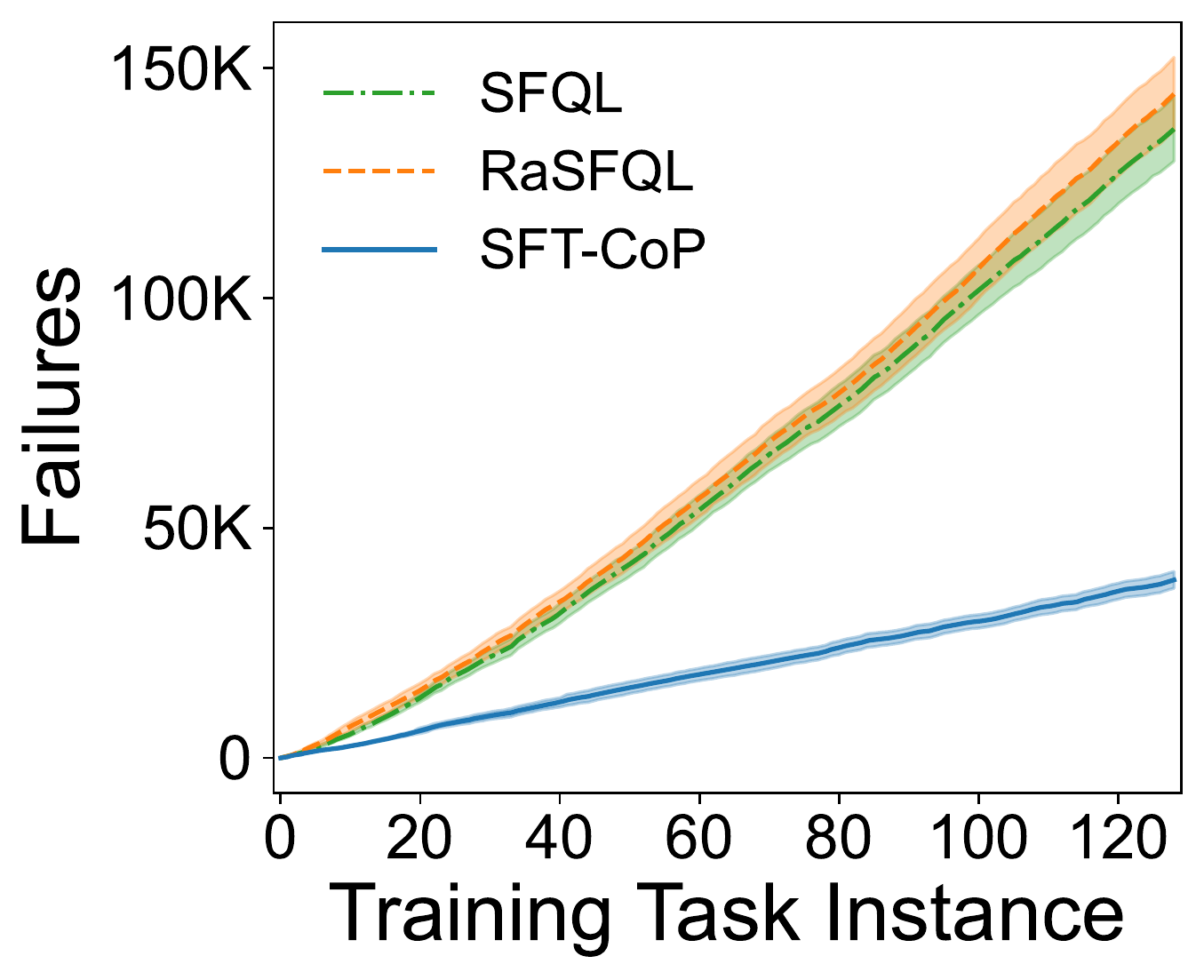}
		\caption{Accumulated failures.}
	\end{subfigure}
	\begin{subfigure}[t]{.24\textwidth}
	    \includegraphics[width=1\textwidth]{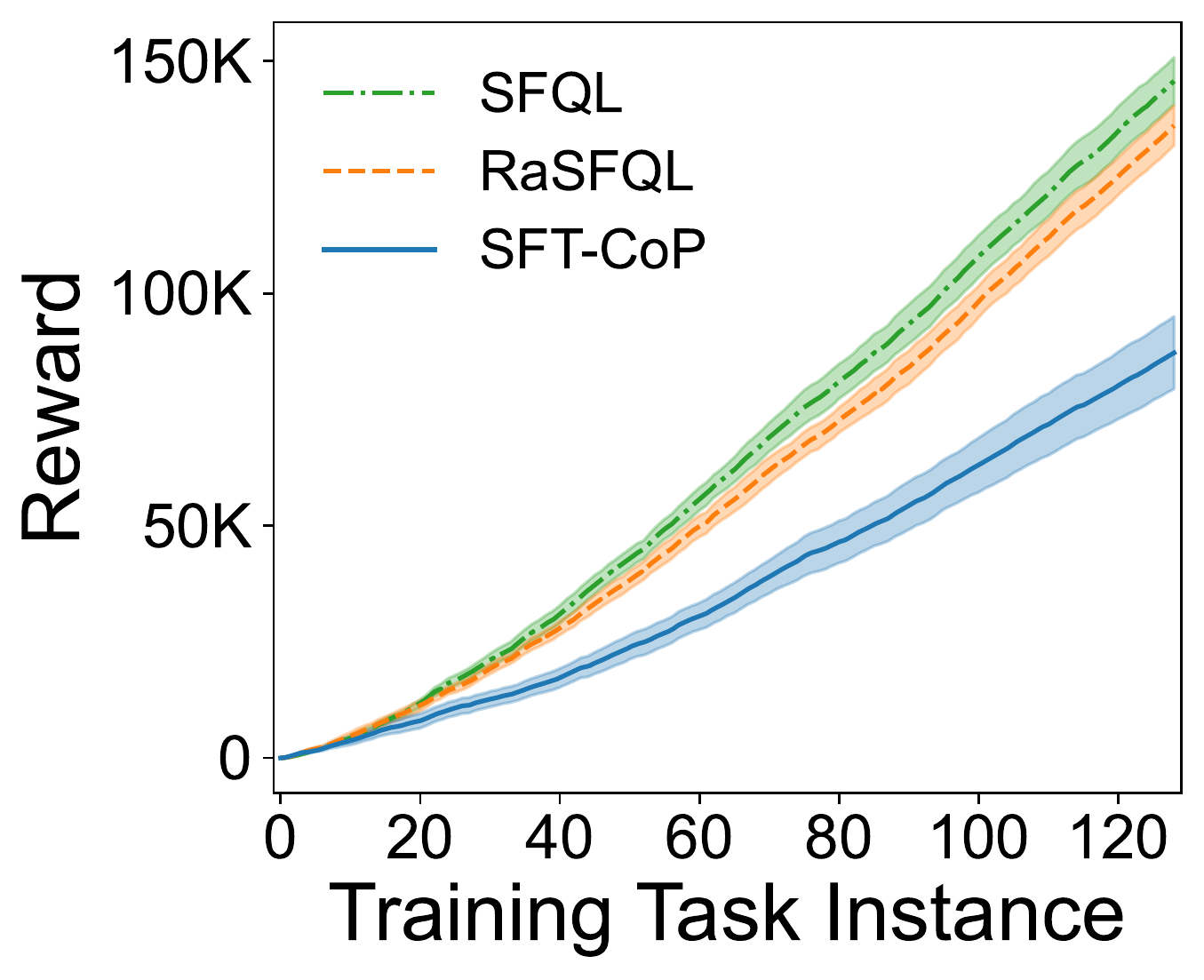}
		\caption{Accumulated rewards.}
	\end{subfigure}
	\begin{subfigure}[t]{.24\textwidth}
		\includegraphics[width=1\textwidth]{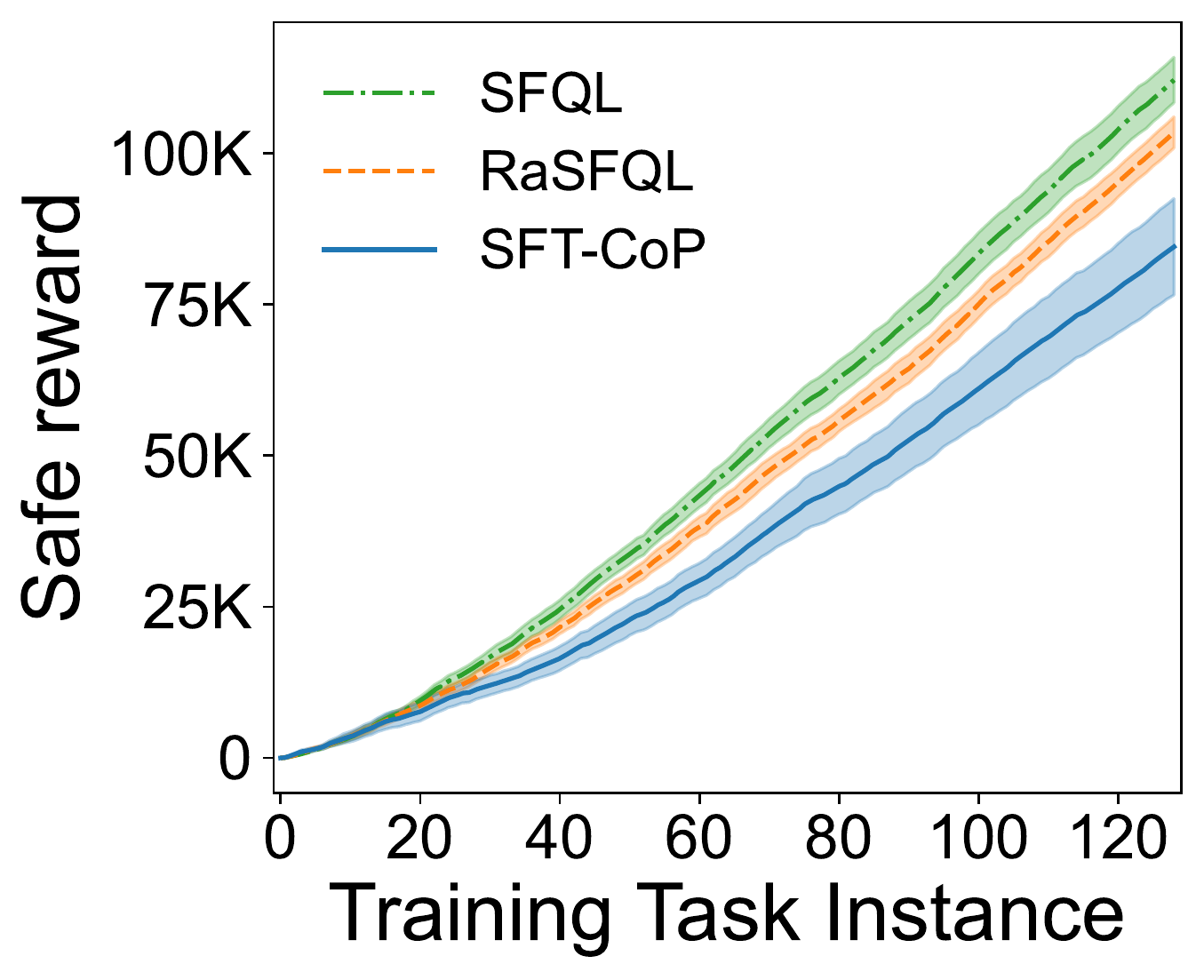}
		\caption{Accumulated rewards from safe objects.}
	\end{subfigure}
	\begin{subfigure}[t]{.24\textwidth}
		\includegraphics[width=1\textwidth]{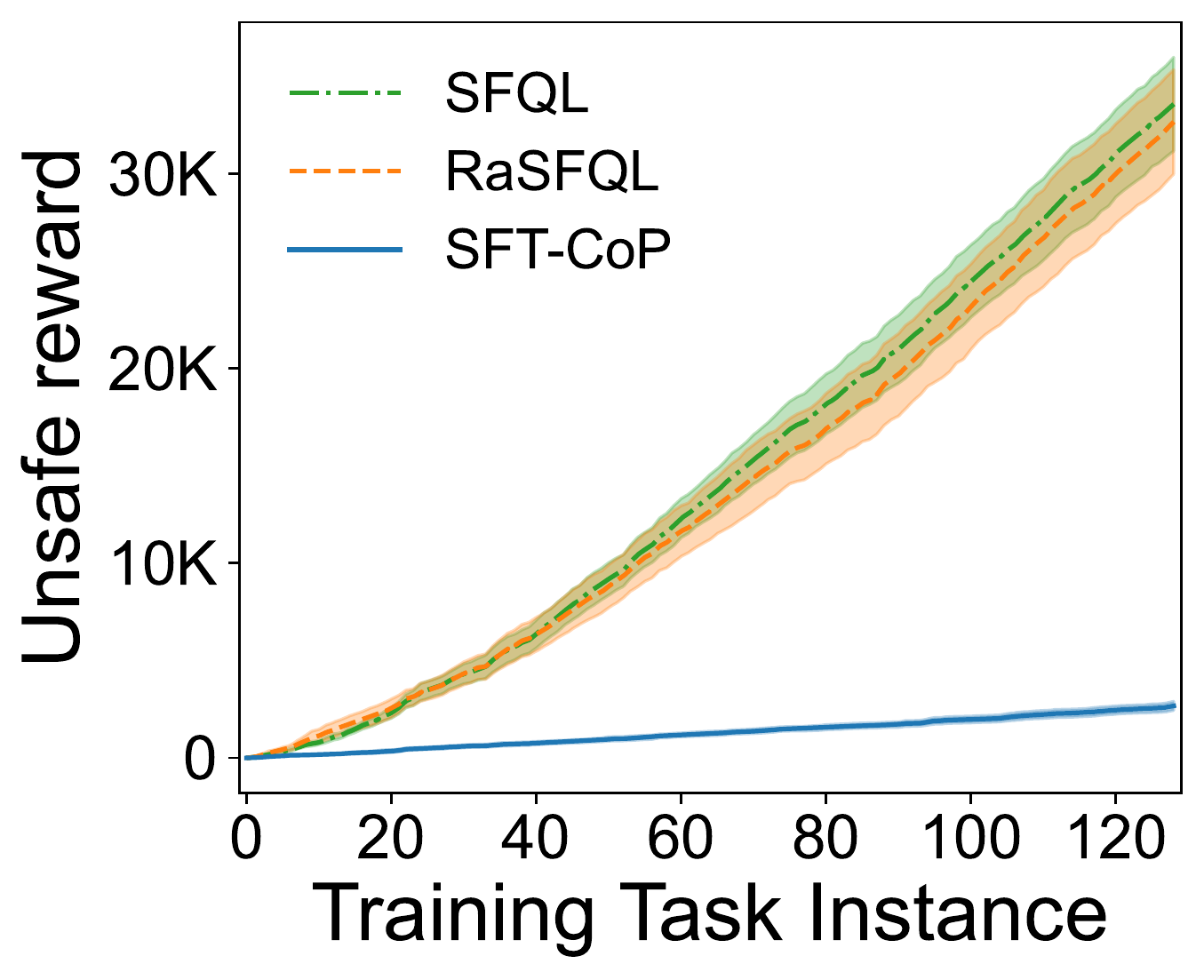}
		\caption{Accumulated rewards from unsafe objects.}
	\end{subfigure}
	\caption{Performance of \sfql, \rasfql{} ($\beta=2$) and \method{} on the Four-Room domain. We compute accumulated (a) failures, (b) total rewards, (c) rewards from safe objects and (d) rewards from unsafe objects, over the training task instances.}
	\label{fig:4room_hplc}
\end{figure*}
\begin{figure*}
    \centering
	\begin{subfigure}[t]{.24\textwidth}
		\includegraphics[width=1.\textwidth]{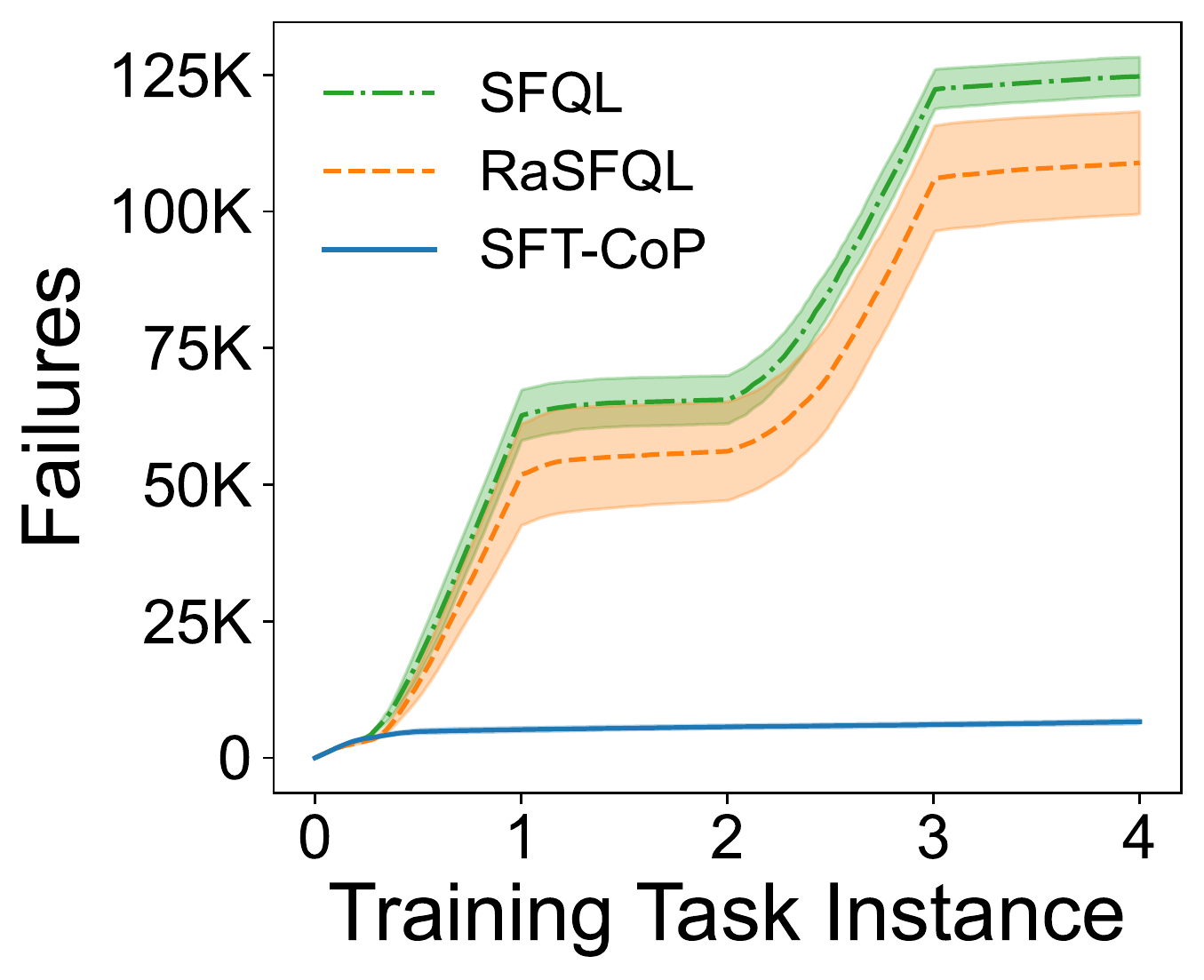}
		\caption{Accumulated failures.}
	\end{subfigure}
	\begin{subfigure}[t]{.24\textwidth}
	    \includegraphics[width=1.\textwidth]{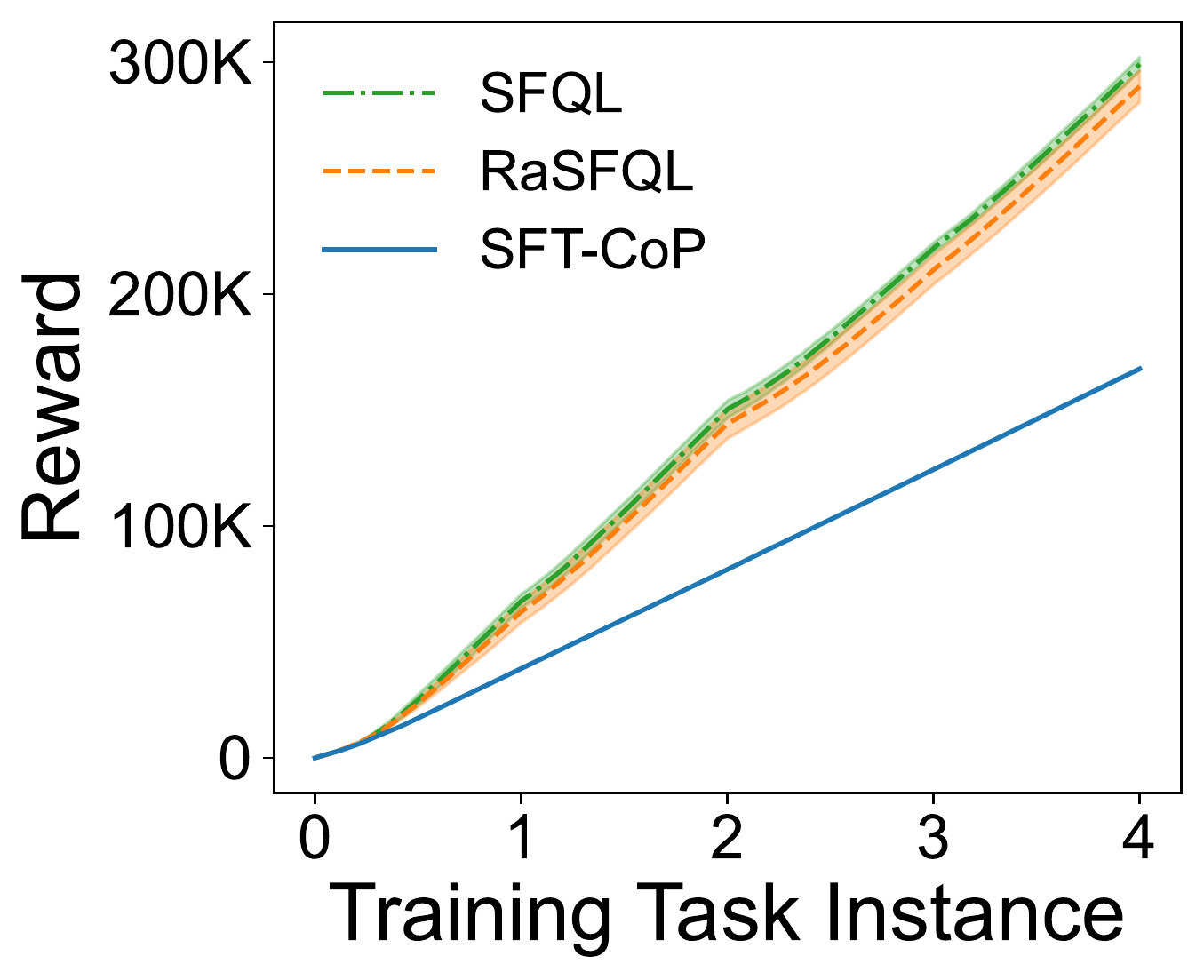}
		\caption{Accumulated rewards.}
	\end{subfigure}
	\begin{subfigure}[t]{.24\textwidth}
		\includegraphics[width=1.\textwidth]{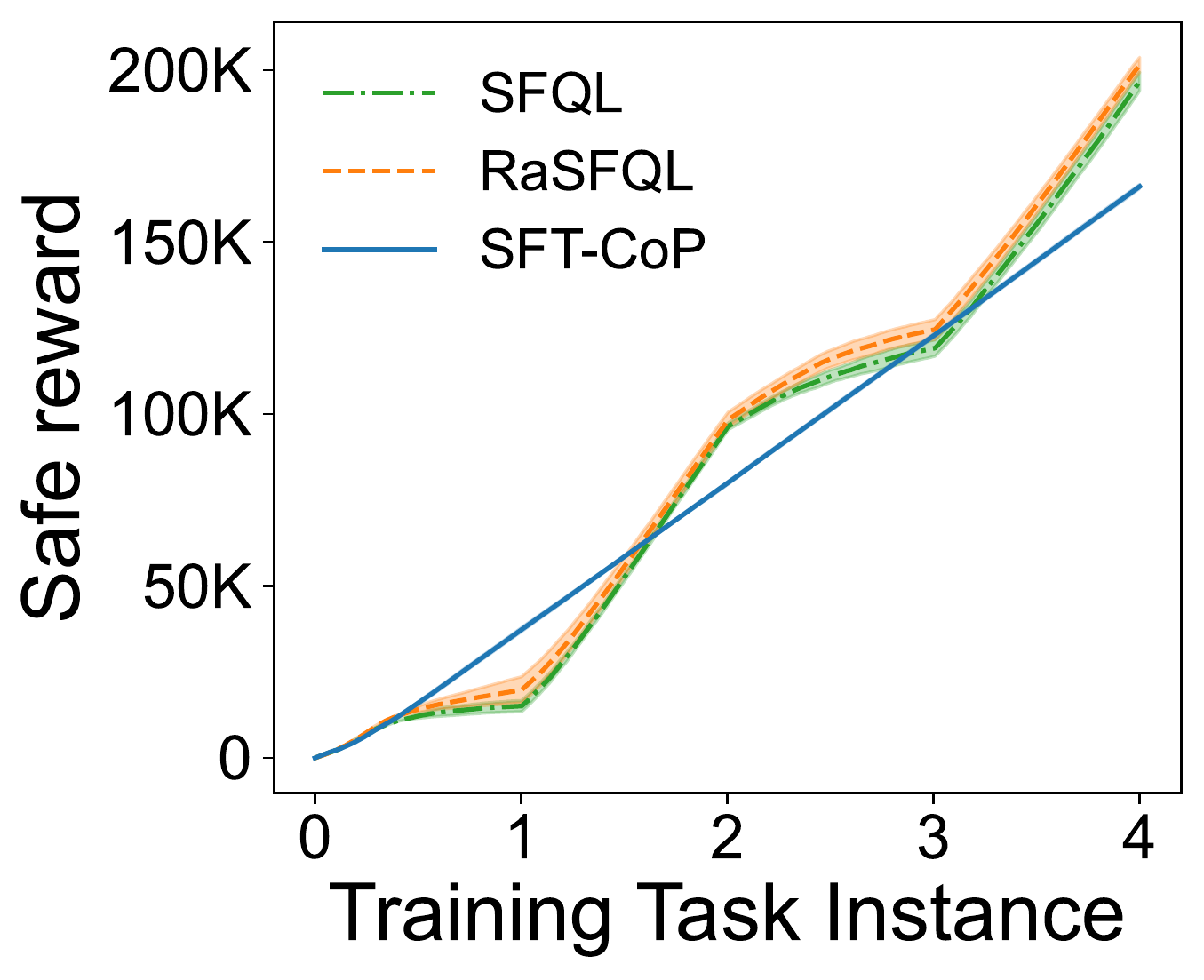}
		\caption{Accumulated rewards from safe regions.}
	\end{subfigure}
	\begin{subfigure}[t]{.24\textwidth}
	    \includegraphics[width=1.\textwidth]{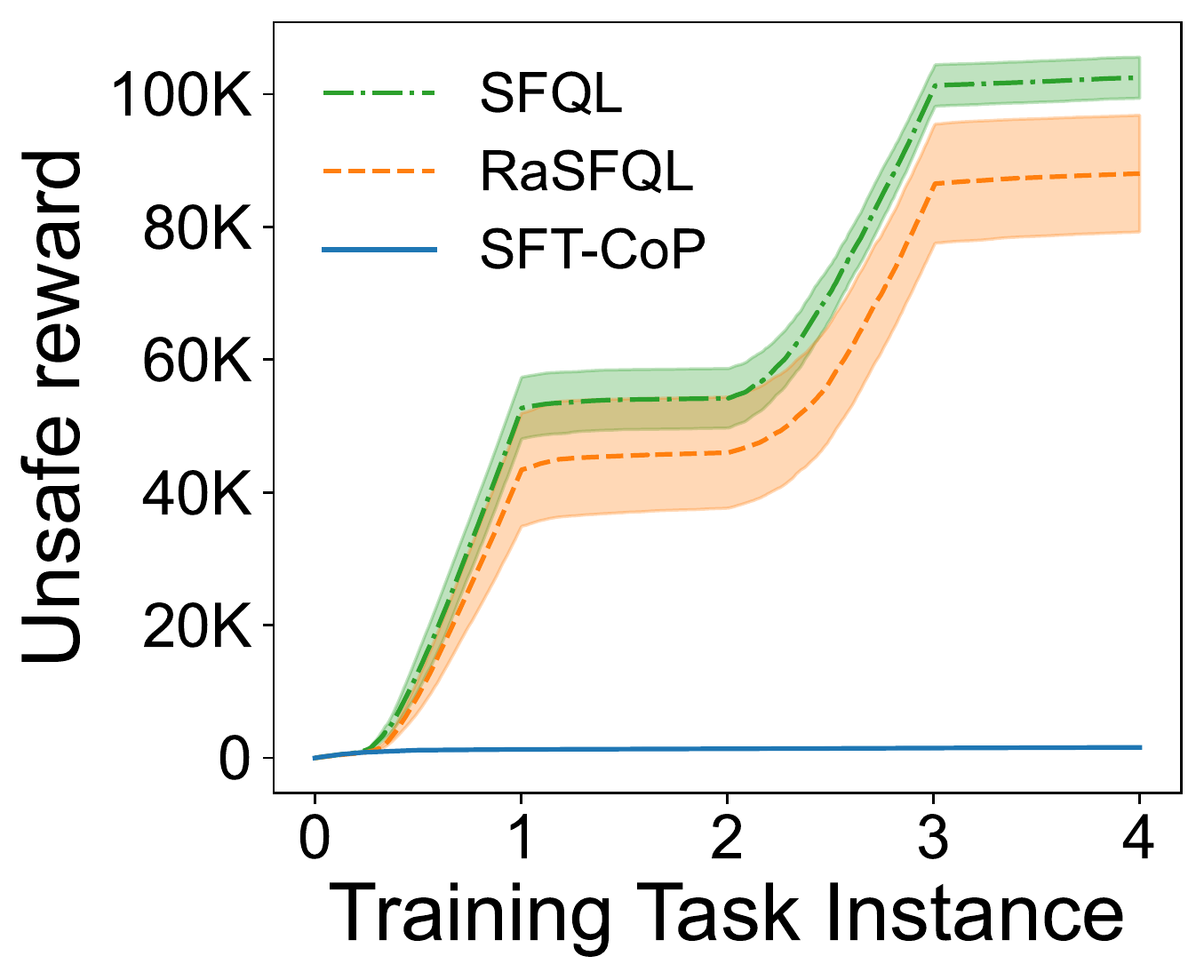}
		\caption{Accumulated rewards from unsafe regions.}
	\end{subfigure}
	\begin{subfigure}[t]{.24\textwidth}
		\includegraphics[width=1.\textwidth]{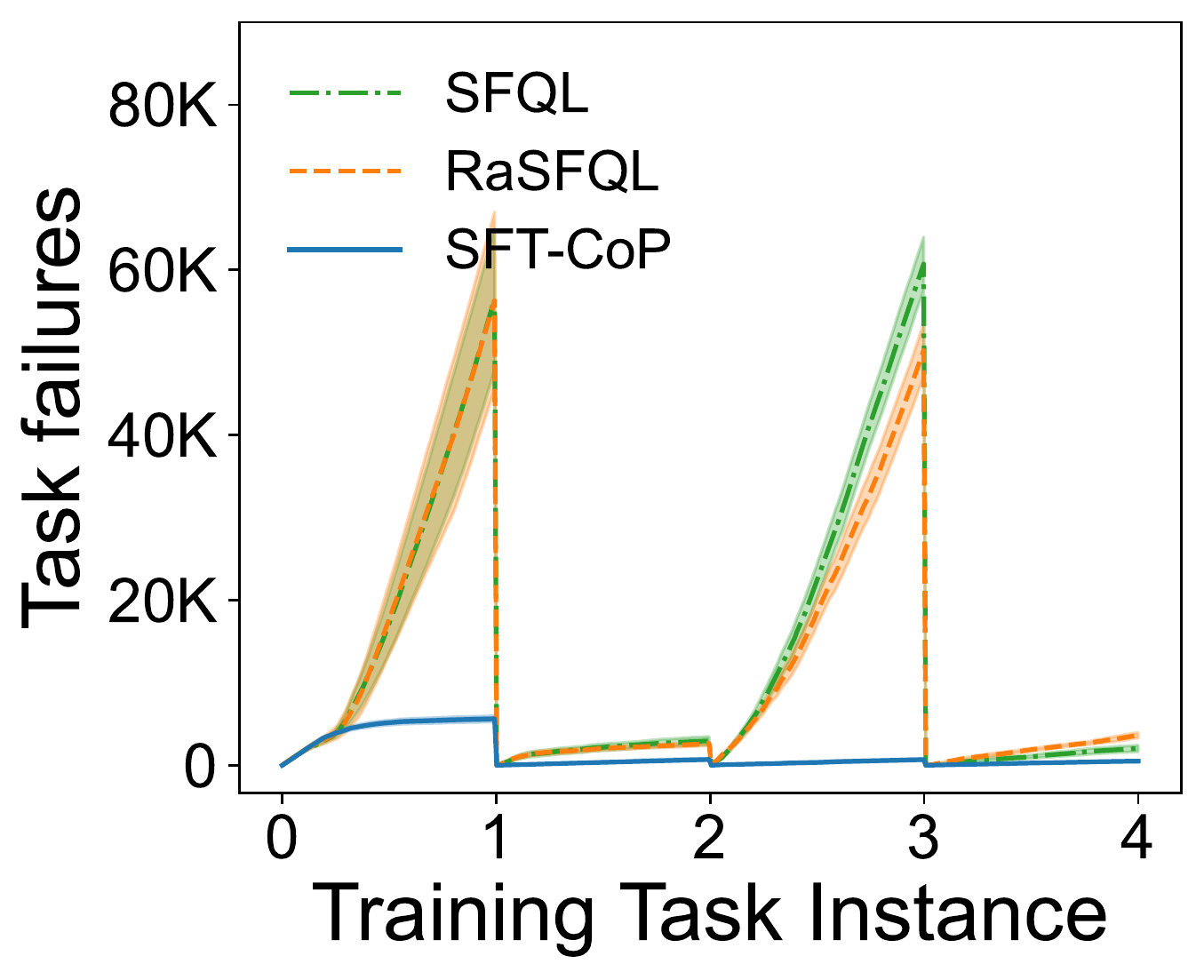}
		\caption{Task failures.}
	\end{subfigure}
	\begin{subfigure}[t]{.24\textwidth}
	    \includegraphics[width=1.\textwidth]{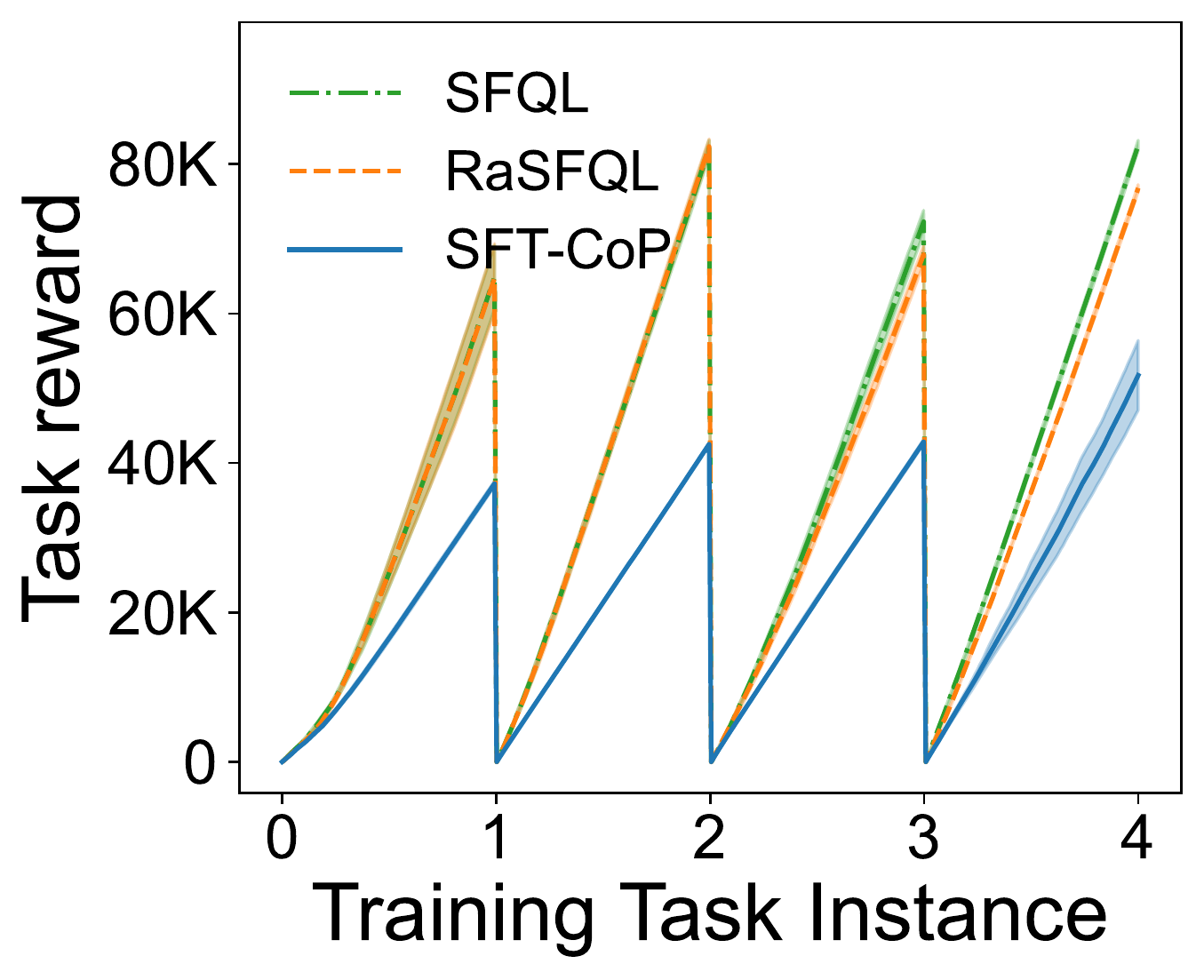}
		\caption{Task rewards.}
	\end{subfigure}
	\begin{subfigure}[t]{.24\textwidth}
		\includegraphics[width=1.\textwidth]{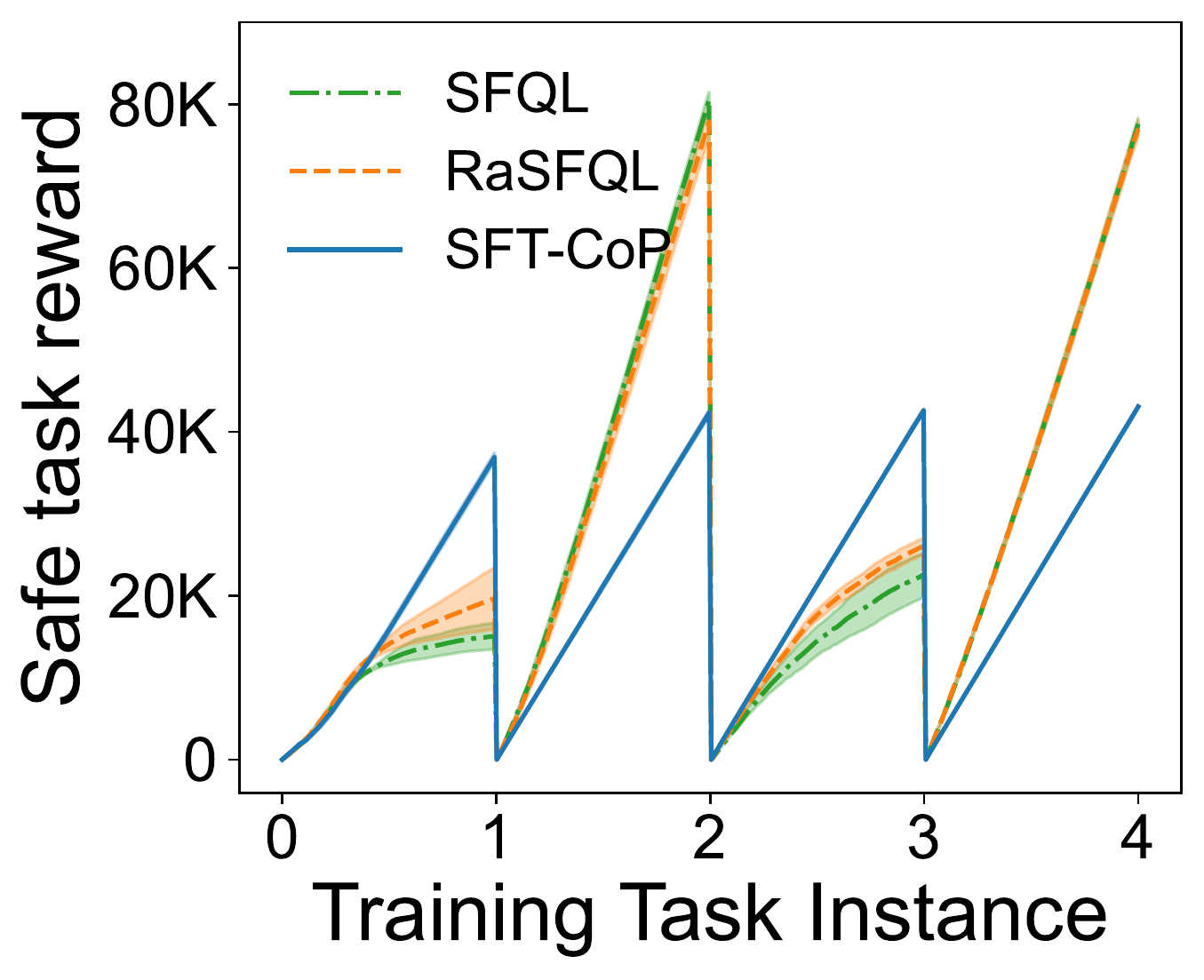}
		\caption{Task rewards from safe regions.}
	\end{subfigure}
	\begin{subfigure}[t]{.24\textwidth}
		\includegraphics[width=1.\textwidth]{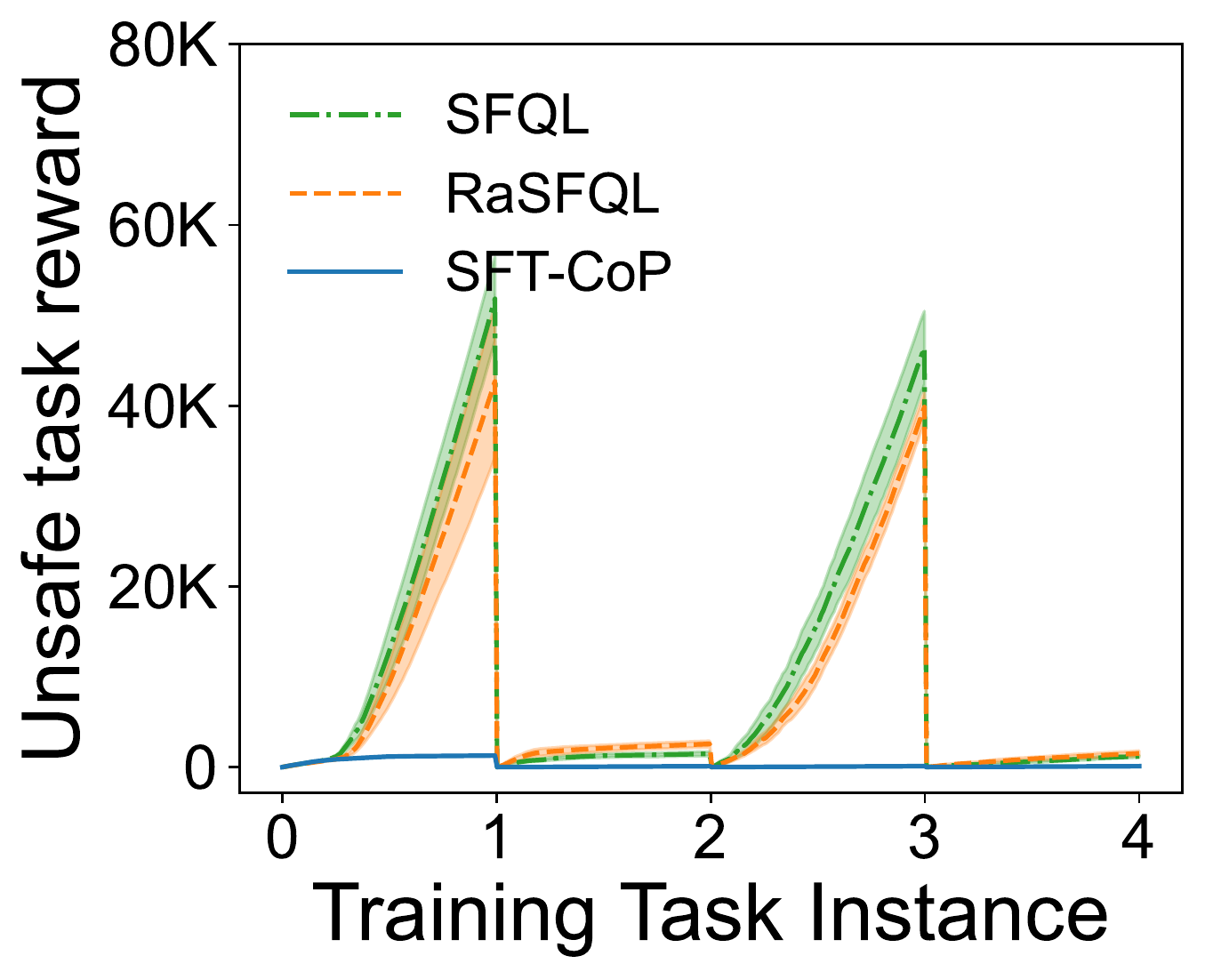}
		\caption{Task rewards from unsafe regions.}
	\end{subfigure}
	\caption{Performance of \sfql, \rasfql{} ($\beta=2$), and \method{} on the Reacher domain. We report accumulated and per-task (a, e) failures, (b, f) total rewards, (c, g) rewards from safe regions and (d, h) rewards from unsafe regions, over the training tasks.}
	\label{fig:reacher_hplc}
\end{figure*}
\begin{figure*}
    \centering
	\begin{subfigure}[t]{.24\textwidth}
		\includegraphics[width=1.\textwidth]{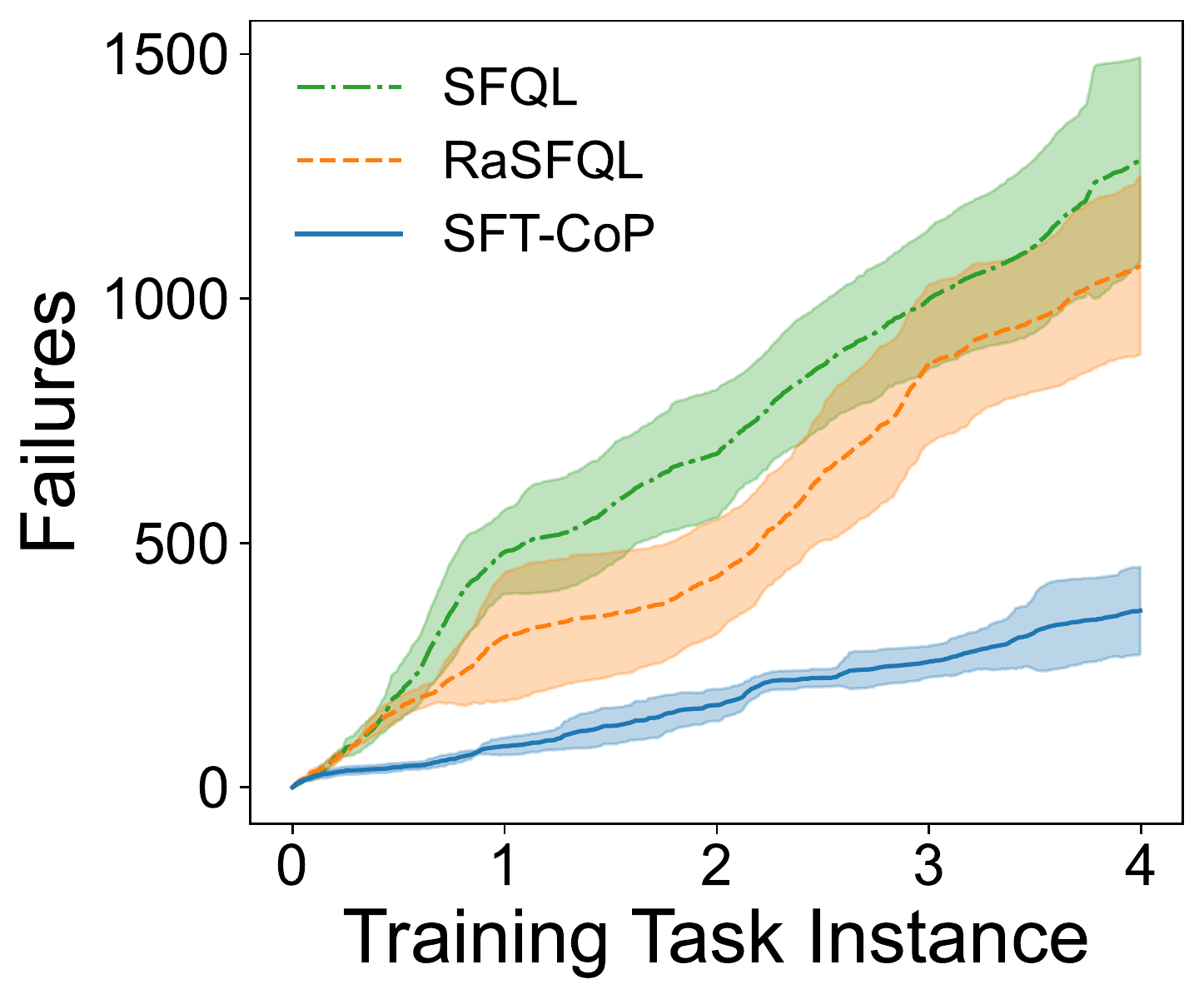}
		\caption{Accumulated failures.}
	\end{subfigure}
	\begin{subfigure}[t]{.24\textwidth}
	    \includegraphics[width=1.\textwidth]{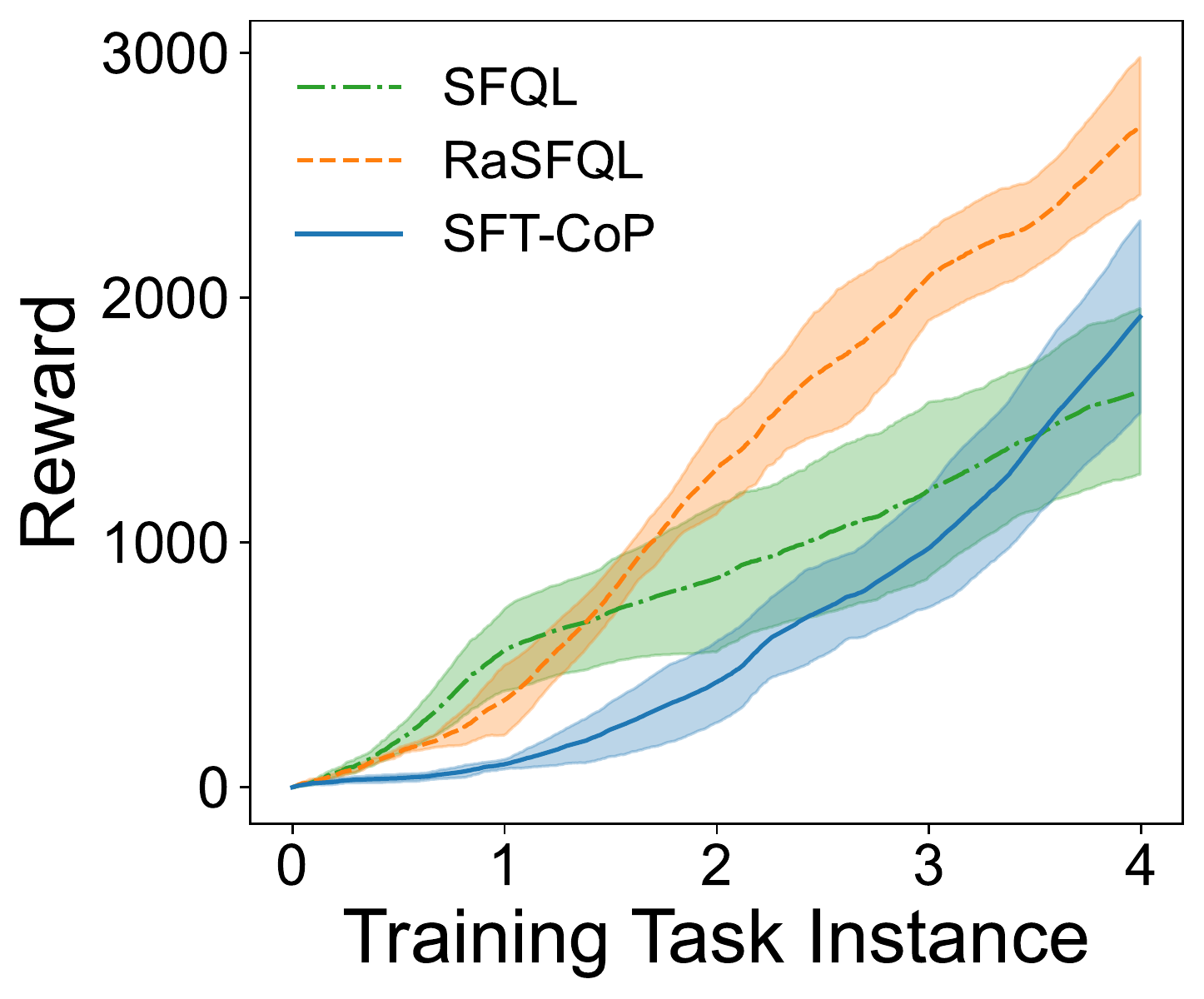}
		\caption{Accumulated rewards.}
	\end{subfigure}
	\begin{subfigure}[t]{.24\textwidth}
	    \includegraphics[width=1.\textwidth]{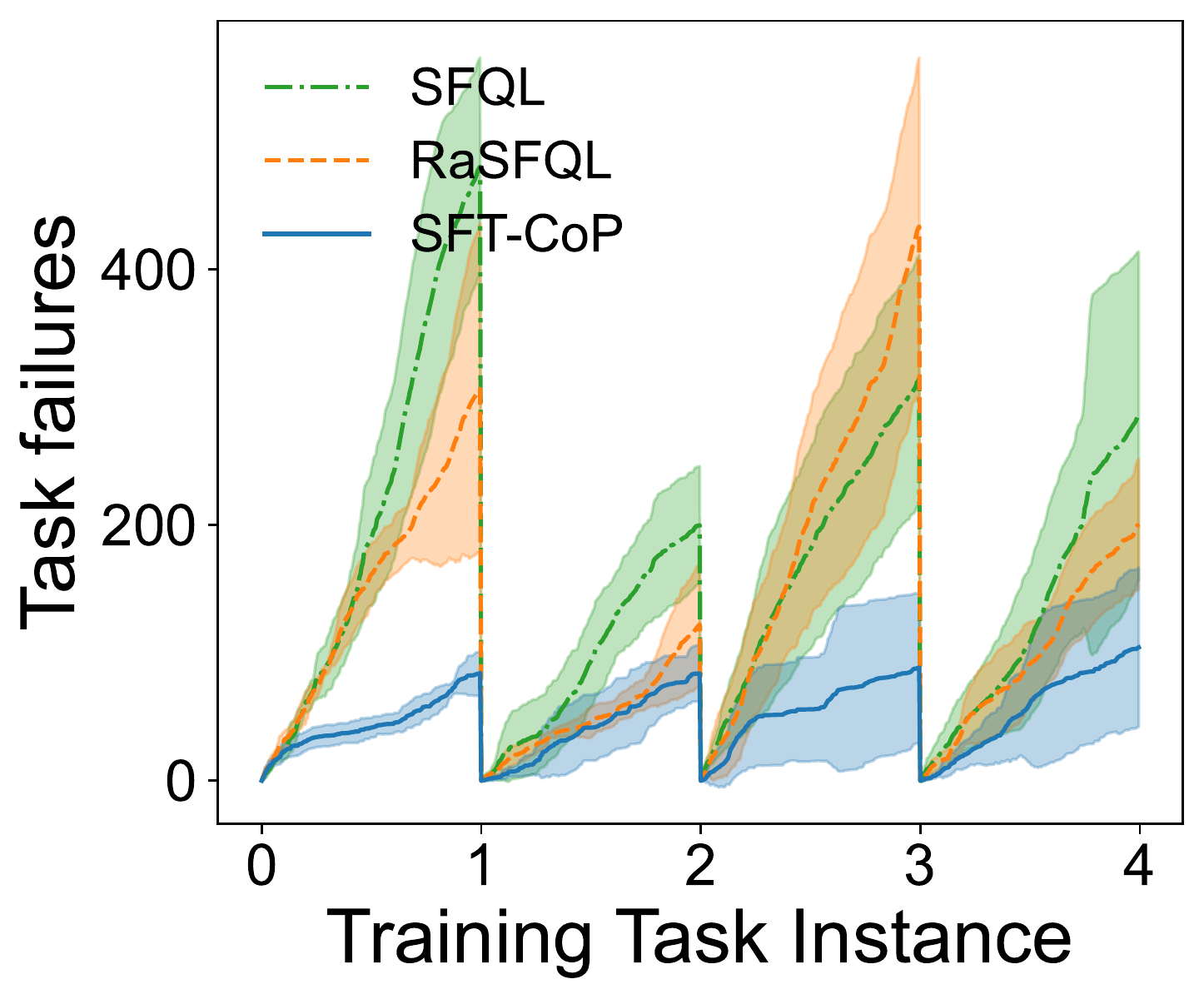}
		\caption{Task failures.}
	\end{subfigure}
	\begin{subfigure}[t]{.24\textwidth}
	    \includegraphics[width=1.\textwidth]{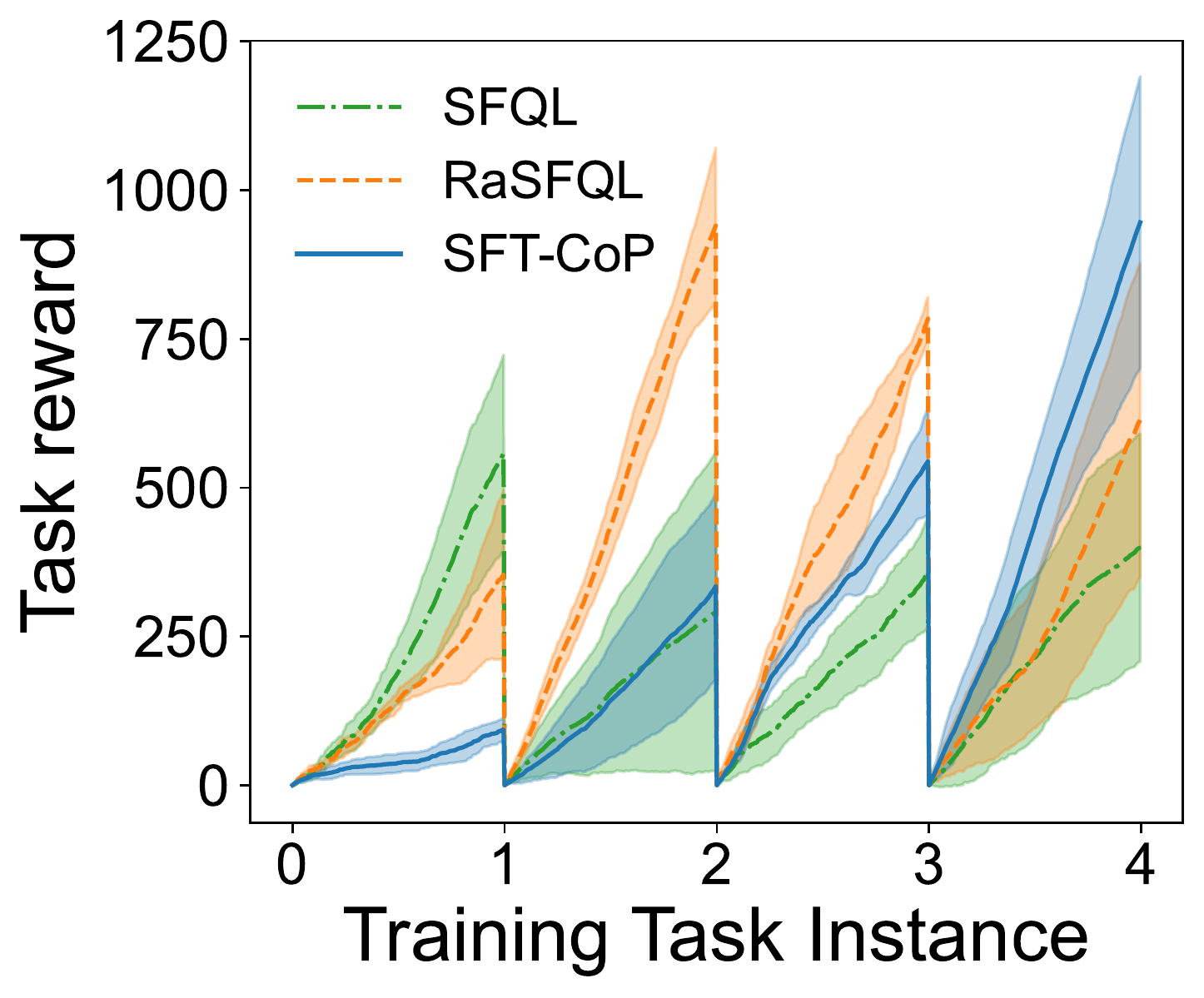}
		\caption{Task rewards.}
	\end{subfigure}
	\caption{Performance of \sfql, \rasfql{} ($\beta=2$) and \method{} on the SafetyGym domain. We report accumulated (a) failures and (b) rewards, and per-task (c) failures and (d) rewards, over the training tasks.}
	\label{fig:safetygym_hplc}
\end{figure*}

\begin{figure*}
    \centering
    \begin{minipage}[t]{0.95\textwidth}
    \centering Failures on 8 test tasks
    \vspace{0.2cm}
    \end{minipage}
    
	\begin{subfigure}[t]{.24\textwidth}
		\includegraphics[width=1.\textwidth]{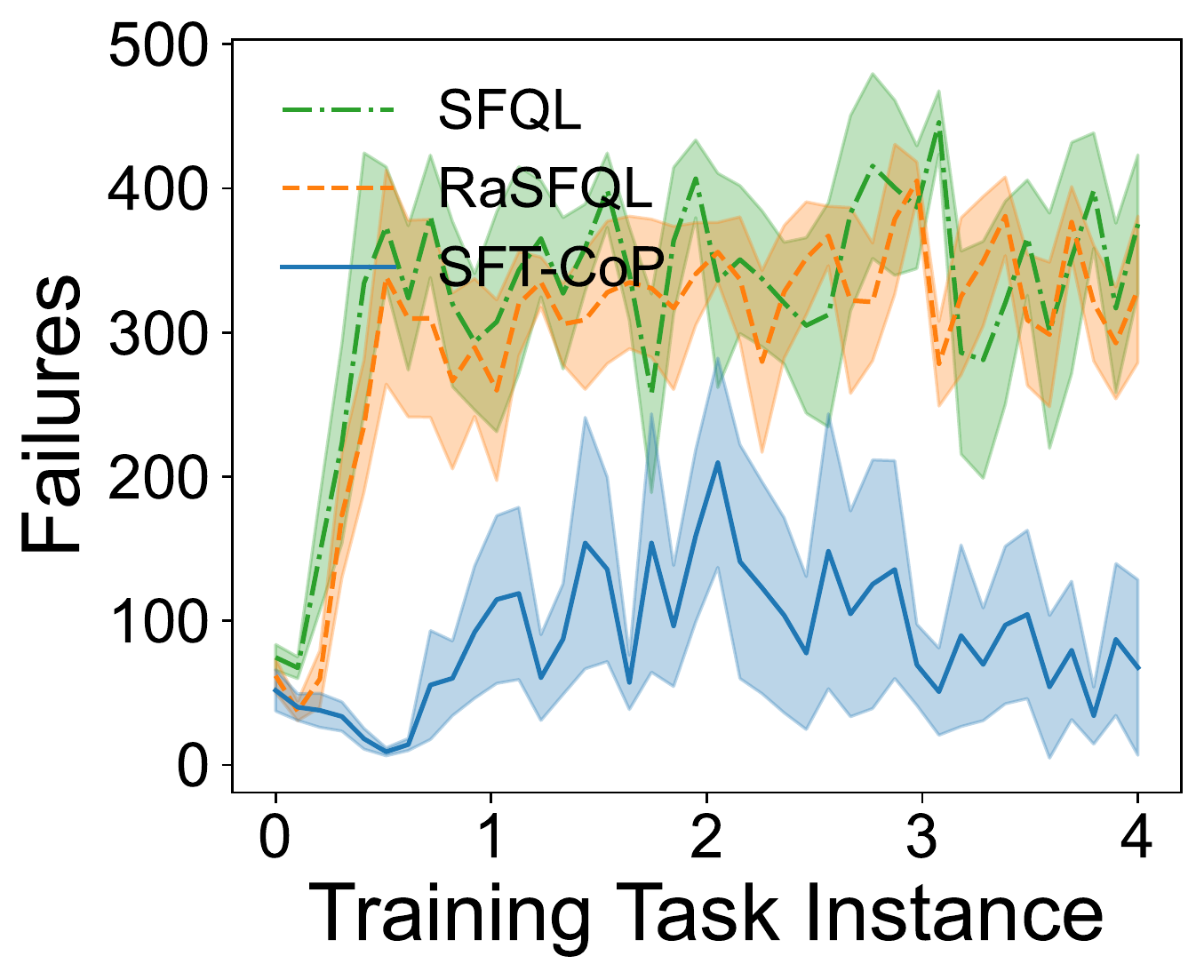}
	\end{subfigure}
	\begin{subfigure}[t]{.24\textwidth}
		\includegraphics[width=1.\textwidth]{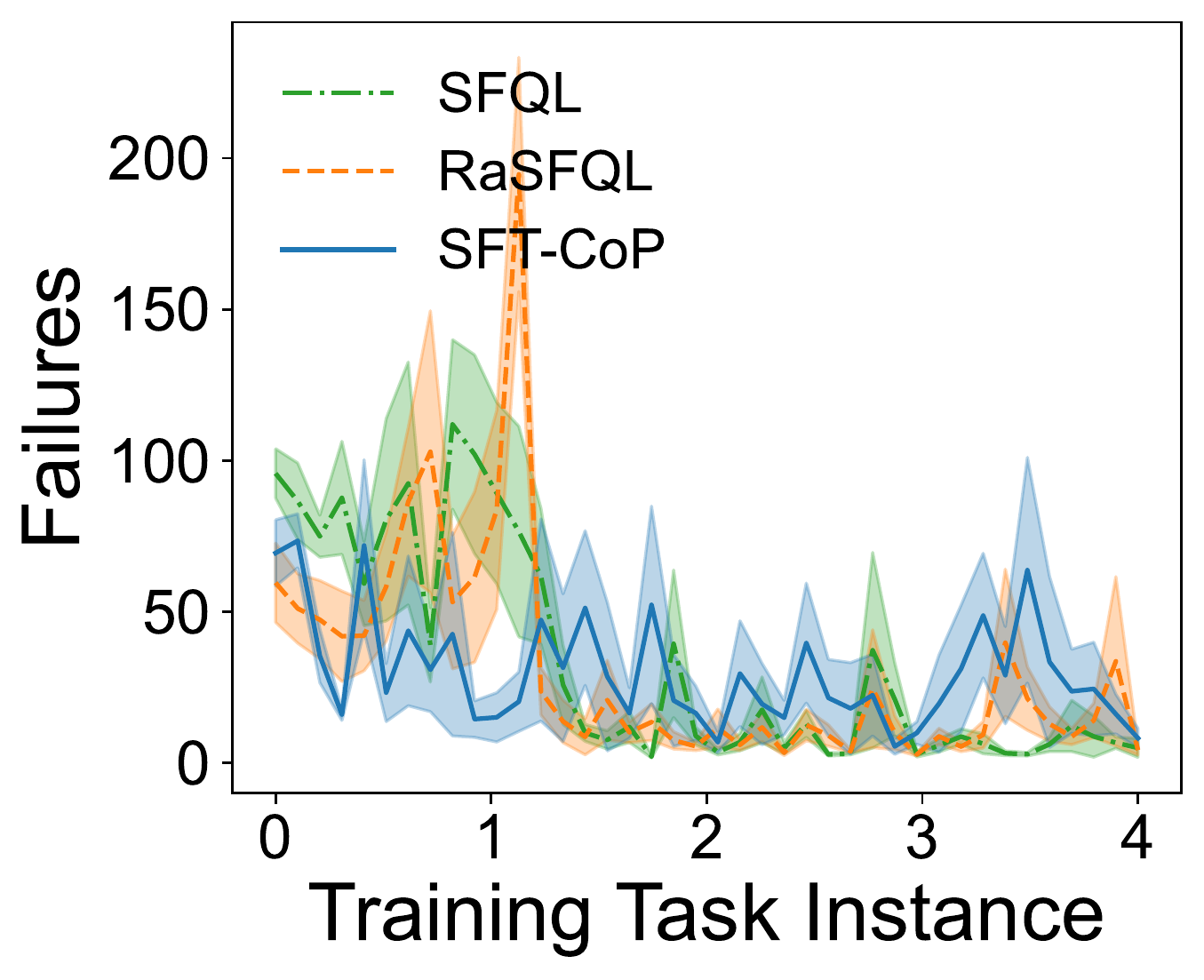}
	\end{subfigure}
	\begin{subfigure}[t]{.24\textwidth}
		\includegraphics[width=1.\textwidth]{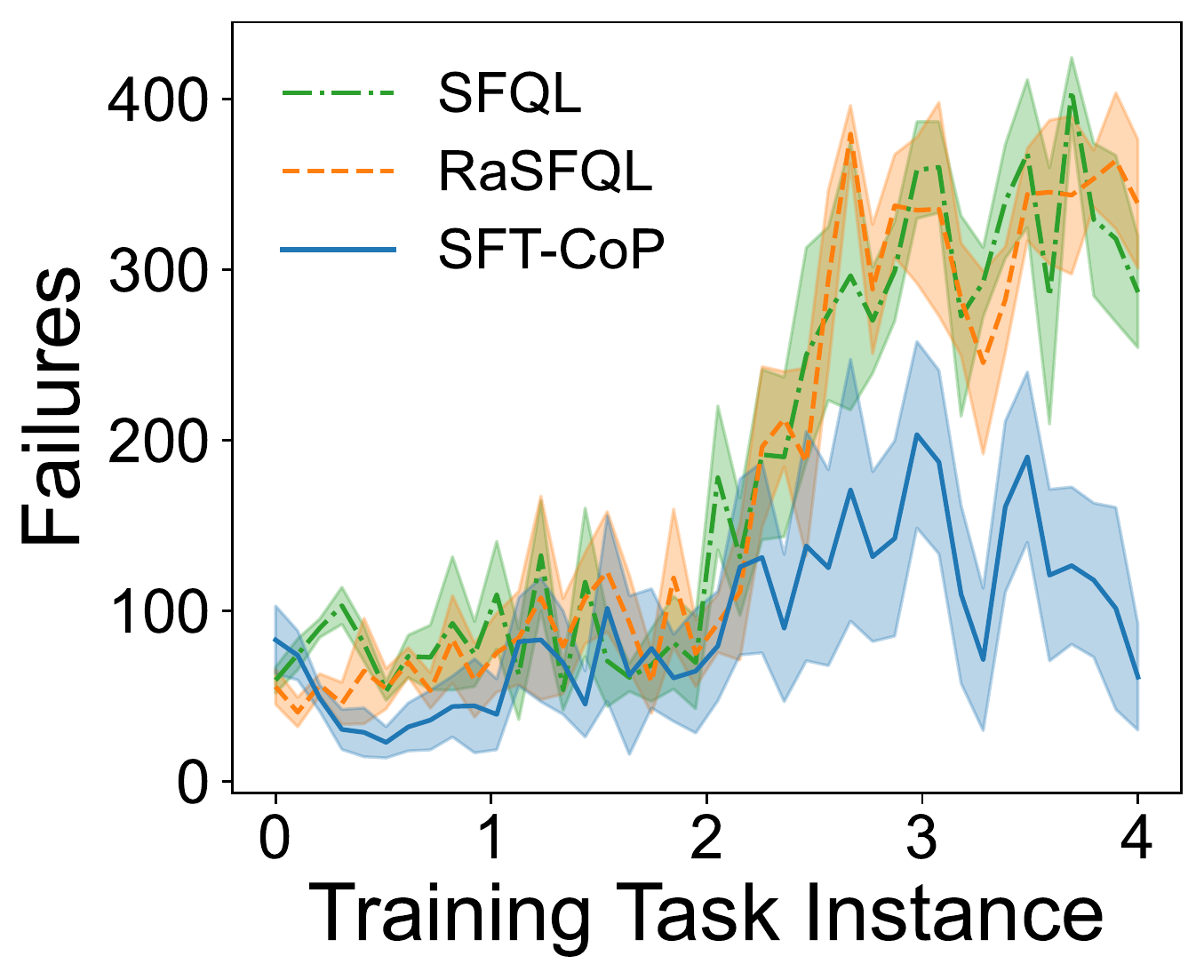}
	\end{subfigure}
	\begin{subfigure}[t]{.24\textwidth}
		\includegraphics[width=1.\textwidth]{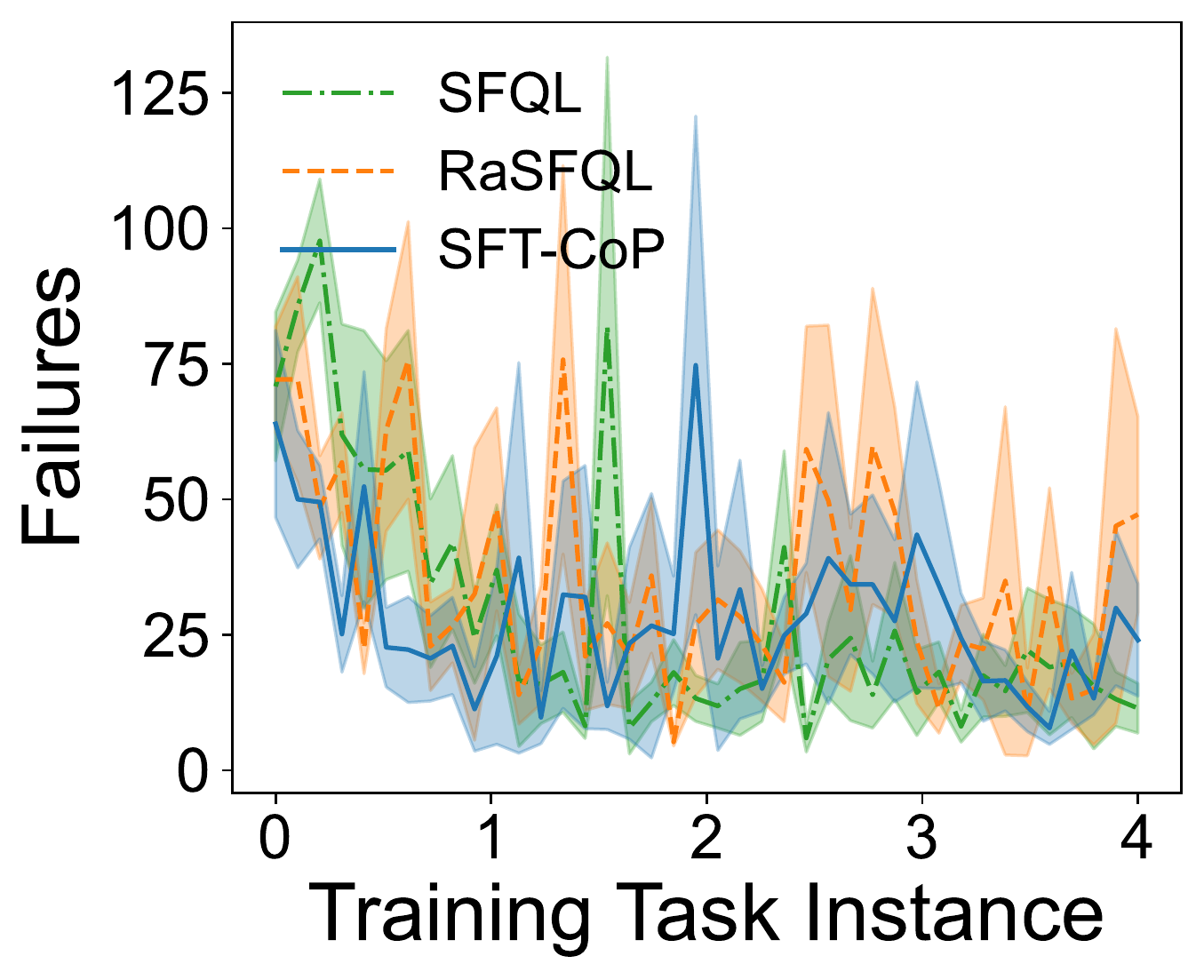}
	\end{subfigure}
	\begin{subfigure}[t]{.24\textwidth}
		\includegraphics[width=1.\textwidth]{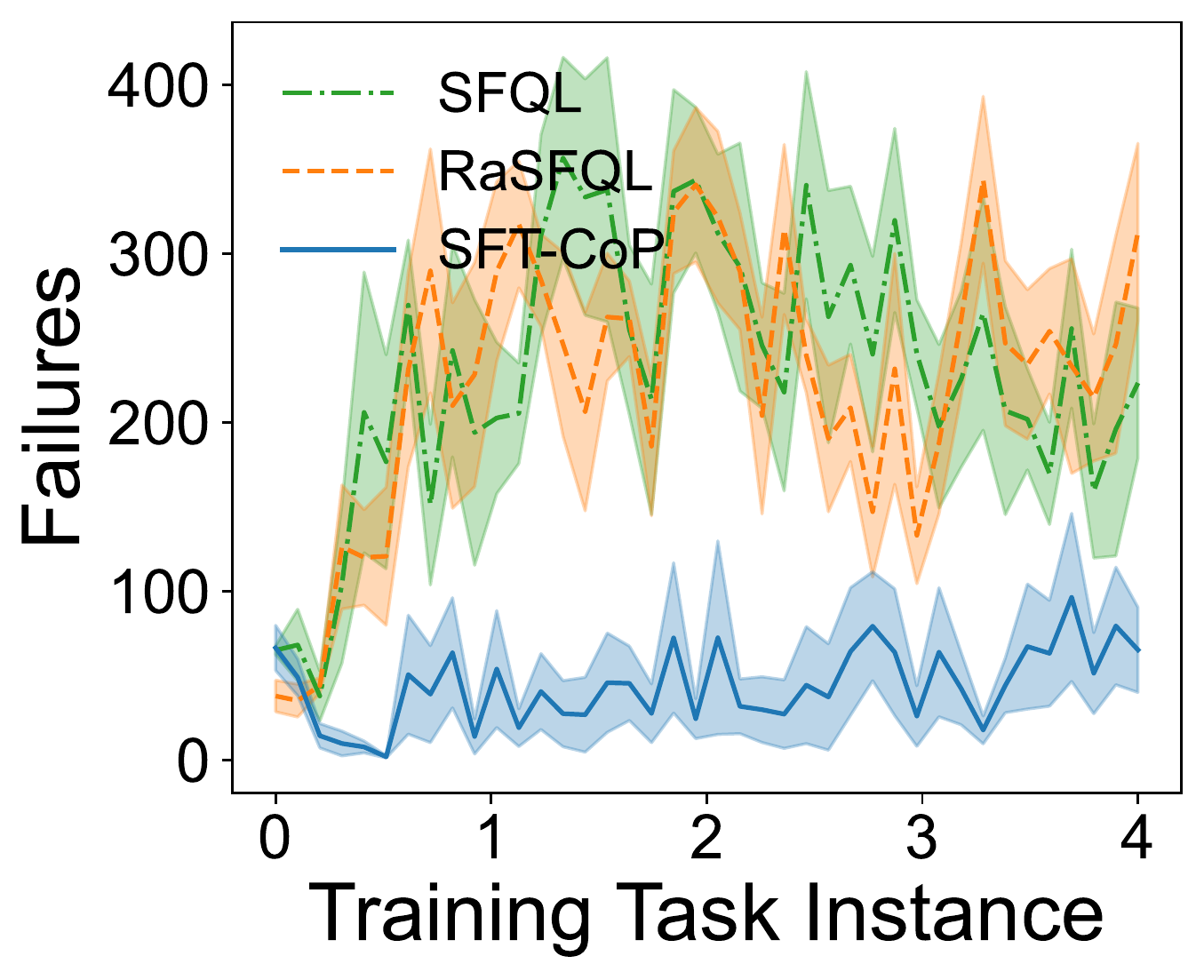}
	\end{subfigure}
	\begin{subfigure}[t]{.24\textwidth}
		\includegraphics[width=1.\textwidth]{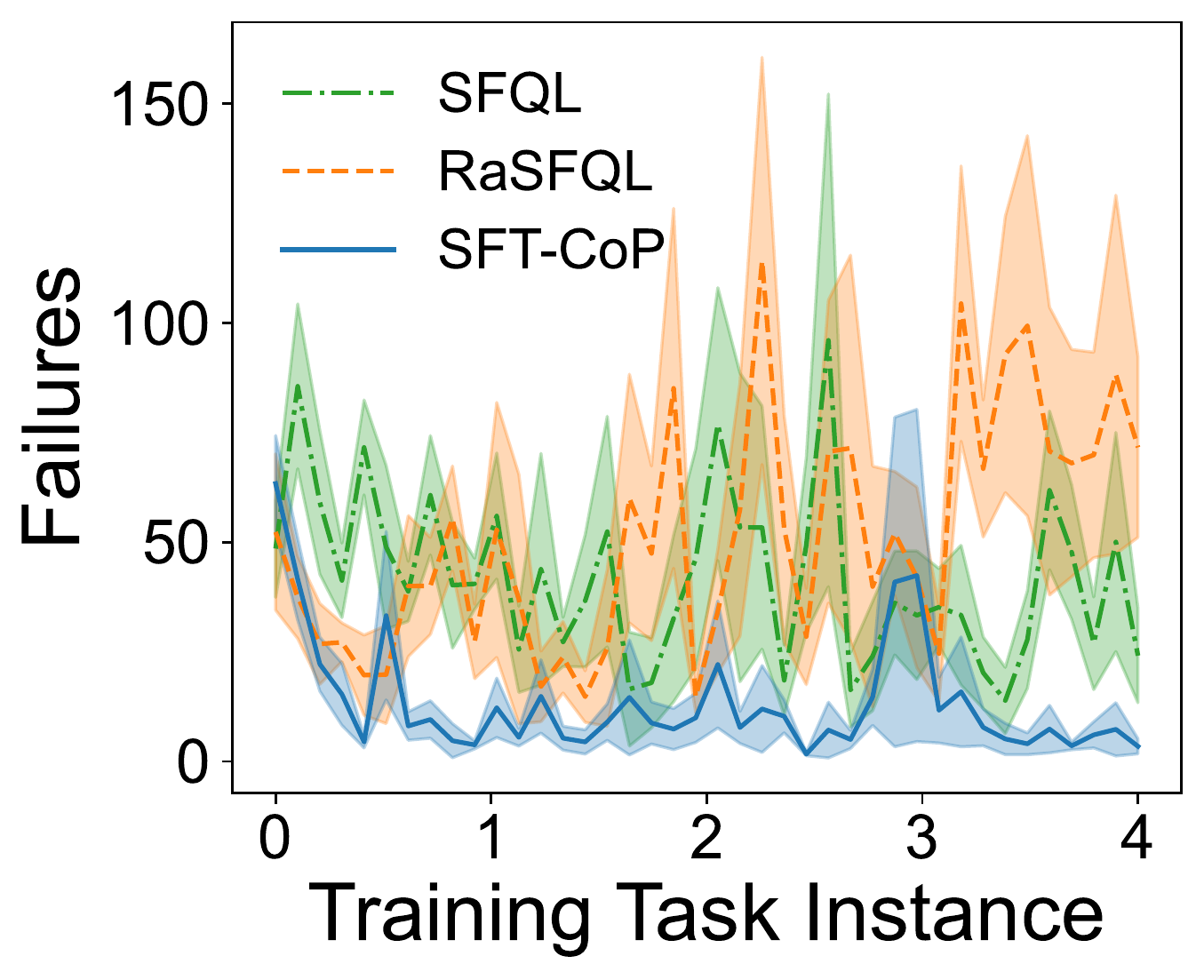}
	\end{subfigure}
	\begin{subfigure}[t]{.24\textwidth}
		\includegraphics[width=1.\textwidth]{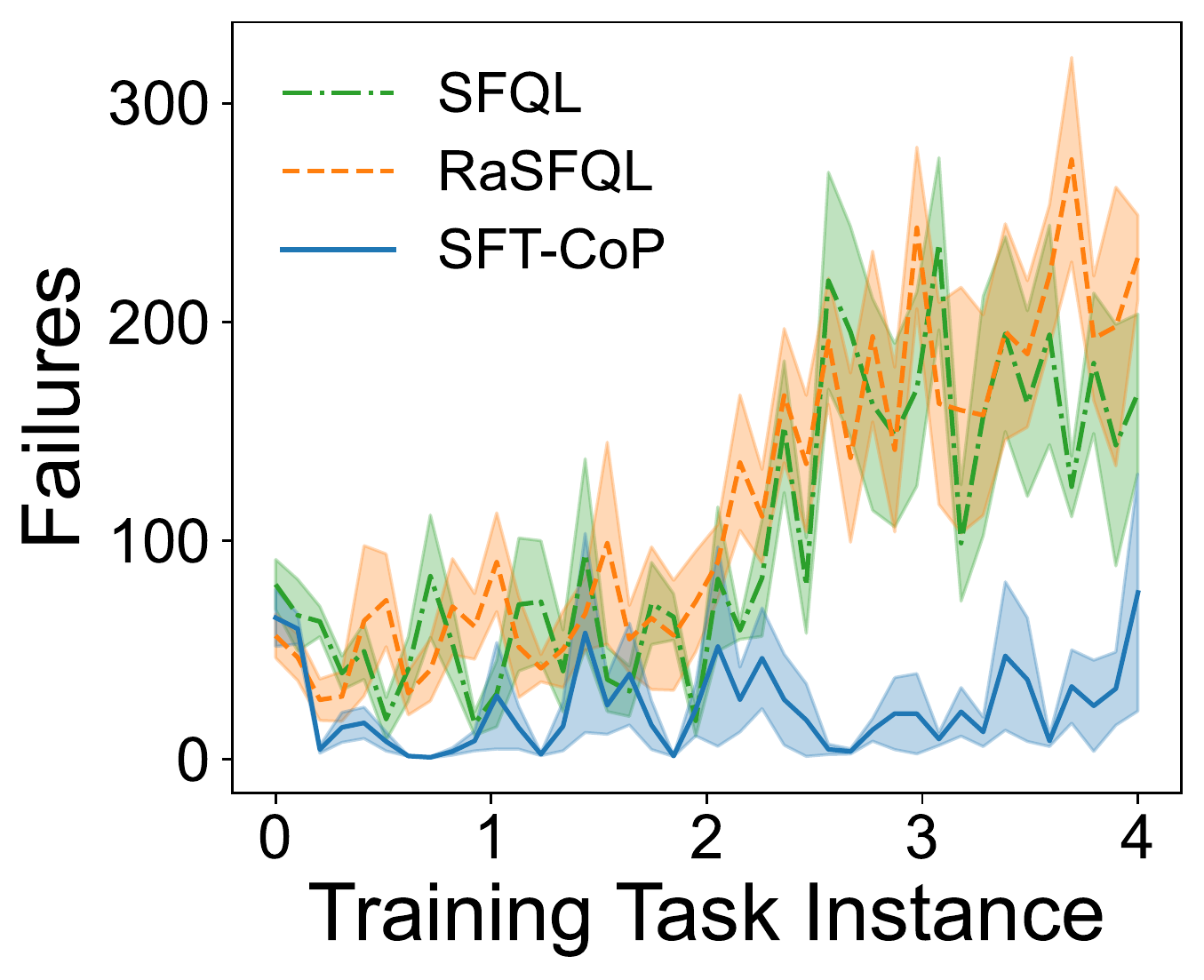}
	\end{subfigure}
	\begin{subfigure}[t]{.24\textwidth}
		\includegraphics[width=1.\textwidth]{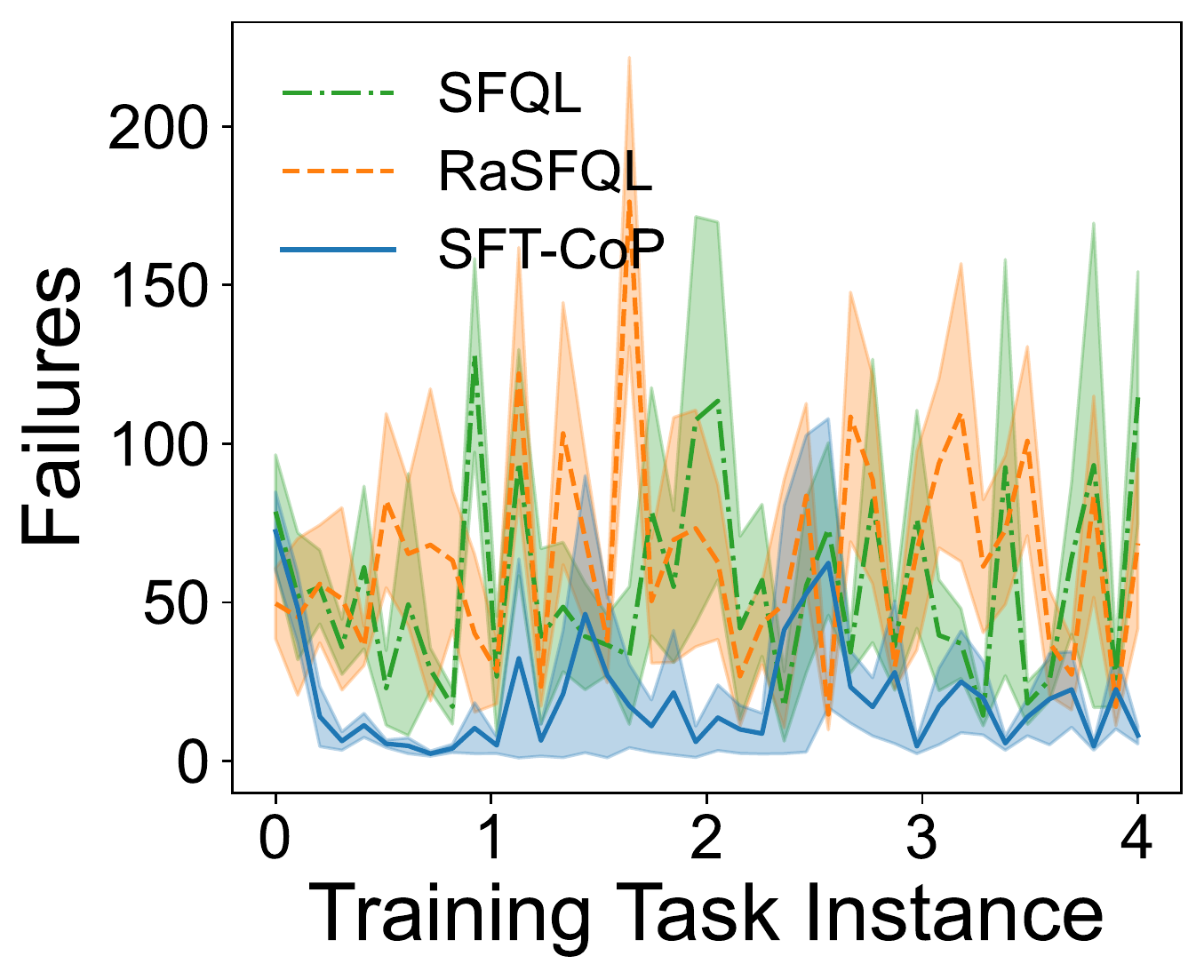}
	\end{subfigure}

    \begin{minipage}[t]{0.95\textwidth}
    \vspace{0.2cm}
    \centering Rewards from safe regions on 8 test tasks
    \vspace{0.2cm}
    \end{minipage}

	\begin{subfigure}[t]{.24\textwidth}
		\includegraphics[width=1.\textwidth]{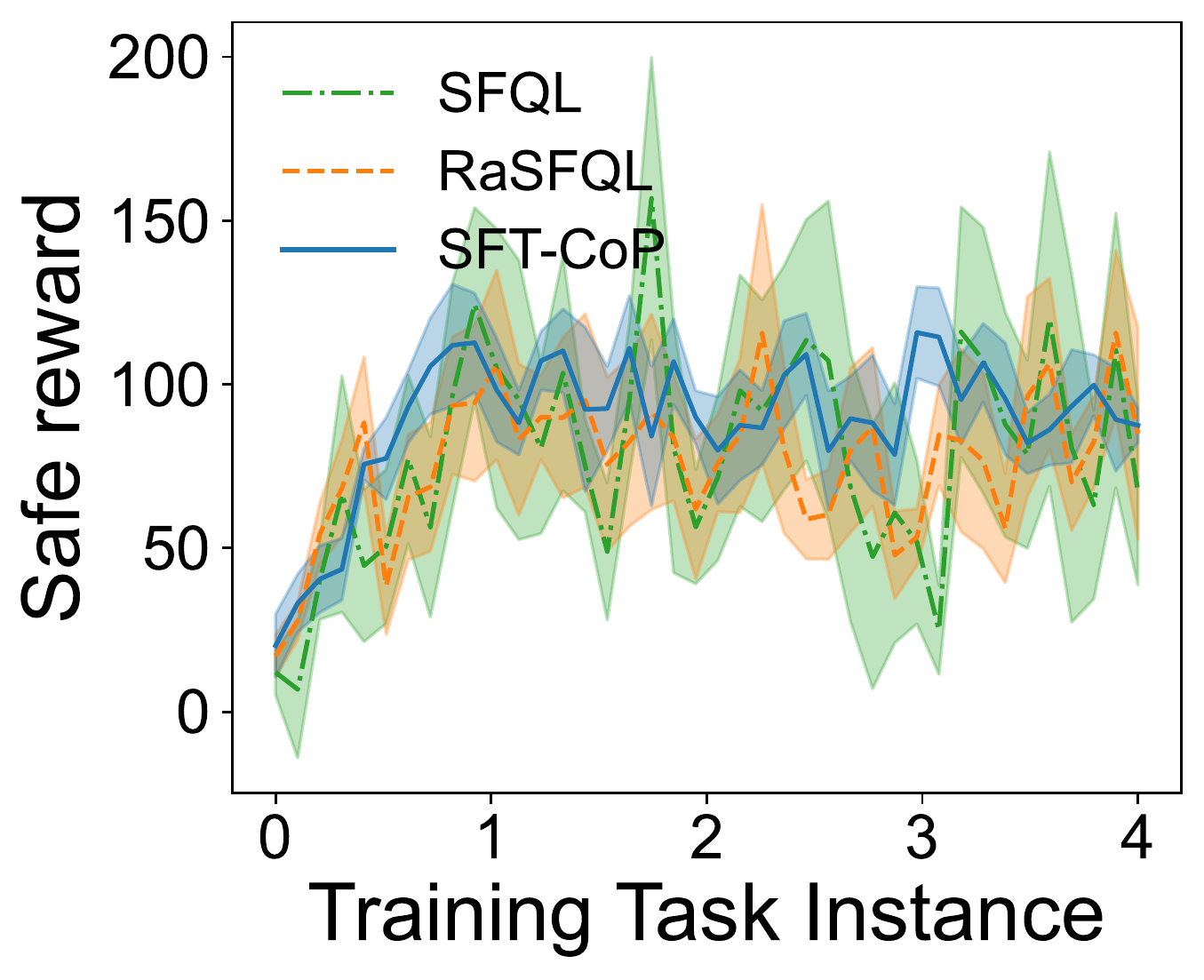}
	\end{subfigure}
	\begin{subfigure}[t]{.24\textwidth}
		\includegraphics[width=1.\textwidth]{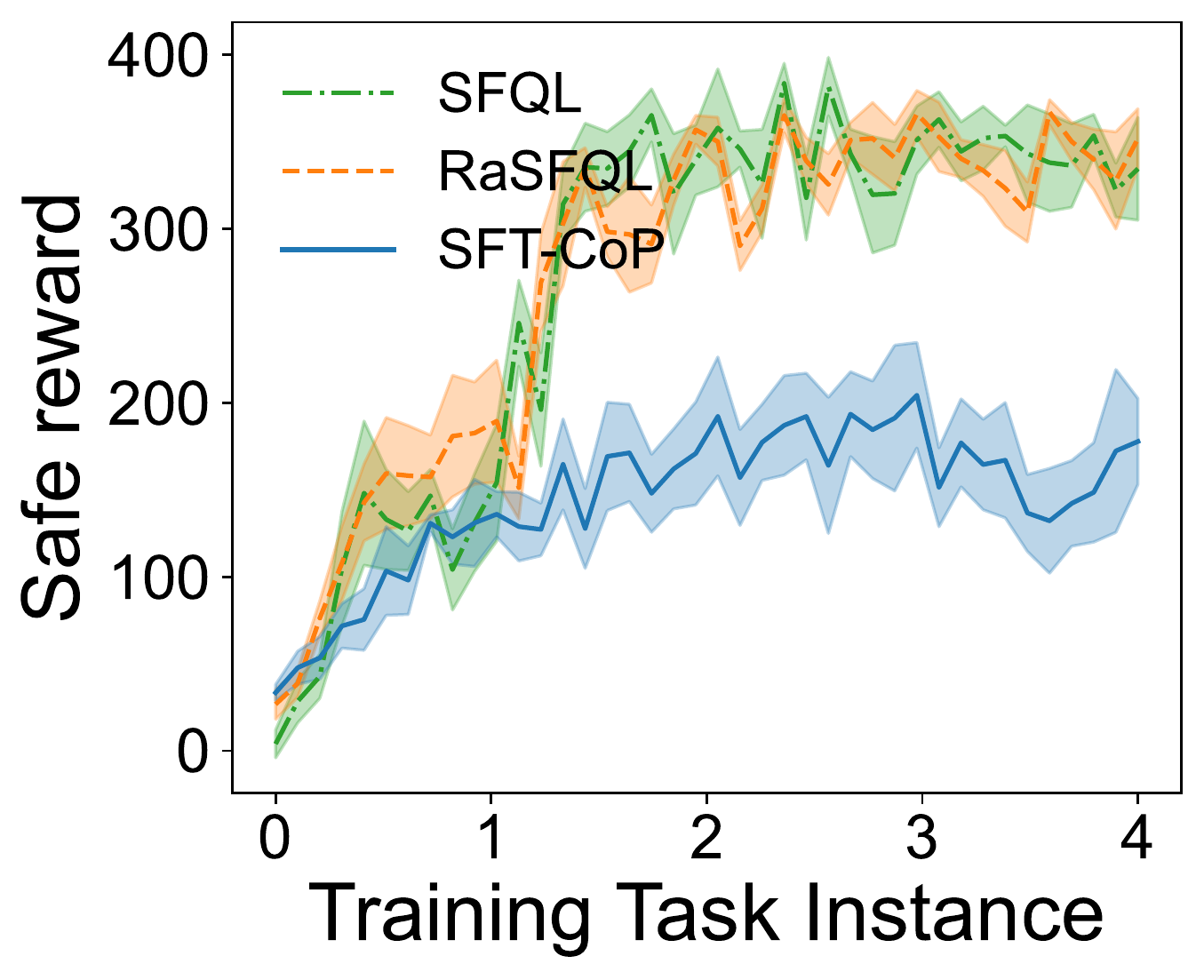}
	\end{subfigure}
	\begin{subfigure}[t]{.24\textwidth}
		\includegraphics[width=1.\textwidth]{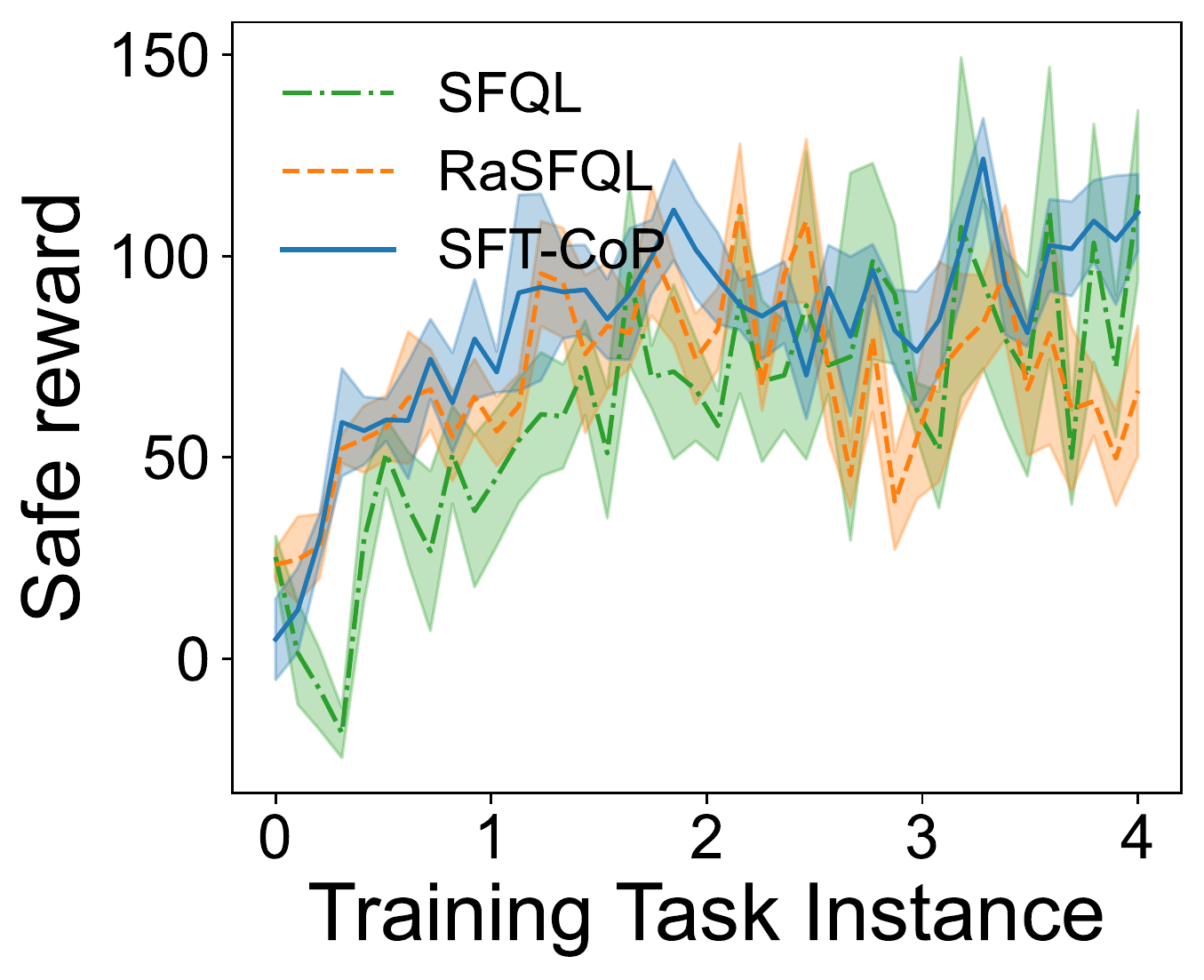}
	\end{subfigure}
	\begin{subfigure}[t]{.24\textwidth}
		\includegraphics[width=1.\textwidth]{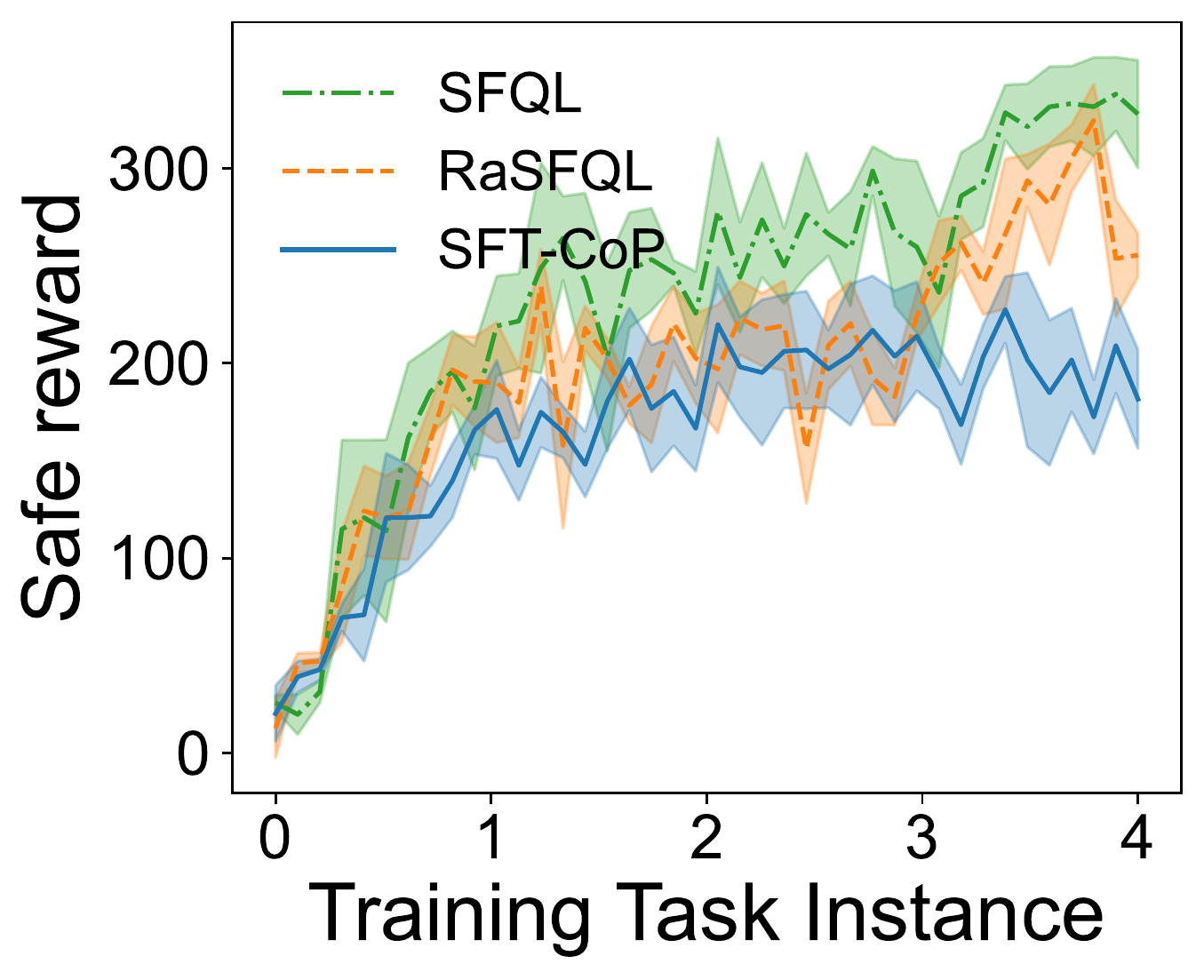}
	\end{subfigure}
	\begin{subfigure}[t]{.24\textwidth}
		\includegraphics[width=1.\textwidth]{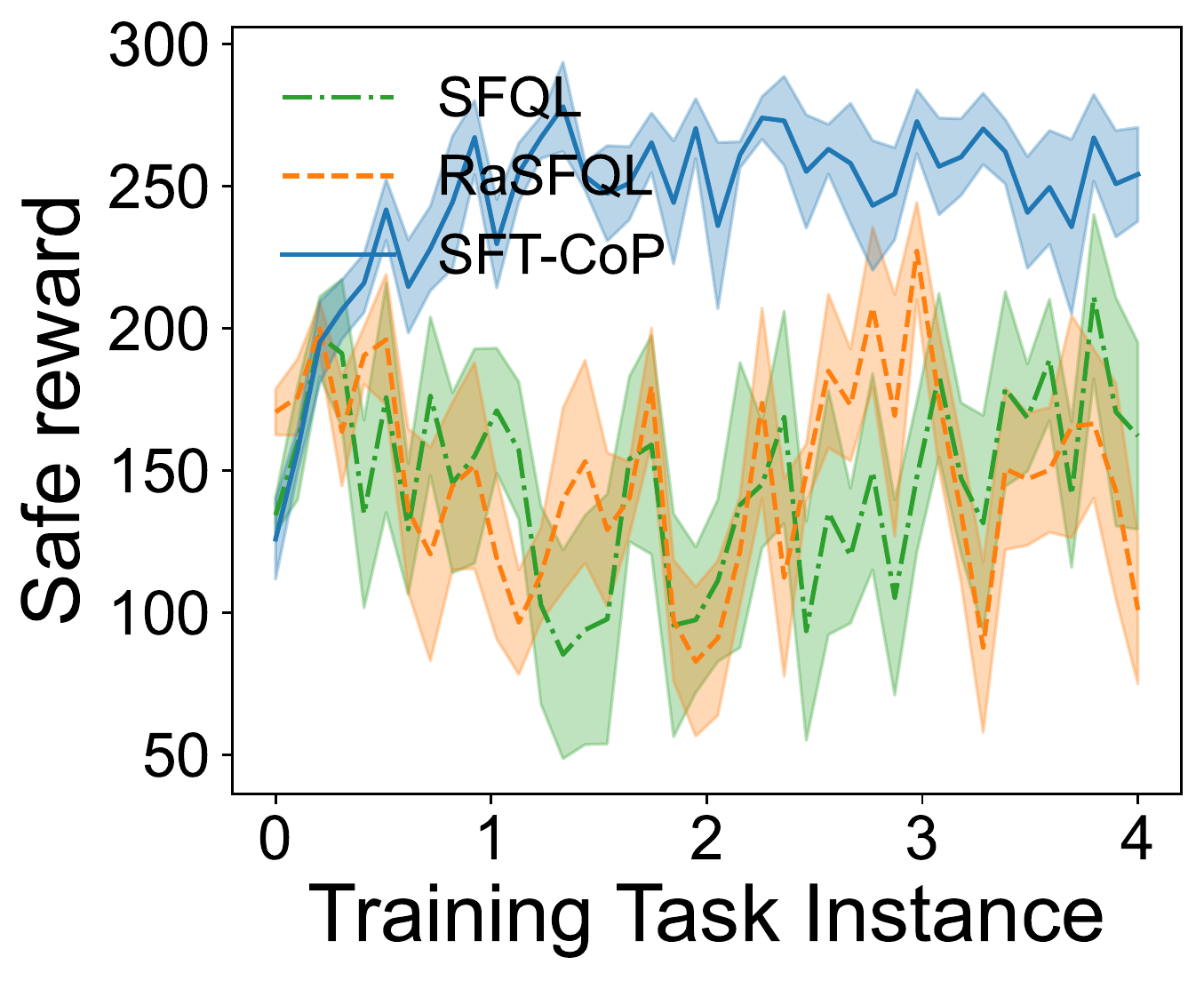}
	\end{subfigure}
	\begin{subfigure}[t]{.24\textwidth}
		\includegraphics[width=1.\textwidth]{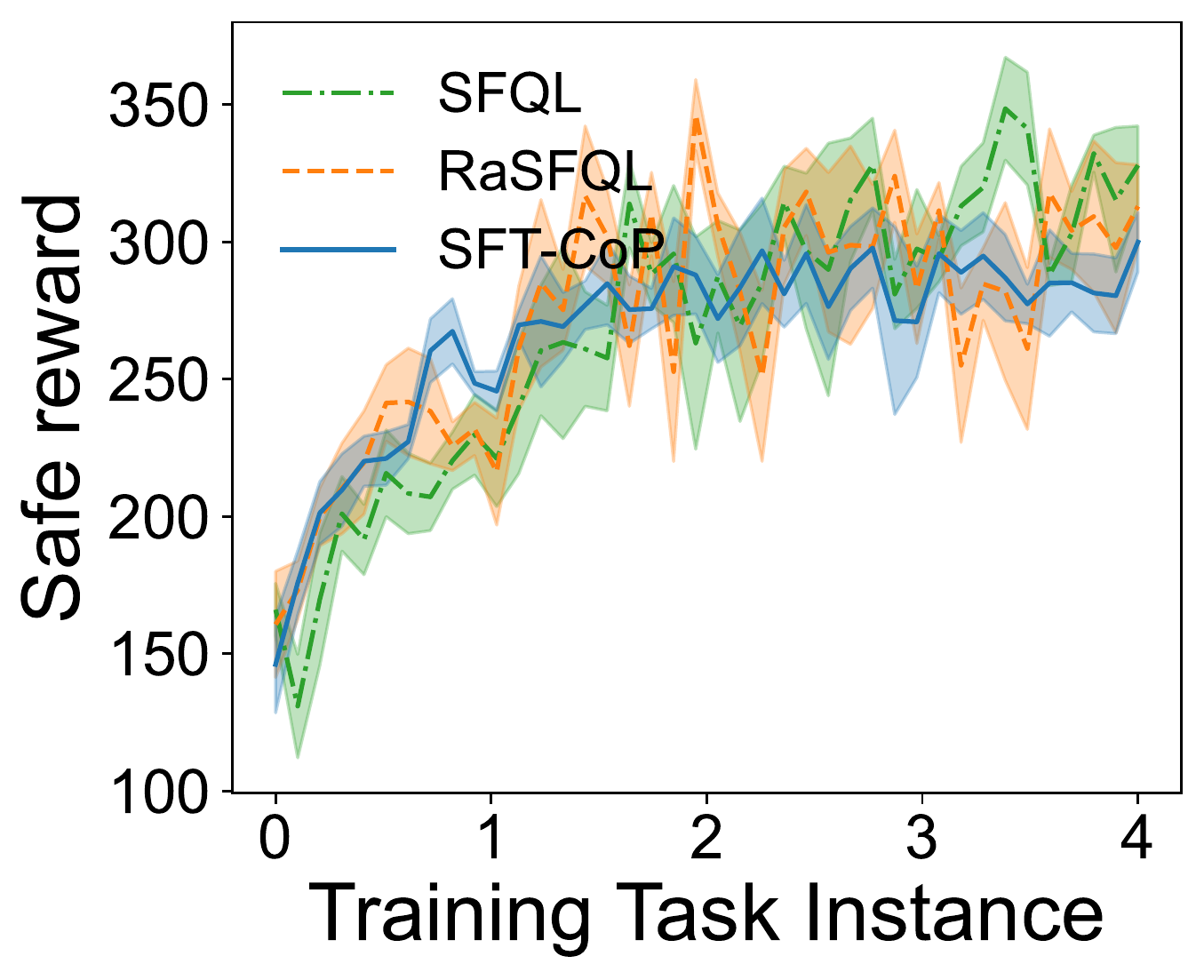}
	\end{subfigure}
	\begin{subfigure}[t]{.24\textwidth}
		\includegraphics[width=1.\textwidth]{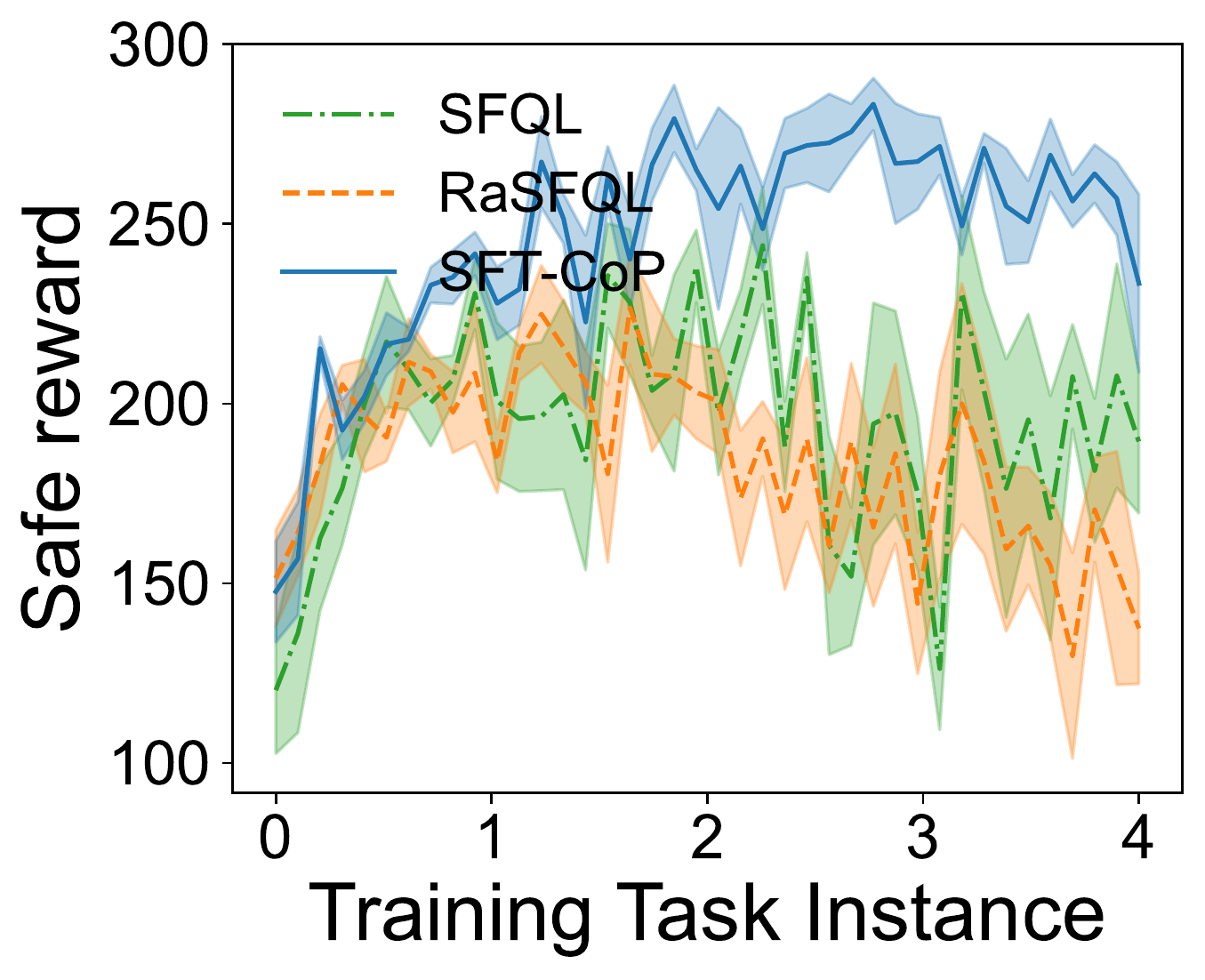}
	\end{subfigure}
	\begin{subfigure}[t]{.24\textwidth}
		\includegraphics[width=1.\textwidth]{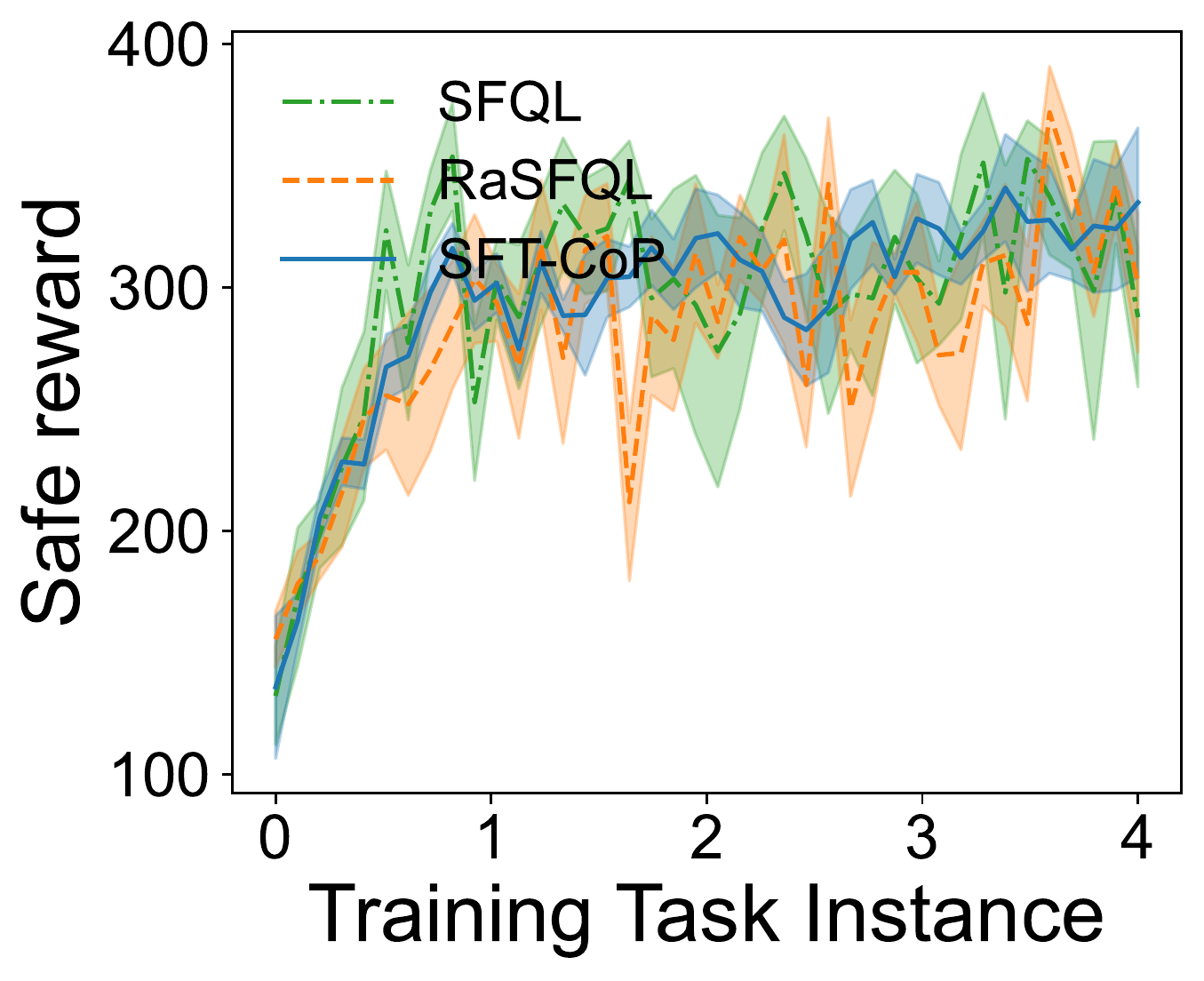}
	\end{subfigure}

    \begin{minipage}[t]{0.95\textwidth}
    \vspace{0.2cm}
    \centering Rewards from unsafe regions on 8 test tasks
    \vspace{0.2cm}
    \end{minipage}

	\begin{subfigure}[t]{.24\textwidth}
		\includegraphics[width=1.\textwidth]{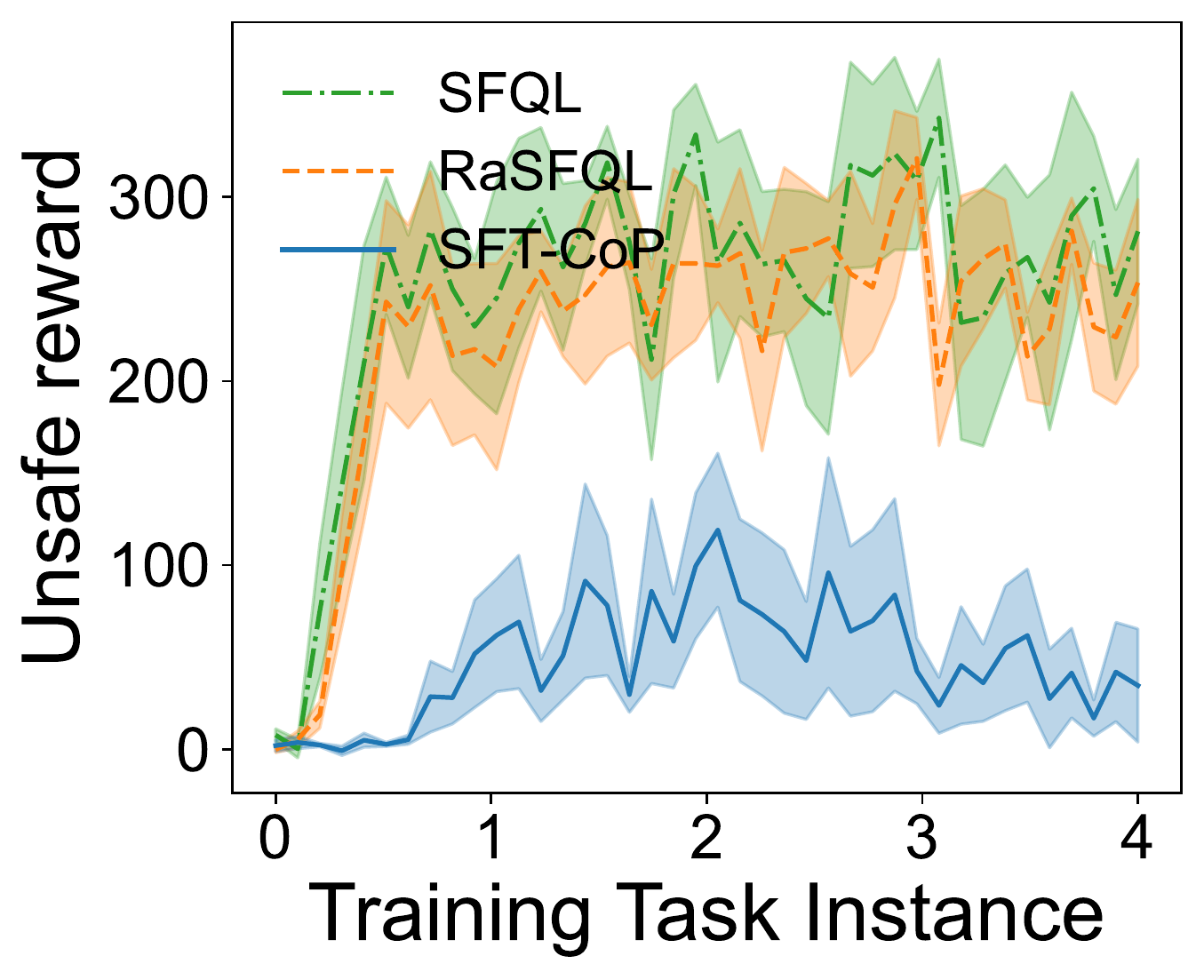}
	\end{subfigure}
	\begin{subfigure}[t]{.24\textwidth}
		\includegraphics[width=1.\textwidth]{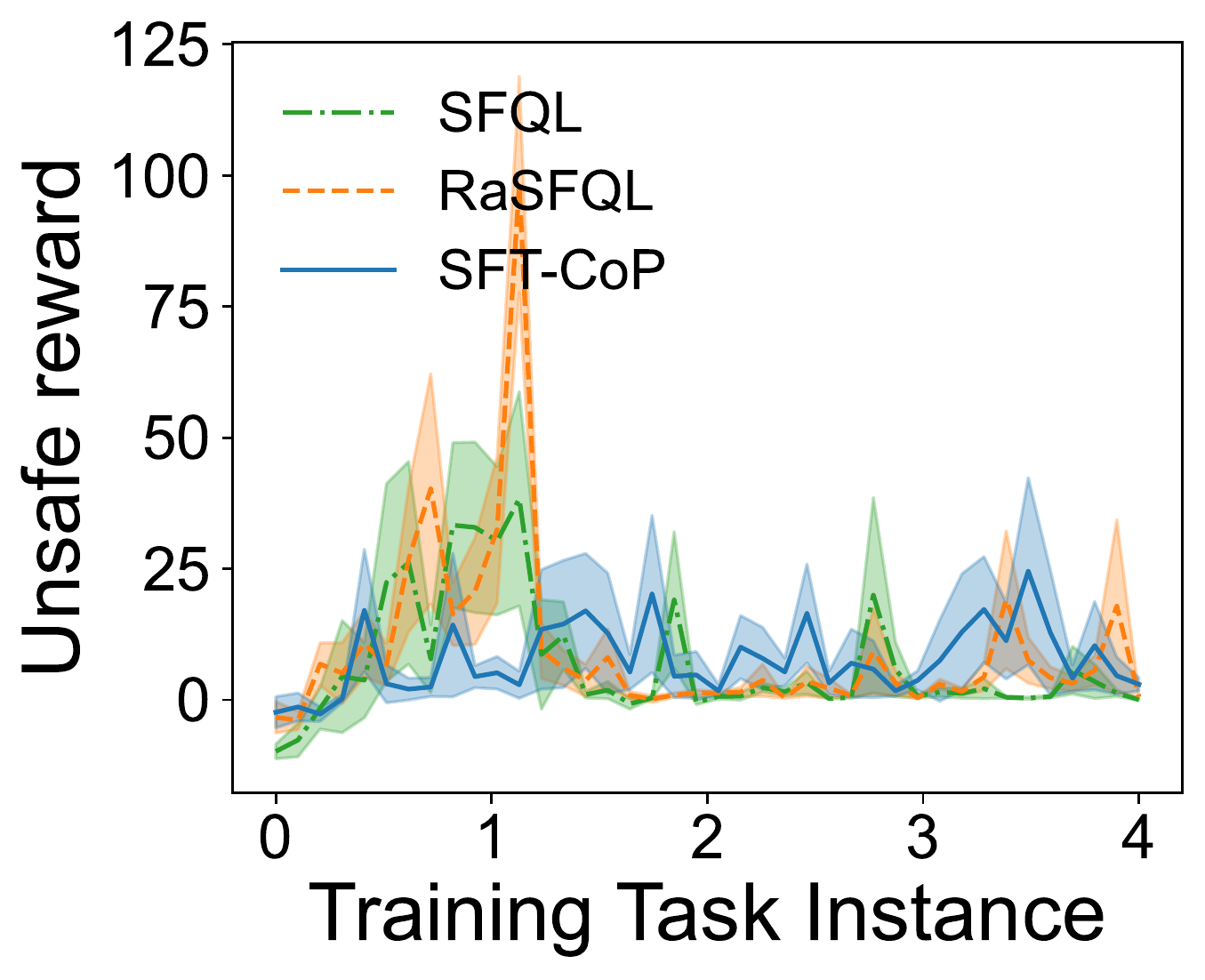}
	\end{subfigure}
	\begin{subfigure}[t]{.24\textwidth}
		\includegraphics[width=1.\textwidth]{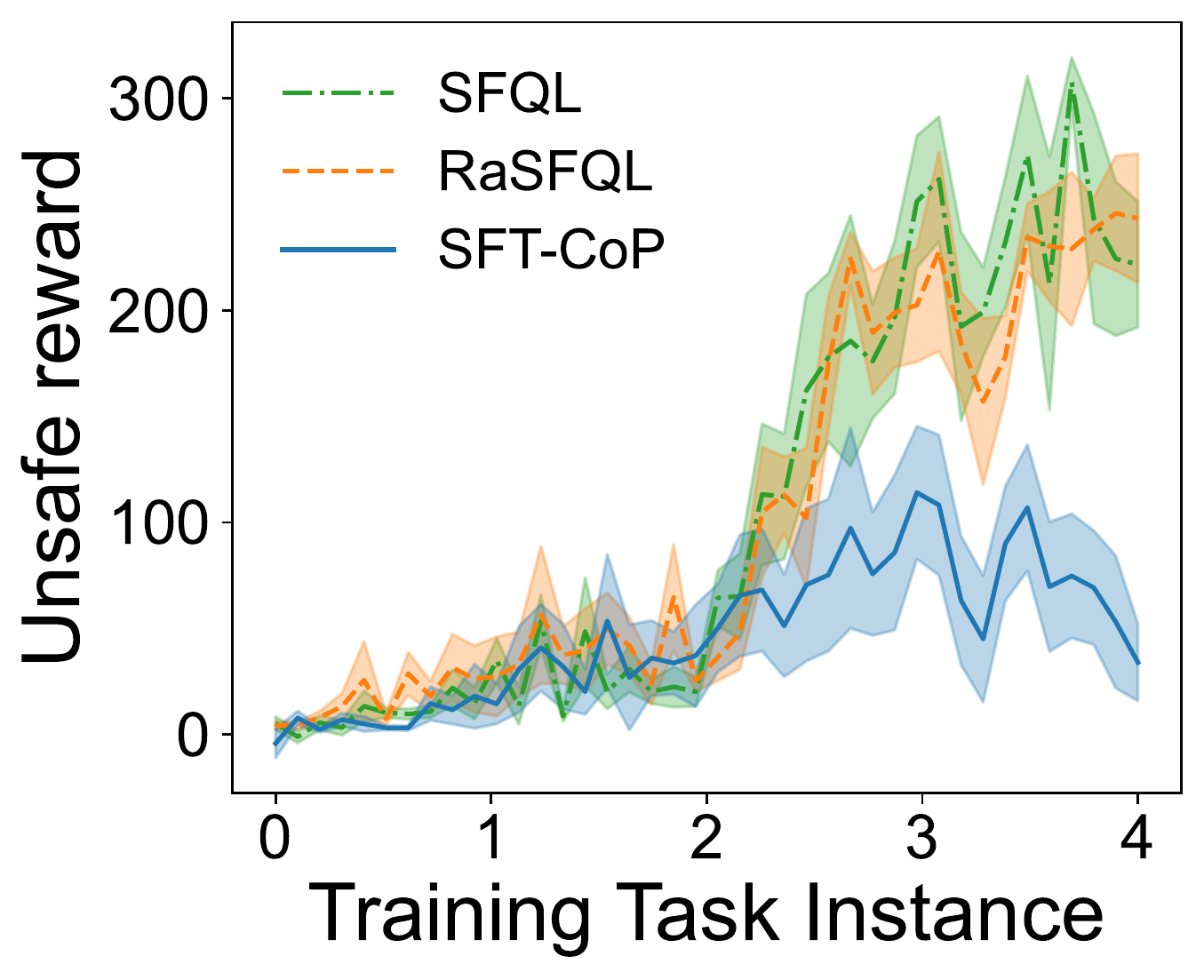}
	\end{subfigure}
	\begin{subfigure}[t]{.24\textwidth}
		\includegraphics[width=1.\textwidth]{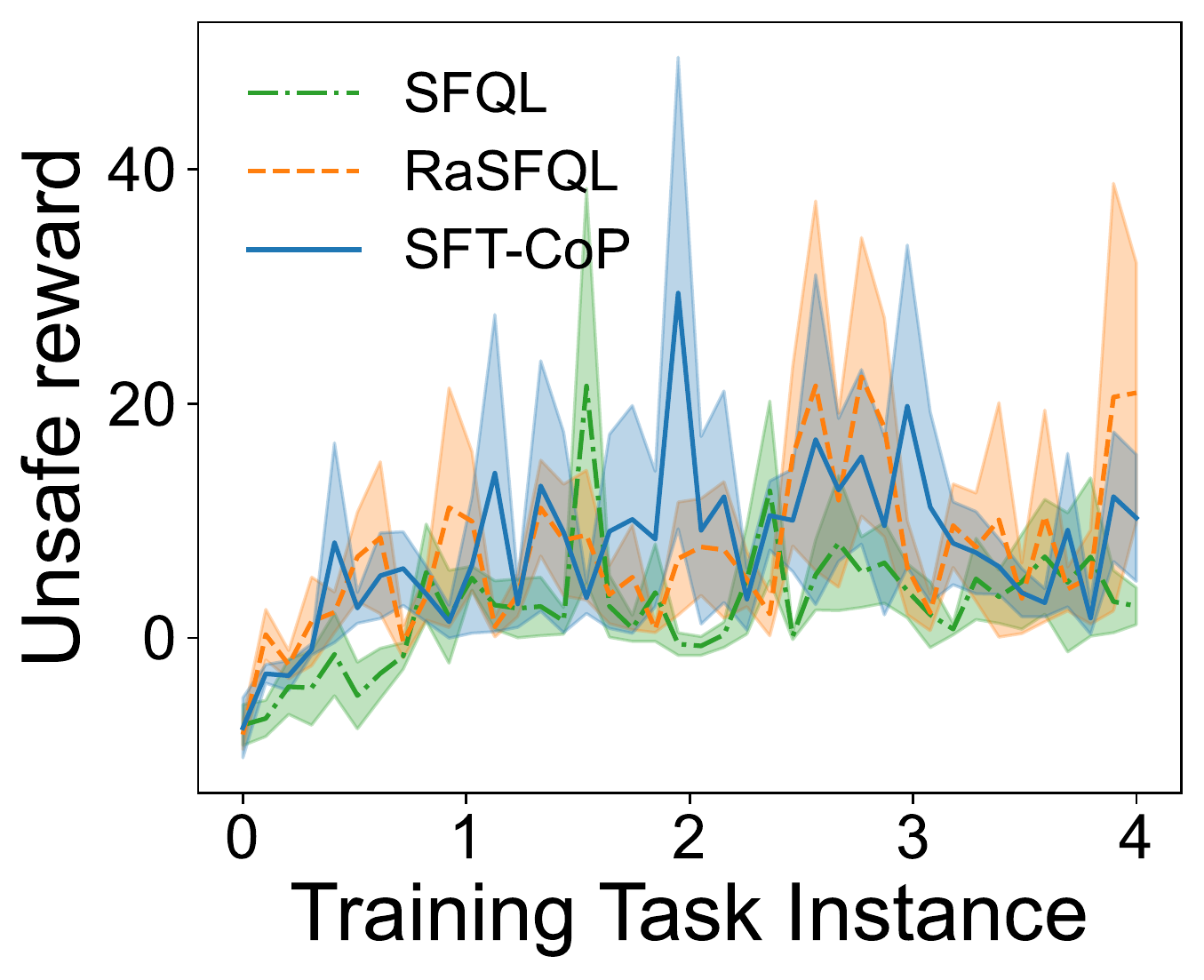}
	\end{subfigure}
	\begin{subfigure}[t]{.24\textwidth}
		\includegraphics[width=1.\textwidth]{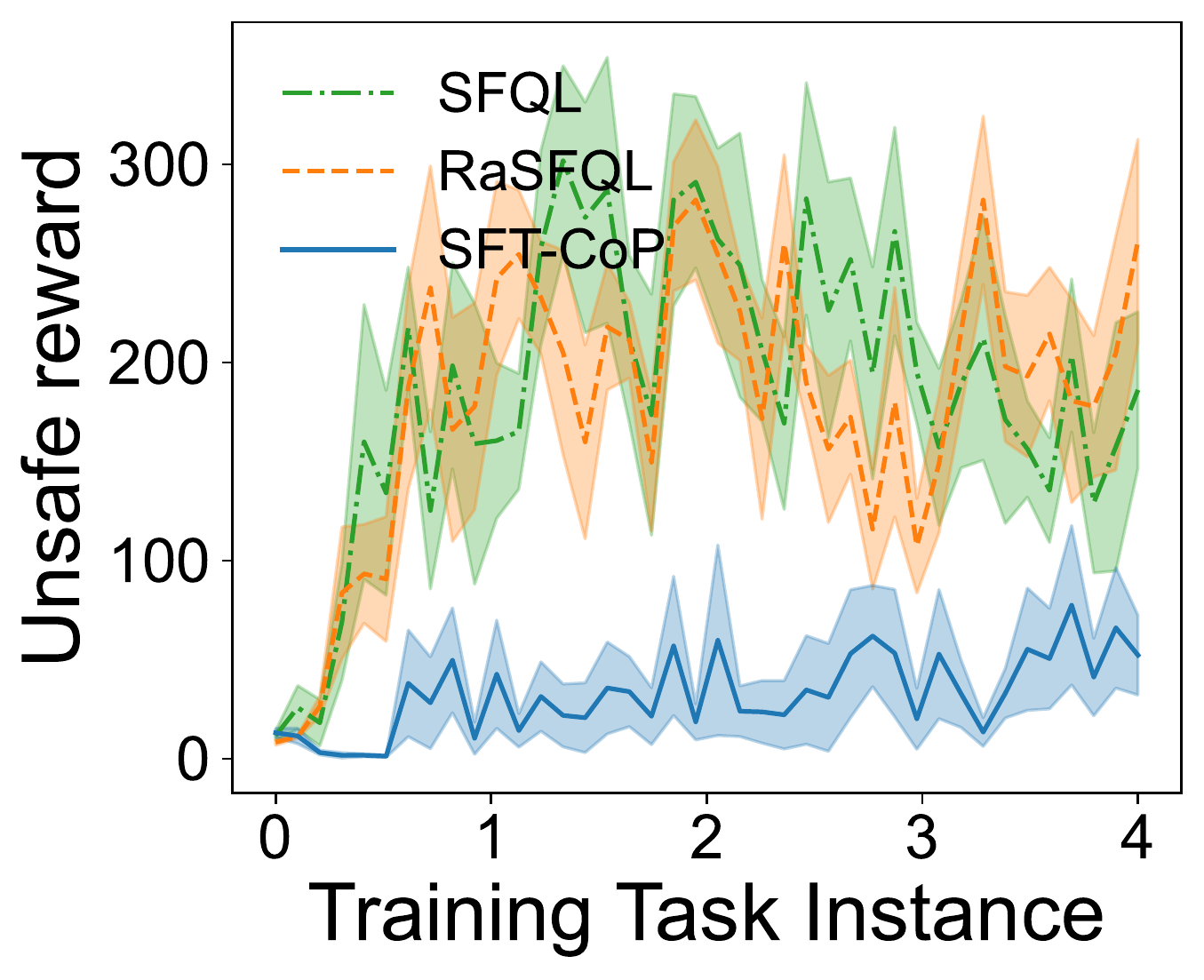}
	\end{subfigure}
	\begin{subfigure}[t]{.24\textwidth}
		\includegraphics[width=1.\textwidth]{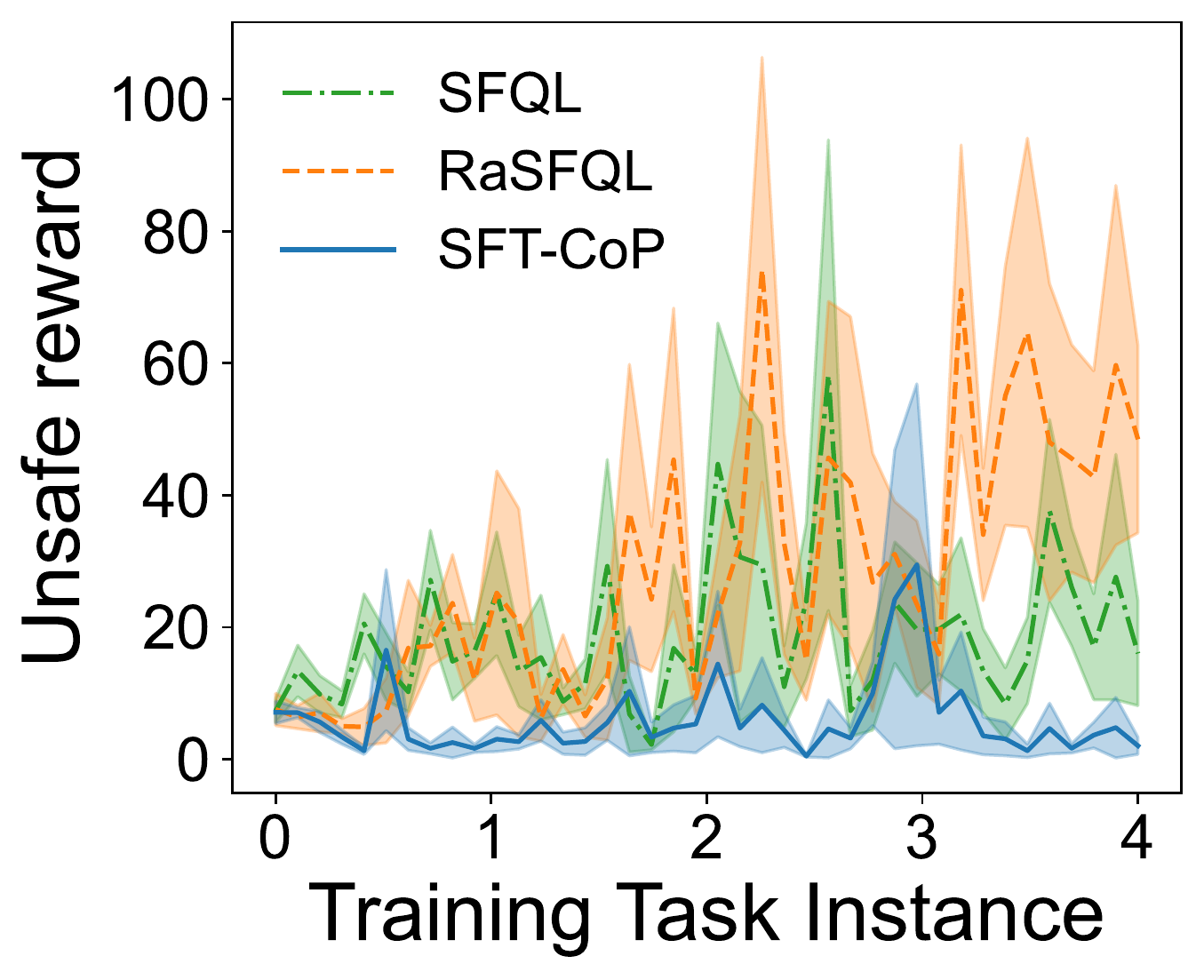}
	\end{subfigure}
	\begin{subfigure}[t]{.24\textwidth}
		\includegraphics[width=1.\textwidth]{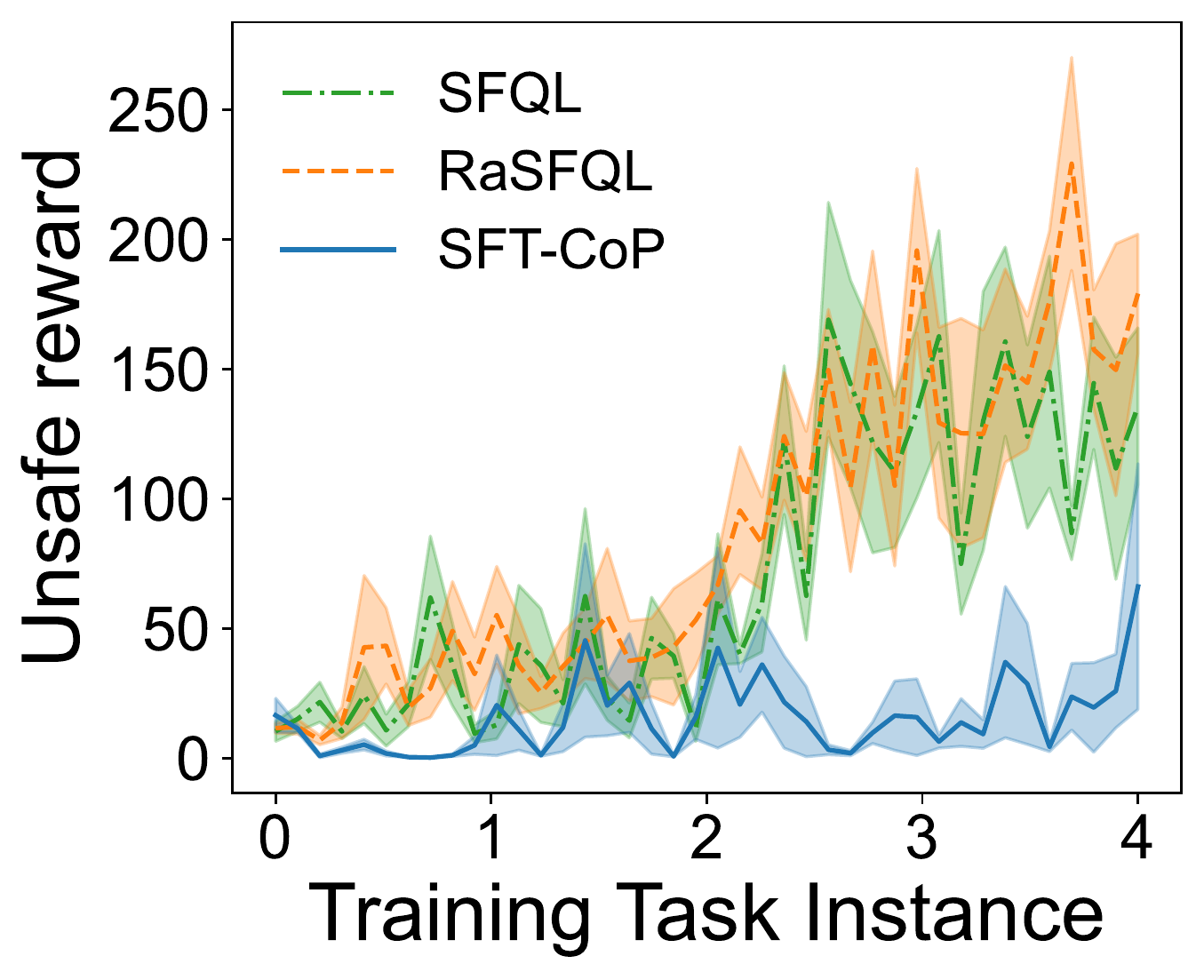}
	\end{subfigure}
	\begin{subfigure}[t]{.24\textwidth}
		\includegraphics[width=1.\textwidth]{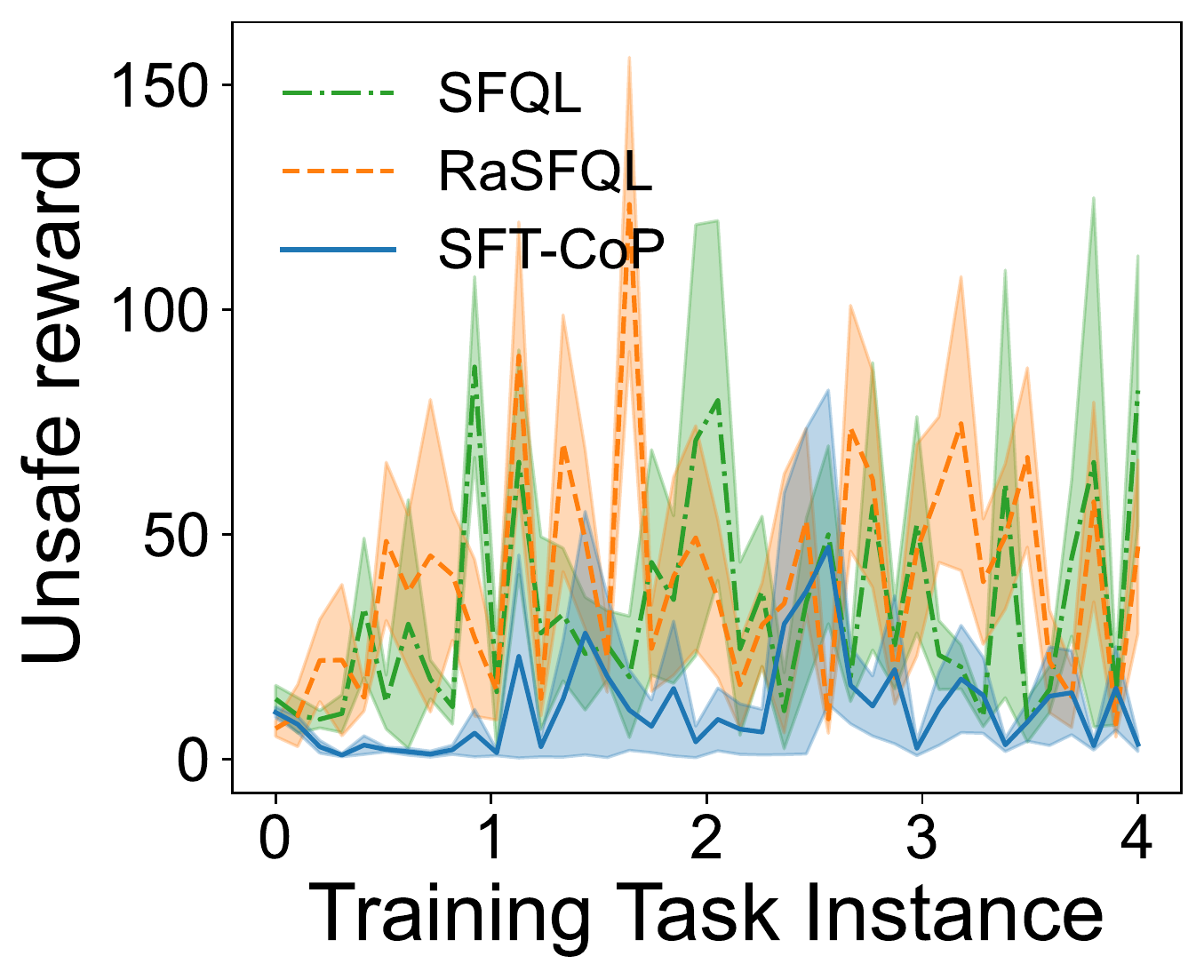}
	\end{subfigure}
	
	\caption{Performance of \sfql, \rasfql{} ($\beta=2$), and \method{} on the Reacher domain tested on 8 test tasks. We report averaged episode failures (top 3 rows), rewards from safe regions (middle 3 rows) and rewards from unsafe regions (bottom 3 rows), during the training course.}
	\label{fig:reacher_hplc_test}
\end{figure*}

Here we present all cumulative and per-task results of failures and different kinds of rewards in support of Fig. 2 and 3 in the main paper. Fig.~\ref{fig:non-transfer} plots additional results of non-transfer comparison. We have tuned the baseline methods and find that \pdql{} and \cpo{} can only converge after a larger number of steps (\eg~350,000 steps on reacher for \cpo{}) to learn a safe policy. Fig.~\ref{fig:4room_hplc}, Fig.~\ref{fig:reacher_hplc} and Fig.~\ref{fig:safetygym_hplc} plot the transfer learning results for Four-Room, Reacher and SafetyGym, respectively. Fig.~\ref{fig:reacher_hplc_test} present the test results of averaged episode failures and rewards on the 8 test tasks of Reacher.

%% file: main.bbl
\begin{thebibliography}{10}
\providecommand{\url}[1]{#1}
\csname url@samestyle\endcsname
\providecommand{\newblock}{\relax}
\providecommand{\bibinfo}[2]{#2}
\providecommand{\BIBentrySTDinterwordspacing}{\spaceskip=0pt\relax}
\providecommand{\BIBentryALTinterwordstretchfactor}{4}
\providecommand{\BIBentryALTinterwordspacing}{\spaceskip=\fontdimen2\font plus
\BIBentryALTinterwordstretchfactor\fontdimen3\font minus
  \fontdimen4\font\relax}
\providecommand{\BIBforeignlanguage}[2]{{%
\expandafter\ifx\csname l@#1\endcsname\relax
\typeout{** WARNING: IEEEtran.bst: No hyphenation pattern has been}%
\typeout{** loaded for the language `#1'. Using the pattern for}%
\typeout{** the default language instead.}%
\else
\language=\csname l@#1\endcsname
\fi
#2}}
\providecommand{\BIBdecl}{\relax}
\BIBdecl

\bibitem{mnih2015human}
V.~Mnih, K.~Kavukcuoglu, D.~Silver, A.~A. Rusu, J.~Veness, M.~G. Bellemare,
  A.~Graves, M.~Riedmiller, A.~K. Fidjeland, G.~Ostrovski \emph{et~al.},
  ``Human-level control through deep reinforcement learning,'' \emph{Nature},
  vol. 518, no. 7540, pp. 529--533, 2015.

\bibitem{Garcia2015}
J.~Garc{\'{i}}a and F.~Fern{\'{a}}ndez, ``A comprehensive survey on safe
  reinforcement learning,'' \emph{Journal of Machine Learning Research},
  vol.~16, no.~42, p. 1437–1480, 2015.

\bibitem{altman1999constrained}
E.~Altman, \emph{Constrained Markov Decision Processes}.\hskip 1em plus 0.5em
  minus 0.4em\relax CRC Press, 1999, vol.~7.

\bibitem{gimelfarb2021riskaware}
M.~Gimelfarb, A.~Barreto, S.~Sanner, and C.-G. Lee, ``Risk-aware transfer in
  reinforcement learning using successor features,'' in \emph{Advances in
  Neural Information Processing Systems}, 2021.

\bibitem{NIPS2017_350db081}
A.~Barreto, W.~Dabney, R.~Munos, J.~J. Hunt, T.~Schaul, H.~P. van Hasselt, and
  D.~Silver, ``Successor features for transfer in reinforcement learning,'' in
  \emph{Advances in Neural Information Processing Systems}, vol.~30.\hskip 1em
  plus 0.5em minus 0.4em\relax Curran Associates, Inc., 2017.

\bibitem{1994MDP}
M.~L. Puterman, \emph{Markov Decision Processes: Discrete Stochastic Dynamic
  Programming}, 1st~ed.\hskip 1em plus 0.5em minus 0.4em\relax USA: John Wiley
  \& Sons, Inc., 1994.

\bibitem{sutton2018reinforcement}
R.~S. Sutton and A.~G. Barto, \emph{Reinforcement learning: An
  introduction}.\hskip 1em plus 0.5em minus 0.4em\relax MIT press, 2018.

\bibitem{paternain2022safe}
S.~Paternain, M.~Calvo-Fullana, L.~F.~O. Chamon, and A.~Ribeiro, ``Safe
  policies for reinforcement learning via primal-dual methods,'' \emph{arXiv
  preprint arXiv:1911.09101}, 2022.

\bibitem{Ray2019}
A.~Ray, J.~Achiam, and D.~Amodei, ``Benchmarking safe exploration in deep
  reinforcement learning,'' \emph{arXiv preprint arXiv:1910.01708}, 2019.

\bibitem{pmlr-v80-barreto18a}
A.~Barreto, D.~Borsa, J.~Quan, T.~Schaul, D.~Silver, M.~Hessel, D.~Mankowitz,
  A.~Zidek, and R.~Munos, ``Transfer in deep reinforcement learning using
  successor features and generalised policy improvement,'' in \emph{Proceedings
  of the 35th International Conference on Machine Learning}, vol.~80.\hskip 1em
  plus 0.5em minus 0.4em\relax PMLR, 10--15 Jul 2018, pp. 501--510.

\bibitem{NEURIPS2019_c1aeb651}
S.~Paternain, L.~Chamon, M.~Calvo-Fullana, and A.~Ribeiro, ``Constrained
  reinforcement learning has zero duality gap,'' in \emph{Advances in Neural
  Information Processing Systems}, H.~Wallach, H.~Larochelle, A.~Beygelzimer,
  F.~d\textquotesingle Alch\'{e}-Buc, E.~Fox, and R.~Garnett, Eds.,
  vol.~32.\hskip 1em plus 0.5em minus 0.4em\relax Curran Associates, Inc.,
  2019.

\bibitem{JMLR:v10:taylor09a}
M.~E. Taylor and P.~Stone, ``Transfer learning for reinforcement learning
  domains: A survey,'' \emph{Journal of Machine Learning Research}, vol.~10,
  no.~56, pp. 1633--1685, 2009.

\bibitem{zhu2020transfer}
Z.~Zhu, K.~Lin, and J.~Zhou, ``Transfer learning in deep reinforcement
  learning: A survey,'' \emph{arXiv preprint arXiv:2009.07888}, 2020.

\bibitem{brunke2021safe}
L.~Brunke, M.~Greeff, A.~W. Hall, Z.~Yuan, S.~Zhou, J.~Panerati, and A.~P.
  Schoellig, ``Safe learning in robotics: From learning-based control to safe
  reinforcement learning,'' \emph{arXiv preprint arXiv:2108.06266}, 2021.

\bibitem{NIPS2017_766ebcd5}
F.~Berkenkamp, M.~Turchetta, A.~Schoellig, and A.~Krause, ``Safe model-based
  reinforcement learning with stability guarantees,'' in \emph{Advances in
  Neural Information Processing Systems}, I.~Guyon, U.~V. Luxburg, S.~Bengio,
  H.~Wallach, R.~Fergus, S.~Vishwanathan, and R.~Garnett, Eds., vol.~30.\hskip
  1em plus 0.5em minus 0.4em\relax Curran Associates, Inc., 2017.

\bibitem{Cheng_Orosz_Murray_Burdick_2019}
R.~Cheng, G.~Orosz, R.~M. Murray, and J.~W. Burdick, ``End-to-end safe
  reinforcement learning through barrier functions for safety-critical
  continuous control tasks,'' \emph{Proceedings of the AAAI Conference on
  Artificial Intelligence}, vol.~33, no.~01, pp. 3387--3395, Jul. 2019.

\bibitem{10.1145/1390156.1390225}
A.~Lazaric, M.~Restelli, and A.~Bonarini, ``Transfer of samples in batch
  reinforcement learning,'' in \emph{Proceedings of the 25th International
  Conference on Machine Learning}, ser. ICML '08.\hskip 1em plus 0.5em minus
  0.4em\relax New York, NY, USA: Association for Computing Machinery, 2008, p.
  544–551.

\bibitem{rajendran2015attend}
J.~Rajendran, A.~Srinivas, M.~M. Khapra, P.~Prasanna, and B.~Ravindran,
  ``Attend, adapt and transfer: Attentive deep architecture for adaptive
  transfer from multiple sources in the same domain,'' in \emph{International
  Conference on Learning Representations}, 2017.

\bibitem{pmlr-v139-vaezipoor21a}
P.~Vaezipoor, A.~C. Li, R.~A.~T. Icarte, and S.~A. Mcilraith, ``Ltl2action:
  Generalizing ltl instructions for multi-task rl,'' in \emph{Proceedings of
  the 38th International Conference on Machine Learning}, ser. Proceedings of
  Machine Learning Research, M.~Meila and T.~Zhang, Eds., vol. 139.\hskip 1em
  plus 0.5em minus 0.4em\relax PMLR, 18--24 Jul 2021, pp. 10\,497--10\,508.

\bibitem{pmlr-v139-araki21a}
B.~Araki, X.~Li, K.~Vodrahalli, J.~Decastro, M.~Fry, and D.~Rus, ``The logical
  options framework,'' in \emph{Proceedings of the 38th International
  Conference on Machine Learning}, vol. 139.\hskip 1em plus 0.5em minus
  0.4em\relax PMLR, 18--24 Jul 2021, pp. 307--317.

\bibitem{NIPS2009_3eb71f62}
E.~Todorov, ``Compositionality of optimal control laws,'' in \emph{Advances in
  Neural Information Processing Systems}, Y.~Bengio, D.~Schuurmans,
  J.~Lafferty, C.~Williams, and A.~Culotta, Eds., vol.~22.\hskip 1em plus 0.5em
  minus 0.4em\relax Curran Associates, Inc., 2009.

\bibitem{NEURIPS2020_6ba3af5d}
G.~Nangue~Tasse, S.~James, and B.~Rosman, ``A boolean task algebra for
  reinforcement learning,'' in \emph{Advances in Neural Information Processing
  Systems}, vol.~33.\hskip 1em plus 0.5em minus 0.4em\relax Curran Associates,
  Inc., 2020, pp. 9497--9507.

\bibitem{Konidaris2006AutonomousSK}
G.~Konidaris and A.~Barto, ``Autonomous shaping: Knowledge transfer in
  reinforcement learning,'' in \emph{Proceedings of the 23rd International
  Conference on Machine Learning}, 2006, p. 489–496.

\bibitem{DayanSR}
P.~Dayan, ``Improving generalization for temporal difference learning: The
  successor representation,'' \emph{Neural Computation}, vol.~5, no.~4, pp.
  613--624, 1993.

\bibitem{pmlr-v139-abdolshah21a}
M.~Abdolshah, H.~Le, T.~K. George, S.~Gupta, S.~Rana, and S.~Venkatesh, ``A new
  representation of successor features for transfer across dissimilar
  environments,'' in \emph{Proceedings of the 38th International Conference on
  Machine Learning}.\hskip 1em plus 0.5em minus 0.4em\relax PMLR, 18--24 Jul
  2021, pp. 1--9.

\bibitem{ma2020universal}
C.~Ma, D.~R. Ashley, J.~Wen, and Y.~Bengio, ``Universal successor features for
  transfer reinforcement learning,'' \emph{arXiv preprint arXiv:2001.04025},
  2020.

\bibitem{alver2021constructing}
S.~Alver and D.~Precup, ``Constructing a good behavior basis for transfer using
  generalized policy updates,'' \emph{arXiv preprint arXiv:2112.15025}, 2021.

\bibitem{HEGER1994105}
M.~Heger, ``Consideration of risk in reinforcement learning,'' in \emph{Machine
  Learning Proceedings 1994}, W.~W. Cohen and H.~Hirsh, Eds.\hskip 1em plus
  0.5em minus 0.4em\relax San Francisco (CA): Morgan Kaufmann, 1994, pp.
  105--111.

\bibitem{10.2307/2629352}
R.~A. Howard and J.~E. Matheson, ``Risk-sensitive markov decision processes,''
  \emph{Management Science}, vol.~18, no.~7, pp. 356--369, 1972.

\bibitem{tamar2014policy}
A.~Tamar, Y.~Glassner, and S.~Mannor, ``Policy gradients beyond expectations:
  Conditional value-at-risk,'' \emph{arXiv preprint arXiv:1404.3862}, 2014.

\bibitem{BORKAR2005207}
V.~Borkar, ``An actor-critic algorithm for constrained markov decision
  processes,'' \emph{Systems \& Control Letters}, vol.~54, no.~3, pp. 207--213,
  2005.

\bibitem{tessler2018reward}
C.~Tessler, D.~J. Mankowitz, and S.~Mannor, ``Reward constrained policy
  optimization,'' in \emph{International Conference on Learning
  Representations}, 2019.

\bibitem{achiam2017constrained}
J.~Achiam, D.~Held, A.~Tamar, and P.~Abbeel, ``Constrained policy
  optimization,'' in \emph{Proceedings of the 34th International Conference on
  Machine Learning}, vol.~70.\hskip 1em plus 0.5em minus 0.4em\relax PMLR,
  06--11 Aug 2017, pp. 22--31.

\bibitem{yang2020projectionbased}
T.-Y. Yang, J.~Rosca, K.~Narasimhan, and P.~J. Ramadge, ``Projection-based
  constrained policy optimization,'' in \emph{International Conference on
  Learning Representations}, 2020.

\bibitem{NEURIPS2018_4fe51490}
Y.~Chow, O.~Nachum, E.~Duenez-Guzman, and M.~Ghavamzadeh, ``A lyapunov-based
  approach to safe reinforcement learning,'' in \emph{Advances in Neural
  Information Processing Systems}, vol.~31.\hskip 1em plus 0.5em minus
  0.4em\relax Curran Associates, Inc., 2018.

\bibitem{NEURIPS2019_873be070}
S.~Miryoosefi, K.~Brantley, H.~Daume~III, M.~Dudik, and R.~E. Schapire,
  ``Reinforcement learning with convex constraints,'' in \emph{Advances in
  Neural Information Processing Systems}, vol.~32, 2019.

\bibitem{gattami2019reinforcement}
A.~Gattami, Q.~Bai, and V.~Agarwal, ``Reinforcement learning for
  multi-objective and constrained markov decision processes,'' \emph{arXiv
  preprint arXiv:1901.08978}, 2019.

\bibitem{pmlr-v97-le19a}
H.~Le, C.~Voloshin, and Y.~Yue, ``Batch policy learning under constraints,'' in
  \emph{Proceedings of the 36th International Conference on Machine Learning},
  ser. Proceedings of Machine Learning Research, K.~Chaudhuri and
  R.~Salakhutdinov, Eds., vol.~97.\hskip 1em plus 0.5em minus 0.4em\relax PMLR,
  09--15 Jun 2019, pp. 3703--3712.

\bibitem{NEURIPS2020_5f7695de}
D.~Ding, K.~Zhang, T.~Basar, and M.~Jovanovic, ``Natural policy gradient
  primal-dual method for constrained markov decision processes,'' in
  \emph{Advances in Neural Information Processing Systems}, H.~Larochelle,
  M.~Ranzato, R.~Hadsell, M.~F. Balcan, and H.~Lin, Eds., vol.~33.\hskip 1em
  plus 0.5em minus 0.4em\relax Curran Associates, Inc., 2020, pp. 8378--8390.

\bibitem{wei2021provably}
H.~Wei, X.~Liu, and L.~Ying, ``A provably-efficient model-free algorithm for
  constrained markov decision processes,'' \emph{arXiv preprint
  arXiv:2106.01577}, 2021.

\bibitem{Ethan2020}
E.~Knight and J.~Achiam, ``Safely transferring to unsafe environments with
  constrained reinforcement learning,'' in \emph{International Conference on
  Learning Representations Workshop: Beyond `tabula rasa' in reinforcement
  learning ({BeTR-RL})}, 2020.

\bibitem{haarnoja2018soft}
T.~Haarnoja, A.~Zhou, P.~Abbeel, and S.~Levine, ``Soft actor-critic: Off-policy
  maximum entropy deep reinforcement learning with a stochastic actor,''
  \emph{arXiv preprint arXiv:1801.01290}, 2018.

\bibitem{anstreicher2009two}
K.~M. Anstreicher and L.~A. Wolsey, ``Two “well-known” properties of
  subgradient optimization,'' \emph{Mathematical Programming}, vol. 120, no.~1,
  pp. 213--220, 2009.

\bibitem{boyd2004convex}
S.~Boyd, S.~P. Boyd, and L.~Vandenberghe, \emph{Convex optimization}.\hskip 1em
  plus 0.5em minus 0.4em\relax Cambridge university press, 2004.

\bibitem{shapiro2002}
A.~Shapiro and A.~Kleywegt, ``Minimax analysis of stochastic problems,''
  \emph{Optimization Methods and Software}, vol.~17, no.~3, pp. 523--542, 2002.

\bibitem{benelot2018}
B.~Ellenberger, ``Pybullet gymperium,'' \url{
  https://github.com/benelot/pybullet-gym}, 2018--2019.

\bibitem{conf/iros/TodorovET12}
E.~Todorov, T.~Erez, and Y.~Tassa, ``Mujoco: A physics engine for model-based
  control.'' in \emph{IROS}.\hskip 1em plus 0.5em minus 0.4em\relax IEEE, 2012,
  pp. 5026--5033.

\end{thebibliography}
